\documentclass[10pt,journal,compsoc]{IEEEtran}
\ifCLASSOPTIONcompsoc
\usepackage[nocompress]{cite}
\else
\usepackage{cite}
\fi
\ifCLASSINFOpdf
\else
\fi
\usepackage{amsthm}
\usepackage{amssymb}
\usepackage[fleqn]{amsmath}
\usepackage{microtype}
\usepackage{pgfplots, pgfplotstable, filecontents}
\usepackage{tikz}
\usetikzlibrary{patterns}

\usepackage{adjustbox}
\usepackage[linesnumbered,ruled]{algorithm2e}
\usepackage{setspace}
\SetKwComment{Comment}{$\///$\ }{}
\SetKwRepeat{Do}{do}{while}%
\newlength\mylen

\usepackage{tabularx}
\usepackage{graphicx}
\usepackage{subfig}
\usepackage{multirow}
\usepackage{booktabs}
\usepackage{url} 
\usepackage{threeparttable}
\usepackage{booktabs,makecell,siunitx}

\newcommand\copyrighttext{%
	\footnotesize This work has been submitted to the IEEE for possible publication. Copyright may be transferred without notice, after which this version may no longer be accessible.}
\newcommand\copyrightnotice{%
	\begin{tikzpicture}[remember picture,overlay]
		\node[anchor=south,yshift=10pt] at (current page.south) {\fbox{\parbox{\dimexpr\textwidth-\fboxsep-\fboxrule\relax}{\copyrighttext}}};
	\end{tikzpicture}%
}

\newenvironment{pics}
{\par\raggedright  \centering
	\setlength\tabcolsep{2pt}%
	\begin{tabular}{*{10}c}}
	{\end{tabular}\par}
\newtheorem{theorem}{Theorem}

\graphicspath{{fig/}}

\begin{document}
	
	\title{Angular triangle distance for ordinal metric learning}
	
	\author{Imam~Mustafa~Kamal,
		Hyerim~Bae
		\IEEEcompsocitemizethanks{\IEEEcompsocthanksitem IM. Kamal was with the Institute of Intelligent Logistics Big Data, Pusan National University, 30-Jan-jeon Dong, Geum-Jeong Gu, 609-753, Busan, South Korea, Email: imamkamal@pusan.ac.kr.
			\IEEEcompsocthanksitem H. Bae was with a major in Industrial Data Science \& Engineering, Department of Industrial Engineering, Pusan National University, 30-Jan-jeon Dong, Geum-Jeong Gu, 609-753, Busan, South Korea, Email: hrbae@pusan.ac.kr (Corresponding author).}% <-this % stops an unwanted space
		}

	\IEEEtitleabstractindextext{%
		\begin{abstract}
			Deep metric learning (DML) aims to automatically construct task-specific distances or similarities of data, resulting in a low-dimensional representation. Several significant metric-learning methods have been proposed. Nonetheless, no approach guarantees the preservation of the ordinal nature of the original data in a low-dimensional space. Ordinal data are ubiquitous in real-world problems, such as the severity of symptoms in biomedical cases, production quality in manufacturing, rating level in businesses, and aging level in face recognition. This study proposes a novel angular triangle distance (ATD) and ordinal triplet network (OTD) to obtain an accurate and meaningful embedding space representation for ordinal data. The ATD projects the ordinal relation of data in the angular space, whereas the OTD learns its ordinal projection. We also demonstrated that our new distance measure satisfies the distance metric properties mathematically. The proposed method was assessed using real-world data with an ordinal nature, such as biomedical, facial, and hand-gestured images. Extensive experiments have been conducted, and the results show that our proposed method not only semantically preserves the ordinal nature but is also more accurate than existing DML models. Moreover, we also demonstrate that our proposed method outperforms the state-of-the-art ordinal metric learning method.

		\end{abstract}
		
		\begin{IEEEkeywords}
			Deep metric learning, ordinal distance, ordinal feature, dimensionality reduction, siamese network, ordinal classification.
	\end{IEEEkeywords}}

	\maketitle
	\copyrightnotice
	
	\IEEEdisplaynontitleabstractindextext

	% \fi
	\IEEEpeerreviewmaketitle

	\IEEEraisesectionheading{\section{Introduction}\label{sec:introduction}}
	
	\IEEEPARstart{W}{ith} the emergence of deep learning, DML has gained significant popularity because it can seamlessly incorporate the strength of deep learning into metric learning \cite{Zheng_2019_CVPR}. The DML automatically constructs task-specific distance metrics in a weakly supervised manner. This results in a low-dimensional embedding representation of the data that preserves the distance between similar data points close to each other and dissimilar data points far from the embedding space \cite{NEURIPS2020_d9812f75}. DML is used in many machine-learning applications because it is motivated by large-scale multimedia data, which are ubiquitous in the modern era \cite{NEURIPS2020_ce016f59}. The learned distance metric in the embedding space representation can be employed to perform simple knowledge-based discovery algorithms, such as the nearest neighbor classifier, hierarchical agglomerative clustering, and information retrieval. Because the data representation in the embedding space is crucial for obtaining a meaningful or semantic representation, the data type (measurement scales) can affect the algorithm's performance.
	
	The level of data measurement determines the embedding space representation. Unlike nominal data, ordinal data are particular categorical data with an ordered nature in the label. Ordinal data are common in the real world. In biomedical cases, cancer severity is classified as normal, benign, or malignant. In manufacturing, the production quality is categorized as quality I (awful), II (low), III (medium), IV (high), and V (excellent). In business cases, the rating level of services is graded as bad, medium, high, or excellent. In face recognition cases, age is grouped as a child, adolescent, adult, and senior adult. Although they have an explicit class ordering, the absolute distances between them remain unknown \cite{NGUYEN201817, TANG202172}. Existing DML methods have achieved an adequate accuracy in ordinal data, although they ignore the inherent ordinal relationship among the labels. Accordingly, the relationships among the categories or labels in the embedding space representation can be semantically scattered. Thus, when the DML encounters ordinal data, it can deviate from its original objective. It is unable to maintain the semantic relation of the original data in the embedding space representation. Nevertheless, generating a DML method to simultaneously deal with embedding precision while maintaining its ordinal nature is a non-trivial task.
	
	In addition, machine learning algorithms frequently require measuring the distance between or among data points. Traditionally, scholars have used previous domain knowledge to choose a standard distance measure, such as Euclidean and cosine similarity. The length of the line segment that connects two points in Euclidean space is the Euclidean distance. The Euclidean distance is powerful; however, it becomes weakly discriminant in high-dimensional datasets. Unlike the Euclidean distance, cosine similarity is determined by the angle of the vectors rather than their magnitudes. In contrast, the term cosine distance is commonly used to complement cosine similarity in a positive space. However, it is worth noting that the cosine distance is not a proper distance metric because it lacks the triangle inequality property. Consequently, these standard distance measures, which existing DML methods commonly rely on, may not be perfectly suitable for dealing with ordinal data.
	
	In this study, we propose a new DML model with a novel distance measure that preserves the ordered nature of ordinal data semantically while obtaining an accurate embedding representation. We extended the angular distance using a triplet representation called the ATD. We mathematically demonstrate that our new distance satisfies basic distance properties, such as non-negativity, identity, symmetry, and triangle inequality. Moreover, a new OTD with triplet input and ordinal regression fashion was devised to learn and maintain the sequence of ordinal data in the embedding space representation. The contributions of this study can be summarized as follows. First, given ordinal data, we demonstrated that existing DML methods are unable to maintain the ordinal nature of data in embedding space representations, even though they obtain high accuracy. Second, we propose a novel OTD with an ATD to solve the ordinal DML. Finally, we demonstrate that our proposed method can semantically preserve the ordinal nature and result in a more accurate embedding representation than existing DML models.
	
	The remainder of this paper is organized as follows. Section \ref{sec:related_work} reviews the literature on the DML models and ordinal classifications. Section \ref{sec:method} describes the proposed method. An extensive experiment and discussion are presented in Section \ref{sec:experiment}, and we outline some conclusions, limitations, and future directions for this research in Section \ref{sec:conclusion}.
	
	\section{Related work}
	\label{sec:related_work}
	Several linear metric learning methods addressing ordinal data have been proposed. Li et al. \cite{6881672} were the first to introduce ordinal metric learning by a designed weighting factor based on ordinal labels; thus, the ordinal relationship of images is expected to be preserved. Yu and Li \cite{DBLP:journals/corr/abs-1902-10284} introduced a multidimensional scaling with Euclidean distance for ordinal metric learning. It has low computational complexity compared with modern metric learning (DML); however, this method merely works under the assumption of a positive definite matrix. Shi et al. \cite{Shi_Li_Sha_2016} introduced linear metric learning for ordinal data using a large margin nearest neighbor, which is computationally intensive for high-dimensional and large-scale data such as images. All of the aforementioned studies can be categorized as conventional metric learning, which has a limited capability to capture non-linearity in particular. In contrast, DML uses (deep) neural networks to automatically learn discriminative features from the data and construct metrics. Therefore, DML can accomplish effective results with non-linear and complex datasets, as demonstrated in recent deep learning research. DML emerged from weakly supervised metric learning methods, leveraging the capacity of deep neural networks \cite{NIPS2002_c3e4035a}. In DML, many efforts have been made to develop pair-based loss functions, such as the Siamese network. The Siamese network \cite{Chicco2021,NIPS1993_288cc0ff} is simple and effective for mapping similarity and dissimilarity samples in the embedding space. Triplet loss \cite{7298682} indicates that the distance of a negative pair is larger than that of a positive pair by a certain margin. Quadruplet loss \cite{DBLP:conf/cvpr/ChenCZH17} is used to devise a larger inter-class variation and a smaller intra-class variation than the triplet loss. The $N$-pair loss \cite{NIPS2016_6b180037} extended the triplet loss and ensured that more than one negative sample simultaneously moved farther away from the anchor than did the positive sample. Nonetheless, these methods treat ordinal data the same as nominal data, which leads to a non-optimum semantic embedding representation.
	
	Ordinal classification, also known as ordinal regression, classifies patterns into naturally ordered labels. Ordinal regression is a relatively new learning paradigm aimed at learning a rule for predicting ordered categories. However, it is essential to note that most previous studies on distance metric learning cannot be directly applied to ordinal classification. Usually, additional constraints are used to indicate ordinal relations between examples of different classes \cite{TANG2021107358}. Metric learning has also been reported to improve classification model performance \cite{SENDIK2019368}. Some algorithms have been proposed to address ordinal regression problems from a machine learning perspective. The simplest idea is to convert ordinal regression into regular regression. Tang et al. \cite{TANG202172} used distance metric learning, which incorporates absolute and relative information, to solve ordinal regression. Nguyen et al. \cite{NGUYEN201817} devised a distance metric learning for ordinal regression by incorporating local triplet constraints containing ordering information into a conventional large-margin distance metric learning approach. Nevertheless, unlike metric learning research, most ordinal regression study focuses on ordinal-continuous data (regression) rather than ordinal-discrete data (classification), which lacks the embedding space representation.
	
	\section{Methodology}
	\label{sec:method}
	The framework of this study comprises a two-fold approach: angular triangle distance and an ordinal triplet network. The angular triangle distance projects the ordinal relation of data in the angular space, whereas the ordinal triplet network learns its ordinal projection. The problem formulation, network design, and learning mechanism of our proposed framework are described below.
	\subsection{Problem formulation}
	\label{subsec:problem_formulation}
	
	We denote $X$ as a set of high-dimensional data with $N$ total amount of data, $X = \{x^{l_{r}}_i | i = 0,1,2,\dotsc,N-1, r = 0,1,2,\dotsc,C-1\}$. $l$ represents the ordinal label with $C$ total number of categories. The relationship among the categories can be denoted as $l_0<l_1<l_2<\dotsc<l_{C-1}$. In this study, we assume that $C \geq 3$ because when $C=2$, it becomes a binary classification problem. The embedding network ($F$) maps a high-dimensional input $X$ to a low-dimensional output $Z$ ($F: X \rightarrow Z$), where $Z = \{z^{l_{r}}_i | i = 0,1,2,\dotsc,N-1, r = 0,1,2,\dotsc,C-1\}$. $Z$ is a vector that reflects the latent representation of $X$ in the embedding space. Let us define two vectors, $z^{l_{r_i}}_i$ and $z^{l_{r_j}}_j$ from $Z$, that correspond to the latent representations of $x^{l_{r_i}}_i$ and $x^{l_{r_j}}_j$, respectively, where $i \neq j$. Their cosine similarity is denoted as ${S_C}(z^{l_{r_i}}_i,z^{l_{r_j}}_j) = \frac{z^{l_{r_i}}_i \cdot z^{l_{r_j}}_j}{\ \| z^{l_{r_i}}_i \|\| z^{l_{r_j}}_j \|}$. The cosine similarity results in the interval $[-1,1]$, which lies in the first and second quadrants of the Cartesian coordinate system; where the ${S_C}(z^{l_{r_i}}_i,z^{l_{r_j}}_j) \approx 1$, ${S_C}(z^{l_{r_i}}_i,z^{l_{r_j}}_j) \approx 0$, ${S_C}(z^{l_{r_i}}_i,z^{l_{r_j}}_j) \approx -1$, if $z^{l_{r_i}}_i$ and $z^{l_{r_j}}_j$ are proportional, orthogonal, or opposite vectors, respectively. In contrast to the similarity measure, which allows both positive and negative directions, the distance metric solely encloses a positive direction. Thus, the term cosine distance is commonly used to complement the cosine similarity in the positive space $[0,1]$. As a result, because the original space is constrained to only the first quadrant of the Cartesian coordinate system, the cosine distance is less expressive than the original cosine similarity. Moreover, because it does not satisfy the triangle inequality property, it may not be a proper distance metric.
	
	\subsection{Angular triangle distance (ATD)}
	\label{subsec:ortho_tri_dist}
	In this section, we introduce our new distance metric, based on the angular distance devised for an ordinal DML. To address the issue of the triangle inequality property in cosine distance, it is necessary to convert it into an angular distance $D_A$. When the vector elements are positive or negative, the $D_A$ is expressed using Eq. \ref{eq:angular_dist} as follows:
	\begin{equation}
		\label{eq:angular_dist}
		{D_A}(z^{l_{r_i}}_i,z^{l_{r_j}}_j) = \frac{{\cos}^{-1}({S_C}(z^{l_{r_i}}_i,z^{l_{r_j}}_j))}{\pi} = \frac{\theta_{z^{l_{r_i}}_i,z^{l_{r_j}}_j}}{\pi}
	\end{equation}
	The ${\cos}^{-1}$ corresponds to $\arccos$, which is the inverse of the $\cos$ function. Thus, it obtains an angle between $z^{l_{r_i}}_i$ and $z^{l_{r_j}}_j$ $(\theta_{z^{l_{r_i}}_i,z^{l_{r_j}}_j})$ rather than a scalar value. Similar to the cosine similarity, the angle space of $D_A$ lies in the first and second quadrants of the Cartesian coordinate system. Therefore, the ${D_A}(z^{l_{r_i}}_i,z^{l_{r_j}}_j)$ functions are bounded between $0$ and $1$, inclusive, by normalizing them with $\pi=180^{\circ}$.
	
	\begin{figure*}[t]
		\centering
		\subfloat[]{\includegraphics[width=0.24\linewidth]{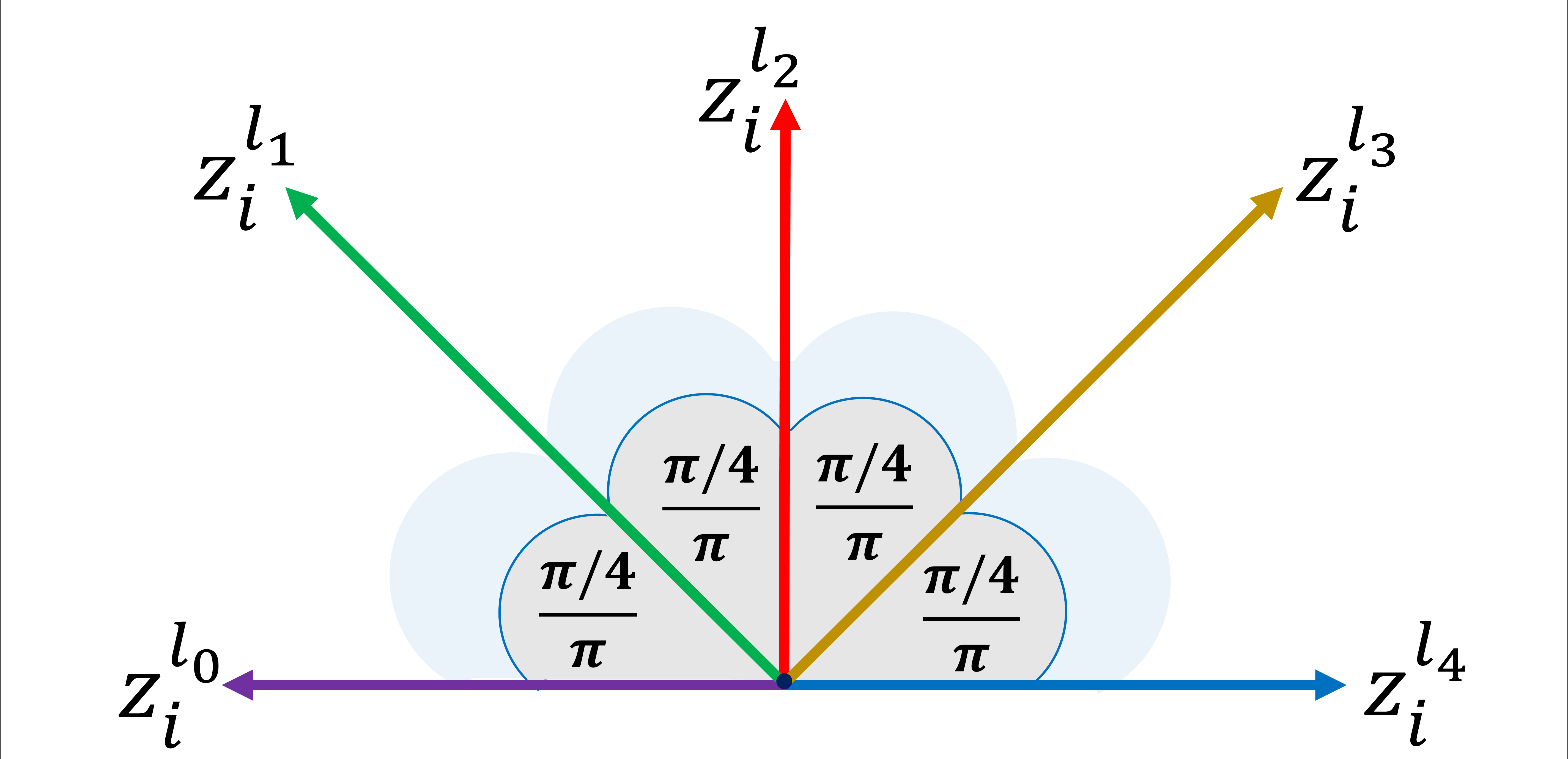}
			\label{ortho_dist1}}%
		\subfloat[]{\includegraphics[width=0.24\linewidth]{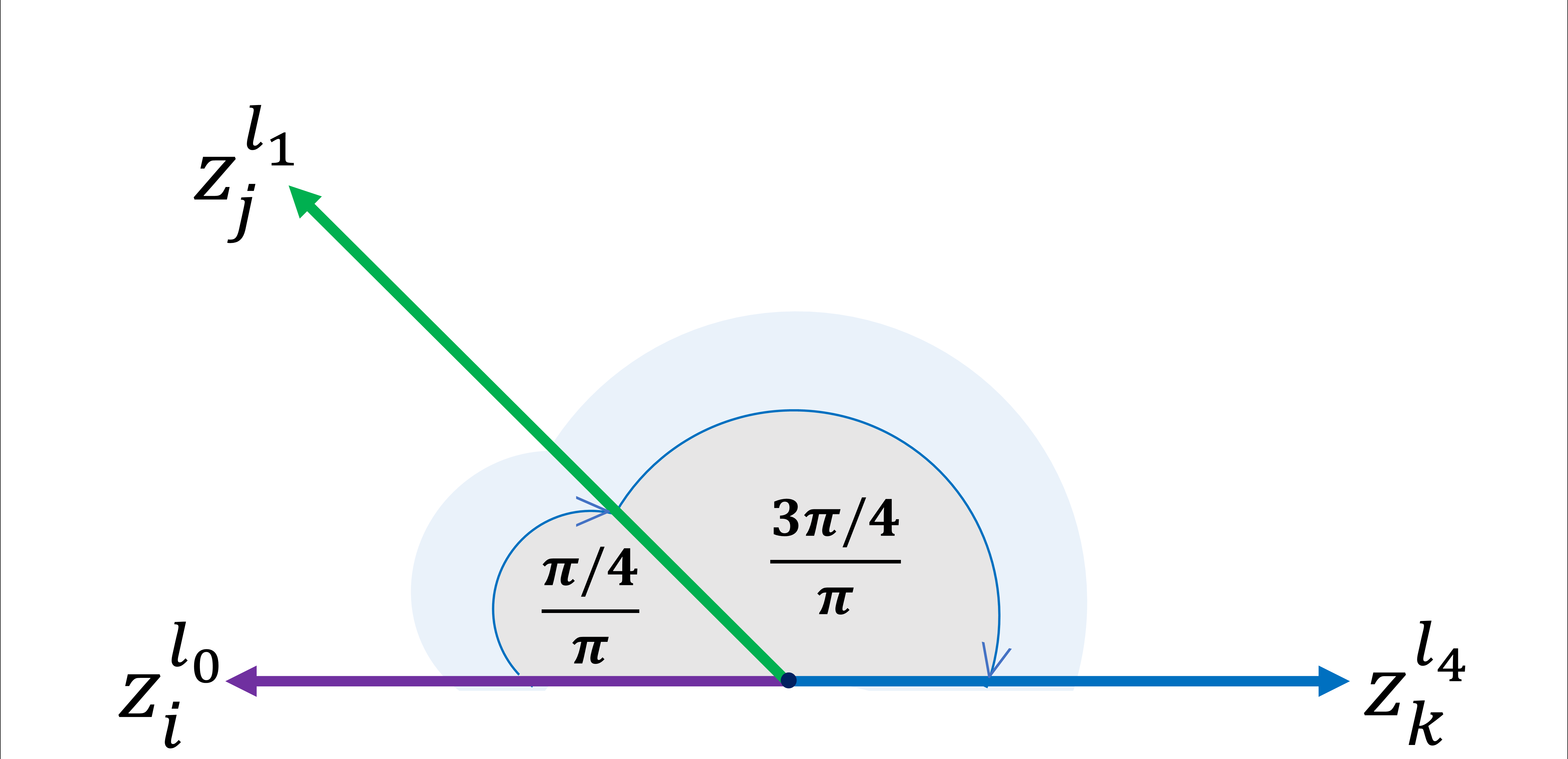}
			\label{ortho_dist2}}%
		\subfloat[]{\includegraphics[width=0.24\linewidth]{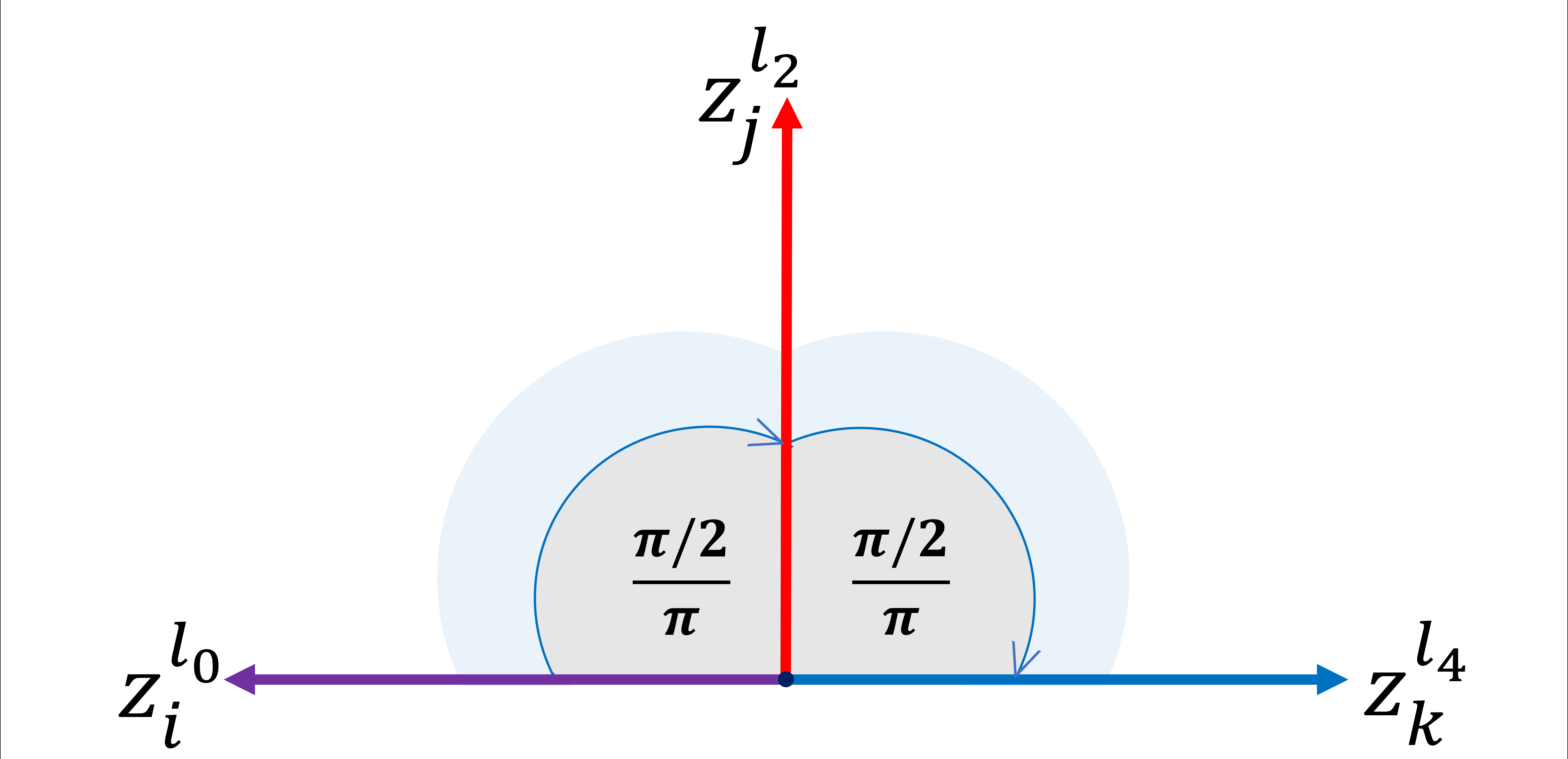}
			\label{ortho_dist3}}\\\hspace*{1pt}
		\subfloat[]{\includegraphics[width=0.24\linewidth]{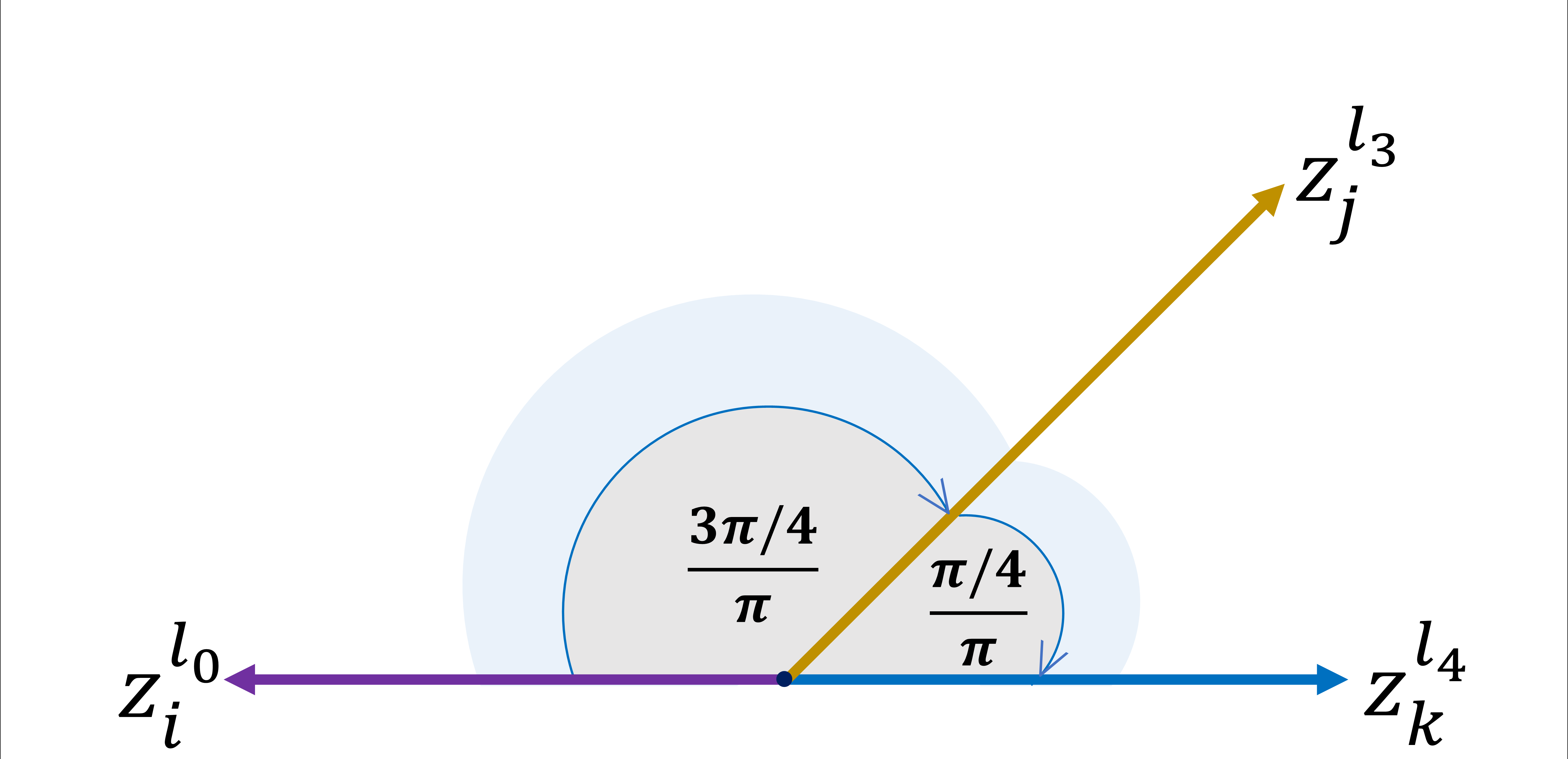}
			\label{ortho_dist4}}%
		\subfloat[]{\includegraphics[width=0.24\linewidth]{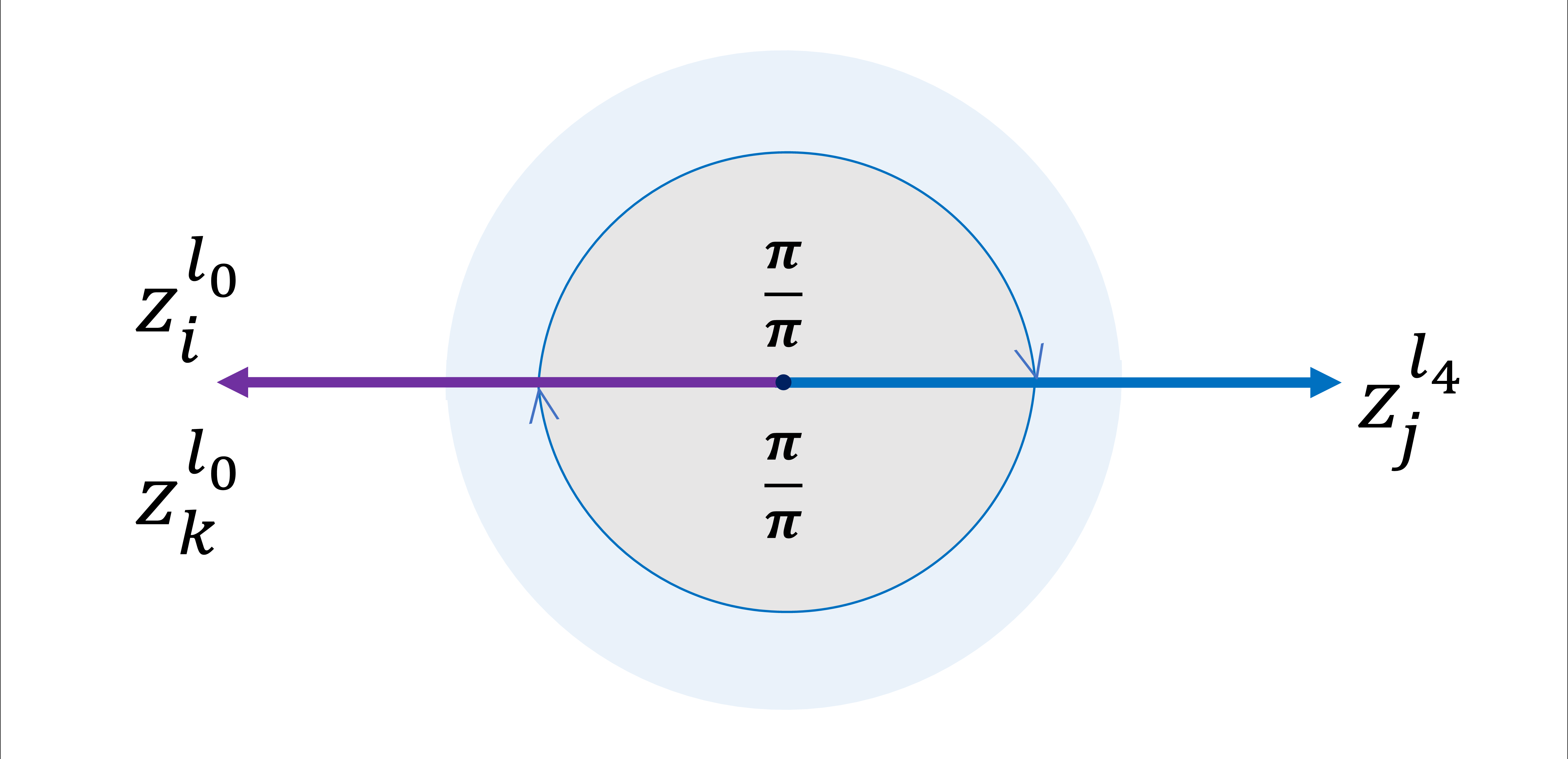}
			\label{ortho_dist5}}%
		\subfloat[]{\includegraphics[width=0.24\linewidth]{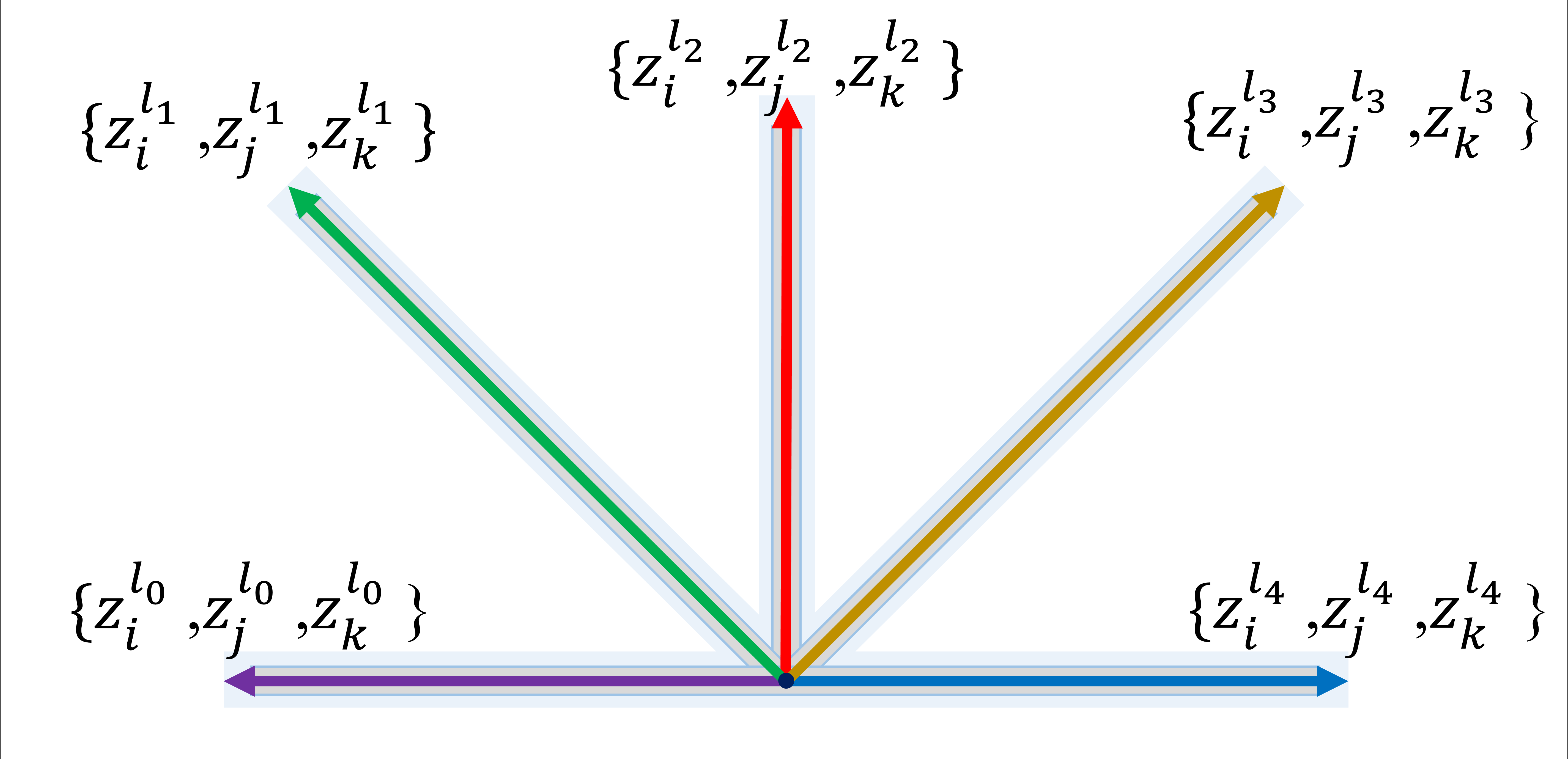}
			\label{ortho_dist6}}
		\caption{Illustration of ATD in $2D$ space: \protect\subref{ortho_dist1}. All possible proportional angles when there are five categories $\{l_0,l_1,l_2,l_3,l_4\}$, where ${\theta}_{z^{l_0}_i,z^{l_1}_i} = {\theta}_{z^{l_1}_i,z^{l_2}_i} = {\theta}_{z^{l_2}_i,z^{l_3}_i} = {\theta}_{z^{l_3}_i,z^{l_4}_i} = 45^{\circ}$, \protect\subref{ortho_dist2}. The distance of category $l_1$ to lower-bound ($l_0$) and upper-bound category ($l_4$) $= \{{\theta}_{z^{l_0}_i,z^{l_1}_j} = 45^{\circ}, {\theta}_{z^{l_1}_j, z^{l_4}_k} = 135^{\circ}\}$, \protect\subref{ortho_dist3}. The distance of category $l_2$ to lower-bound ($l_0$) and upper-bound category ($l_4$) $={\theta}_{z^{l_0}_i,z^{l_2}_j} = {\theta}_{z^{l_2}_j,z^{l_4}_k} = 90^{\circ}$, \protect\subref{ortho_dist4}. The distance of category $l_3$ to lower-bound ($l_0$) and upper-bound category ($l_4$) $= \{{\theta}_{z^{l_0}_i,z^{l_3}_j}= 135^{\circ}, {\theta}_{z^{l_3}_j, z^{l_4}_k} = 45^{\circ}\}$, \protect\subref{ortho_dist5}. The distance between lower-bound ($l_0$) and upper-bound category ($l_4$) $= \{{\theta}_{z^{l_0}_i,z^{l_4}_j} = 180^{\circ}, {\theta}_{z^{l_4}_j,z^{l_0}_i} = 180^{\circ}\}$ (a complete rotation), and \protect\subref{ortho_dist6}. The distance of inner-bound categories comprise of ${\theta}_{z^{l_0}_i,z^{l_0}_j,z^{l_0}_k} = {\theta}_{z^{l_1}_i,z^{l_1}_j,z^{l_1}_k} = {\theta}_{z^{l_2}_i,z^{l_2}_j,z^{l_2}_k} = {\theta}_{z^{l_3}_i,z^{l_3}_j,z^{l_3}_k} = {\theta}_{z^{l_4}_i,z^{l_4}_j,z^{l_4}_k} = 0^{\circ}$.}
		\label{fig:angular_distance}
	\end{figure*}
	
	For example, in manufacturing production, let us define $C=5$, where $l_0=\text{awful}$, $l_1=\text{low}$, $l_2=\text{medium}$, $l_3=\text{high}$, and $l_4=\text{excellent}$ quality. The possible number of angles in the angular space is $C-1$. The degree for each angle was $\frac{180^{\circ}}{C-1}$, as depicted in Fig. \ref{fig:angular_distance} \protect\subref{ortho_dist1}. Henceforth, we can define the $D_A$ between the categories based on this degree, that is, the angular distance between $z^{l_0}_i$ and $z^{l_0}_j$ is denoted as $D_A(z^{l_0}_i, z^{l_0}_j)=\frac{0}{\pi}=0$, similarly for $D_A(z^{l_0}_i, z^{l_1}_j)=\frac{{\pi}/4}{\pi}$, $D_A(z^{l_0}_i, z^{l_2}_j)=\frac{{\pi}/2}{\pi}$, $D_A(z^{l_0}_i, z^{l_4}_j)=\frac{\pi}{\pi}=1$, and so forth. Hence, to accurately map $X$ to a low-dimensional embedding space representation ($Z$) based on its label, while preserving its ordinal nature, we aim to project $X$ based on its categories ($l$) in the angular space. Unfortunately, when using pairwise comparison to determine the angular distance among the categories, the possible combination can be expressed as $C+\binom{C}{2}$. Consequently, as $C$ increases, the pairwise combination increases as well.
	
	To address this issue, we propose a triplet-wise representation to simply calculate the ordinal distance in the angular space. Here, there are three inputs: $z^{l_{r_i}}_i, z^{l_{r_j}}_j$, and $z^{l_{r_k}}_k$, where $i \neq j \neq k$. The ATD ($D_{AT}$) between them is expressed in Eq. \ref{eq:angular_triangle_dist}. First, because we aim to learn the order of categories for each sample, learning the position or distance of middle-bound $\{l_1,l_2,l_3\}$ categories to the lower-bound ($l_0$) and upper-bound ($l_4$) categories is sufficient. In this case, the possible triplet combinations are $D_{AT}(z^{l_0}_i, z^{l_1}_j, z^{l_4}_k)$, $D_{AT}(z^{l_0}_i, z^{l_2}_j, z^{l_4}_k)$, $D_{AT}(z^{l_0}_i, z^{l_3}_j, z^{l_4}_k)$, and $D_{AT}(z^{l_0}_i, z^{l_4}_j, z^{l_0}_k)$, as illustrated in Fig. \ref{fig:angular_distance} \protect\subref{ortho_dist2}, \protect\subref{ortho_dist3}, \protect\subref{ortho_dist4}, and \protect\subref{ortho_dist5}, respectively. Second, for the sake of mapping accuracy, we also learn the relation within its category, called inner-bound mapping, such as $D_{AT}(z^{l_0}_i, z^{l_0}_j, z^{l_0}_k)$, $D_{AT}(z^{l_1}_i, z^{l_1}_j, z^{l_1}_k)$, $D_{AT}(z^{l_2}_i, z^{l_2}_j, z^{l_2}_k)$, and so forth, as depicted in Fig. \ref{fig:angular_distance} \protect\subref{ortho_dist6}. Therefore, the total number of combinations to estimate the ordinal distance with triplet representation can be denoted as $C+C-1 = 2C-1$. 
	\begin{flalign}
		\label{eq:angular_triangle_dist}
		{D_{AT}}(z^{l_{r_i}}_i,z^{l_{r_j}}_j,z^{l_{r_k}}_k) &= &&\\\nonumber \frac{{\cos}^{-1}({S_C}(z^{l_{r_i}}_i,z^{l_{r_j}}_j))+{\cos}^{-1}({S_C}(z^{l_{r_j}}_j,z^{l_{r_k}}_k))}{\pi} &= &&\\\nonumber
		\frac{\theta_{z^{l_{r_i}}_i,z^{l_{r_j}}_j} + \theta_{z^{l_{r_j}}_j,z^{l_{r_k}}_k}}{\pi}&&
	\end{flalign}

	\begin{theorem}
		\label{theorem1}
		Suppose there are $C$ ordinal categories: $L = \{l_r |r=0,1,2,\dotsc,m,n,o,p,q,\dotsc,C-1\}$, where their order can be denoted as $l_m<l_n<l_o<l_p<l_q$. The ATD $D_{AT}(z^{l_{m}}_i, z^{l_{n}}_j, z^{l_{o}}_k)$ is a distance metric. Hence, it must satisfy the following properties. Non-negativity ($D_{AT}(z^{l_{m}}_i, z^{l_{n}}_j, z^{l_{o}}_k)\geq0$), identity ($D_{AT}(z^{l_{m}}_i, z^{l_{m}}_j, z^{l_{m}}_k)=0$), symmetry ($D_{AT}(z^{l_m}_i, z^{l_n}_j, z^{l_o}_k) = D_{AT}(z^{l_o}_i, z^{l_n}_j, z^{l_m}_k))$, and triangle inequality ($D_{AT}(z^{l_m}_i, z^{l_n}_j, z^{l_o}_k)+D_{AT}(z^{l_o}_i, z^{l_p}_j, z^{l_q}_k)\geq D_{AT}(z^{l_m}_i, z^{l_o}_j, z^{l_q}_k)$).
	\end{theorem}
	As a new distance metric, our ATD satisfies the basic distance metric properties as described in Theorem \ref{theorem1}. The proof of this is provided in Appendix \ref{ap:proof_dist_metric}. 
	
	\subsection{Network \& learning system}
\begin{figure}[t]
		\centering
		\includegraphics[width=0.65\linewidth]{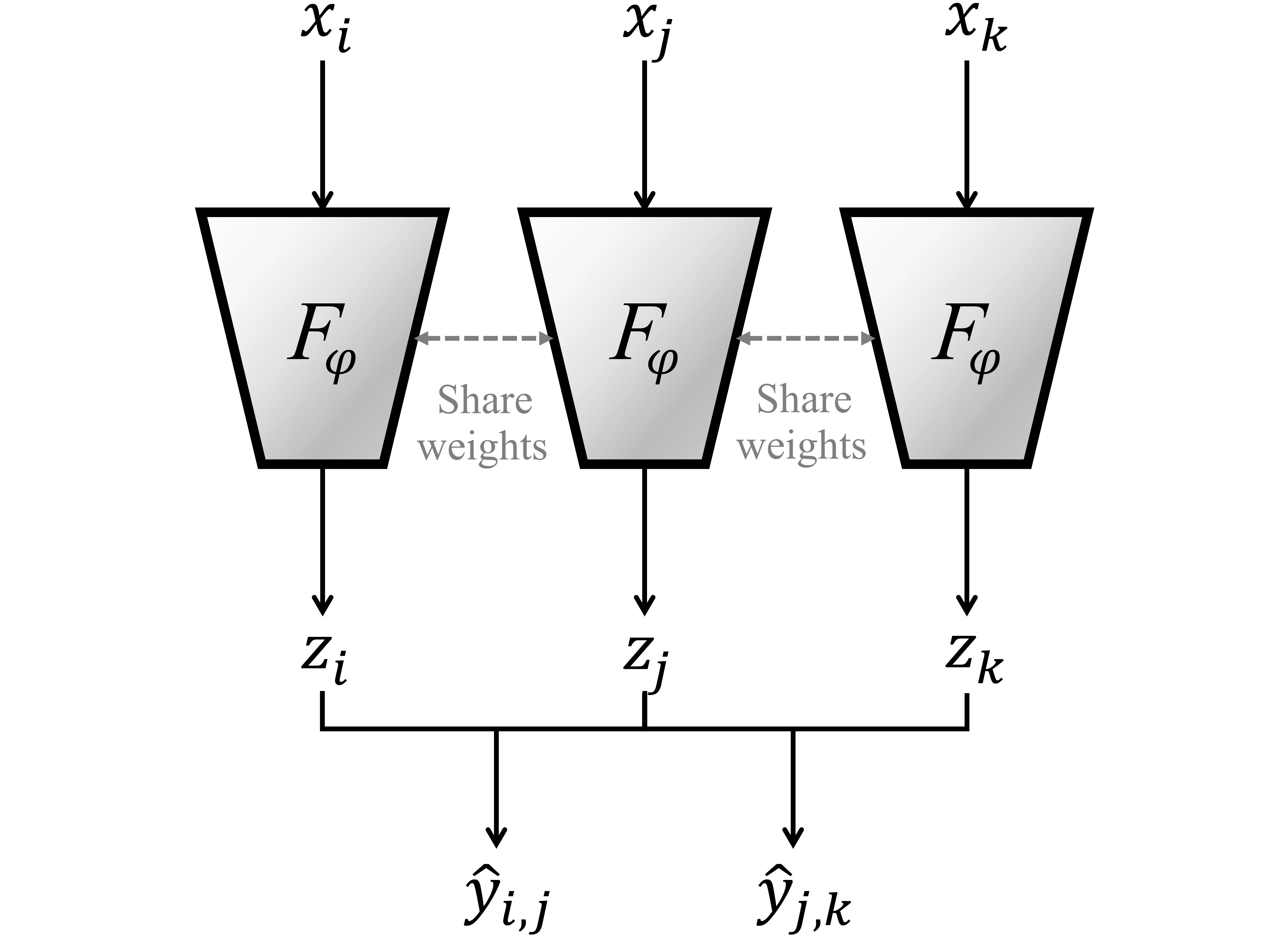}
		\caption{OTD with triplet input and duplet output to learn the sequence of ordinal data.}
		\label{fig:proposed_model}
	\end{figure}

	The network architecture of ordinal metric learning is shown in Fig. \ref{fig:proposed_model}. Unlike the existing DML models, it has three inputs and two outputs. The inputs represent the triplet representation ($\{x_i,x_j,x_k\}$) described in Section \ref{subsec:ortho_tri_dist}. The outputs are expected to be $y_{i,j}$ and $y_{j,k}$ corresponding to $D_A(x_i,x_j)$ and $D_A(x_j,x_k)$, respectively, as shown in Eq. \ref{eq:xlabel}. The estimated $\hat{y}_{i,j}$ and $\hat{y}_{j,k}$, which represent the estimated distances of $\widehat{D}_A(x_i,x_j)$ and $\widehat{D}_A(x_j,x_k)$, respectively (Eq. \ref{eq:ylabel}), are produced by $F_{\varphi}$. We employed two output representations rather than one because we intended to learn the triplet inputs’ sequence order and their mapping accuracy. As illustrated, there are three embedding networks ($F_{\varphi}$) in our framework. However, they are identical; therefore, they are physically only a single network because they share their weights. 
	\begin{align}
		\label{eq:xlabel}
		y_{i,j} = D_A(x_i,x_j),  \, y_{j,k} = D_A(x_j,x_k) \\
		\label{eq:ylabel}
		\hat{y}_{i,j} = \widehat{D}_A(x_i,x_j),  \, \hat{y}_{j,k} = \widehat{D}_A(x_j,x_k) \\
		\label{eq:loss_func}
		\mathcal{L_{\varphi}} = \frac{1}{T}\sum_{t=0}^{T-1}\Big({(y_{i,j}-{\hat{y}}_{i,j})}_t + {(y_{j,k}-{\hat{y}}_{j,k})}_t\Big) 
	\end{align}
	Algorithm \ref{algo:ordinal_metric_net} describes the learning mechanism of an OTD. We denote $\mathcal{X}$ as a set of triplet representations of $X$. $\mathcal{X}$ consists of tuples containing three values, $\{x_i,x_j,x_k\}$ and $T$ is the total number of triplet samples or tuples in $\mathcal{X}$. The angular distance between them ($\{y_{i,j},y_{j,k}\}$) is defined as label $\mathcal{Y}$. The network was trained using ordinal regression under the supervision of the label. Similar to a standard neural network, the model was trained using the stochastic gradient descent optimization algorithm, and the weights were updated using the backpropagation of the error algorithm to minimize Eq. \ref{eq:loss_func}. Given triplet inputs ($\{x_i,x_j,x_k\}$), the $F_{\varphi}$ projects them to latent representations, $\{z_i,z_j,z_k\}$, respectively. Their estimated labels ${\hat{y}}_{i,j}$ and ${\hat{y}}_{j,k}$, are obtained using Eq. \ref{eq:angular_dist}. In this study, we employed the mean square error (MSE) as a loss function to minimize the discrepancy error between the real and estimated labels, as formulated in Eq. \ref{eq:loss_func}. At the end of the iteration, the embedding model parameter $\varphi$ was updated based on the loss function value. These learning processes were repeated from $e=0$ until the predefined number of epochs ($Epoch - 1$), as shown in Algorithm \ref{algo:ordinal_metric_net} from lines 1 to 6. In practice, the network is trained with a predefined number of batches. The optimum embedding model ($F^*_{\varphi}$) was determined based on the best accuracy in the validation set.

			\begin{algorithm}
				\setstretch{0.85}
				\SetAlgoLined
				\caption{training procedure of OTD}
				\label{algo:ordinal_metric_net}
				\KwIn{$\mathcal{X}={\{x_i,x_j,x_k\}}_0,...,{\{x_i,x_j,x_k\}}_{T-1}$, $\mathcal{Y} = {\{y_{ij},y_{jk}\}}_0,...,{\{y_{ij},y_{jk}\}}_{T-1}$} 
				\KwOut{$F^*_{{\varphi}}$}
				\For{$e \gets 0$ \KwTo $Epoch - 1$}{
					$z_i = F_{{\varphi}_e}(x_i), z_j = F_{{\varphi}_e}(x_j), z_k = F_{{\varphi}_e}(x_k)$ \DontPrintSemicolon \Comment*[r]{extract embedding reprsentation}
					$\hat{y}_{ij} = D_A(z_i,z_j), \hat{y}_{jk} = D_A(z_j,z_k)$ \DontPrintSemicolon \Comment*[r]{$D_A$ is expressed in Eq. \ref{eq:angular_dist}}
					$\mathcal{L}_{{\varphi}_e} =\text{MSE}(y_{i,j},{\hat{y}}_{i,j}) + \text{MSE}(y_{j,k},{\hat{y}}_{j,k}$) 
					\DontPrintSemicolon \Comment*[r]{calculate loss (Eq. \ref{eq:loss_func})}
					${\varphi}_e \gets {\varphi}_e - \eta \nabla_{{\varphi}_e}{\mathcal{L}}_{{\varphi}_e}$ \DontPrintSemicolon \Comment*[r]{update model parameter}
				}
			\end{algorithm}

	\section{Experiment}
	\label{sec:experiment}
	We evaluate our ATD, with the OTD, using four public datasets with an ordinal nature, such as breast ultrasonic images (Busi) \cite{ALDHABYANI2020104863}, counting finger images (Finger)\footnote{\url{https://www.kaggle.com/datasets/koryakinp/fingers}}, and face age recognition (FG-Net \cite{ranking2016PAMI} and Adience \cite{7490057}). The effectiveness of our proposed method ($O$-Net) is mainly assessed in terms of both the mapping accuracy and semantic correctness of the embedding space representation. We compare the performance with state-of-the-art DML models, such as the Siamese network with contrastive loss ($S$-Net \cite{Chicco2021}), triplet network with triplet loss ($T$-Net \cite{7298682}), quadruplet network with quadruplet loss ($Q$-Net \cite{DBLP:conf/cvpr/ChenCZH17}), and a DML model with N-pair loss ($N$-Net \cite{NIPS2016_6b180037}). For fairness, we employed the same embedding network model for each model, as shown in Table \ref{tab:net_architecture} (Appendix \ref{ap:detailed_architect}). The overall architecture and parameter comparison for each model are presented in Table \ref{tab:model_archict_comp} (Appendix \ref{ap:detailed_architect}).
	\begin{figure*}
		\centering
		\subfloat[Busi]{\includegraphics[width=0.19\linewidth]{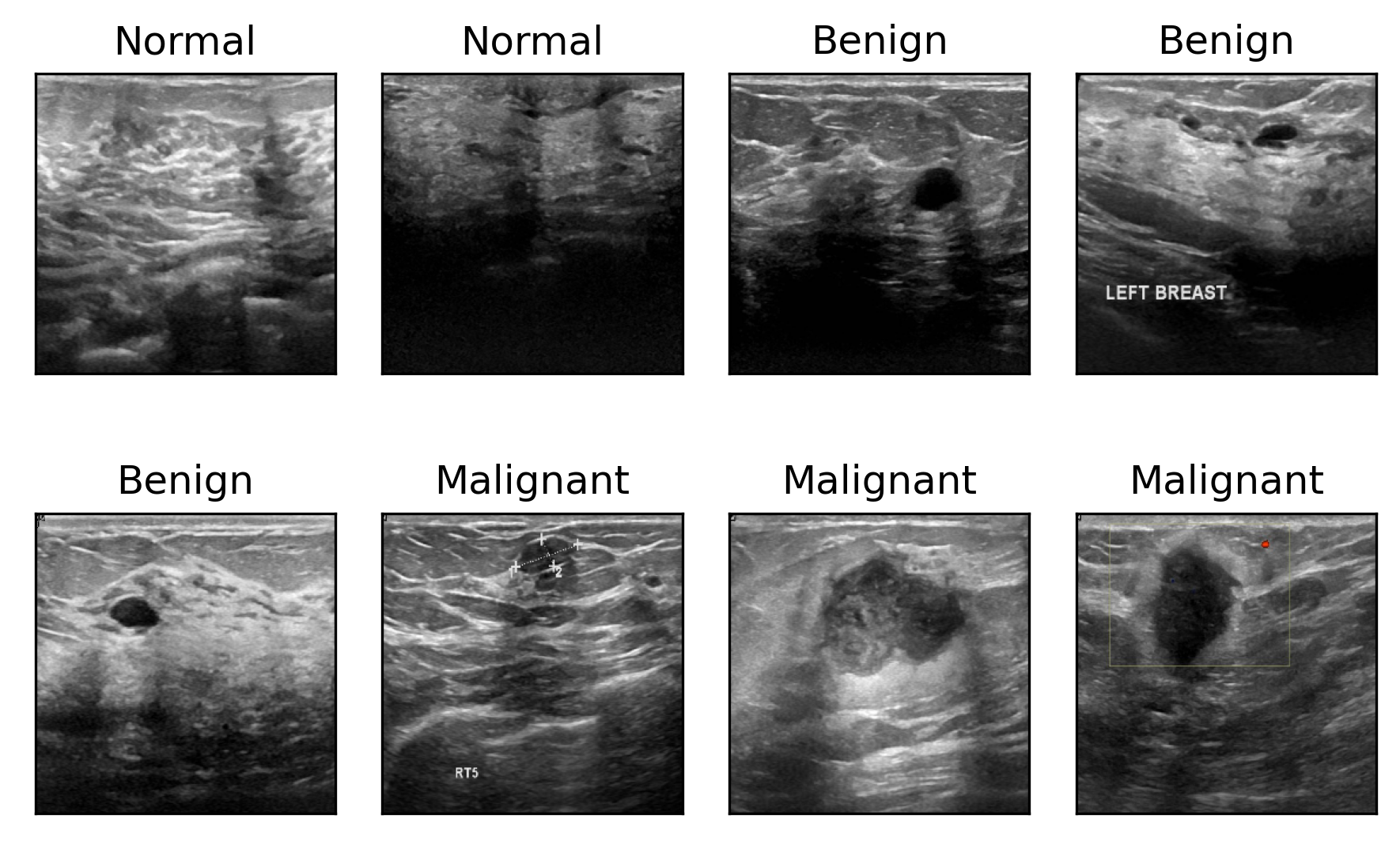}
			\label{busi_data}}\hspace{3pt}%
		\subfloat[Finger]{\includegraphics[width=0.19\linewidth]{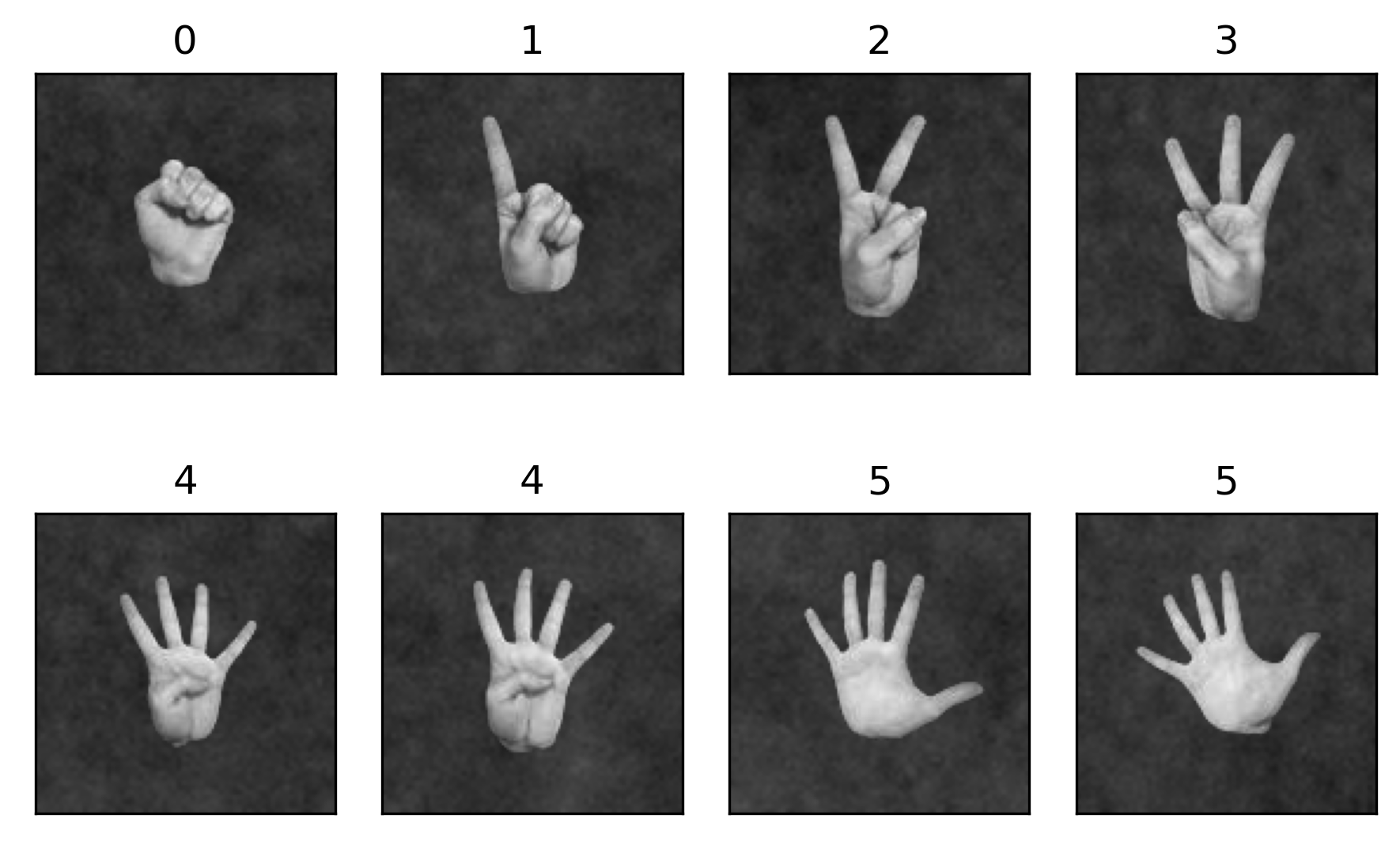}
			\label{finger_data}}\hspace{3pt}%
		\subfloat[FG-Net]{\includegraphics[width=0.19\linewidth]{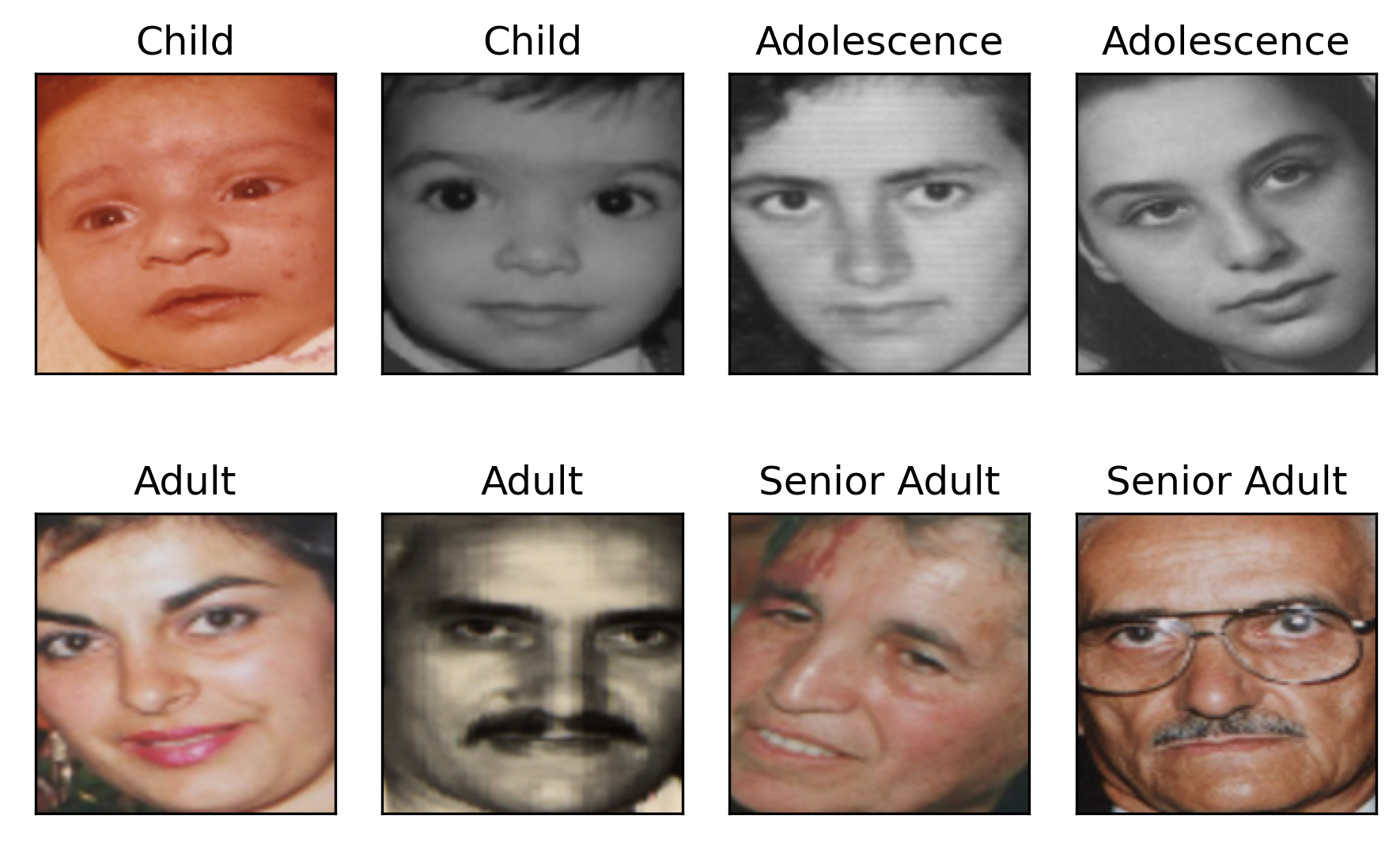}
			\label{FG-Net_data}}\hspace{3pt}%
		\subfloat[Adience]{\includegraphics[width=0.19\linewidth]{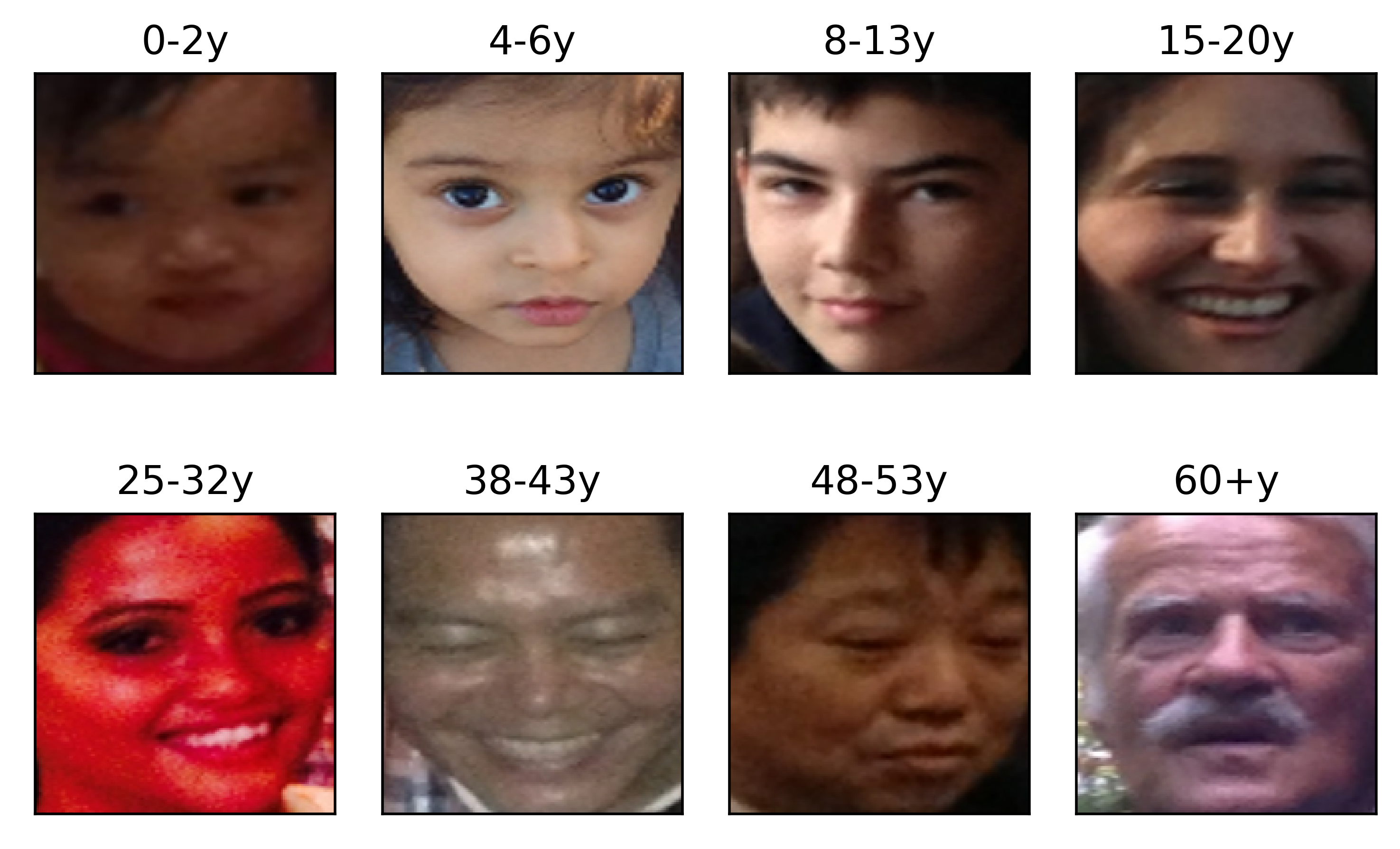}
			\label{adience_data}}
		\caption{Dataset samples with their ordinal labels (category).}
		\label{fig:dataset_overview}
	\end{figure*}
	\subsection{Data \& setting}
	The Busi dataset is divided into three categories: normal, benign, and malignant images, as shown in Fig. \ref{fig:dataset_overview} \protect\subref{busi_data}. The dataset consists of 780 images, with an average image size of $500\times500$ pixels. For simplicity, we normalized the image size to $256\times256$ with a range of $[0,1]$. For data augmentation, we also doubled the images by flipping them horizontally. The Finger dataset created to automatically count fingers in both the left and right hands has six categories: $0,1,2,3,4,$,and $5$ (Fig. \ref{fig:dataset_overview} \protect\subref{finger_data}). The data dimensions were $500\times500$, with 18,000 and 3,600 images as the training and testing sets, respectively. We also normalized the size to $128\times128$ pixels with an interval $[0,1]$. The FG-Net dataset is an aging database consisting of 1,002 images of various face age groups. Originally, the age range of the patients was 0–69 years. We converted them into the following groups: child (0-12 years), adolescents (13-18 years), adult (19-59 years), and senior adults (60 years and above), as depicted in Fig. \ref{fig:dataset_overview} \protect\subref{FG-Net_data}. For data augmentation, we doubled the image using a horizontal flip. The Adience dataset is a face age recognition dataset obtained from the \textit{Flickr} album. The data size 26,580 with eight categories: 0-2, 4-6, 8-13, 15-20, 25-32, 38-43, 48-53, and 60+ years old, as shown in Fig. \ref{fig:dataset_overview} \protect\subref{adience_data}. Moreover, we conducted a pre-processing technique for both FG-Net and Adience by cropping the image centered on the face with a size of $150\times150$ pixels and normalizing it by subtracting its mean and dividing it by its standard deviation. In this study, we randomly divided the data into five blocks (except the Adience dataset already divided into five blocks). The data were randomly split into training and testing sets of 80\% and 20\% for each block, respectively. Moreover, we randomly subtracted 20\% of the training set for the validation set to obtain the optimum model during training. Subsequently, due to the stochastic nature of the deep learning-based model, we conducted ten repetitive experiments for each block to obtain the average and standard deviation of model performance. Furthermore, we used FaceNet \cite{7298682} backbone in the embedding network of the FG-Net and Adience dataset.
	\begin{figure*}
		\centering
		\begin{tikzpicture}[scale=0.4,font=\large]
			\begin{axis}[
				title=\LARGE{Busi},
				width  = 0.8*\textwidth,
				height = 9cm,
				%major y tick style = transparent,
				xbar=2*\pgflinewidth,
				bar width=6pt,
				xlabel = {\Large{Accuracy}},
				symbolic y coords={$O$-Net,$S$-Net,$T$-Net,$Q$-Net,$N$-Net$_1$,$N$-Net$_2$},
				ytick = data,scaled x ticks = false, 
				enlarge y limits=0.1,
				nodes near coords={\pgfmathprintnumber\pgfplotspointmeta},
				nodes near coords align={horizontal},
				every node near coord/.append style={font=\small,xshift=14pt,/pgf/number format/.cd,precision=3},
				xmax = 1.06,
				legend image code/.code={%
					\draw[#1, draw=none] (0cm,-0.1cm) rectangle (0.65cm,0.22cm);
				},  
				legend style={at={(0.5,-0.17)},
					anchor=north,legend columns=-1},
				every axis plot/.append style={fill opacity=1.0},
				every tick label/.append style={font=\Large}
				]
				\addplot[xbar, black,fill=red!50,postaction={pattern=north east lines},error bars/.cd,
				x dir=both,x explicit] coordinates {
					(0.931201923,{$O$-Net}) += (0.02735982,0) -=(0.02735982,0)
					(0.886538462,{$S$-Net}) += (0.03690188,0) -=(0.03690188,0) 
					(0.509214744,{$T$-Net}) += (0.038530569,0) -=(0.038530569,0)
					(0.759935897,{$Q$-Net}) += (0.026957184,0) -=(0.026957184,0) 
					(0.54150641,{$N$-Net$_1$}) += (0.026957184,0) -=(0.026957184,0) 
					(0.88525641,{$N$-Net$_2$}) += (0.036097833,0) -=(0.036097833,0)
				};
				\addlegendentry{$k$=2}
				\addplot[xbar,black,fill=yellow,postaction={pattern=horizontal lines},error bars/.cd,
				x dir=both,x explicit] coordinates {
					(0.930011276,{$O$-Net}) += (0.0218294121,0) -=(0.0218294121,0)
					(0.87786859,{$S$-Net}) += (0.040906524,0) -=(0.040906524,0)
					(0.497088675,{$T$-Net}) += (0.026699443,0) -=(0.026699443,0)
					(0.723114316,{$Q$-Net}) += (0.031795072,0) -=(0.031795072,0)
					(0.502857906,{$N$-Net$_1$}) += (0.021813241,0) -=(0.021813241,0)
					(0.885229701,{$N$-Net$_2$}) += (0.031257494,0) -=(0.031257494,0)
				};
				\addlegendentry{$k$=6}
				\addplot[xbar,black,fill=green!50,postaction={pattern=dots},error bars/.cd,
				x dir=both,x explicit] coordinates {
					(0.948285256,{$O$-Net}) += (0.0258163104,0) -=(0.0258163104,0)
					(0.889871795,{$S$-Net}) += (0.037315311,0) -=(0.037315311,0)
					(0.487163462,{$T$-Net}) += (0.019761322,0) -=(0.019761322,0)
					(0.699871795,{$Q$-Net}) += (0.035795072,0) -=(0.035795072,0)
					(0.491378205,{$N$-Net$_1$}) += (0.02176812,0) -=(0.02176812,0)
					(0.885080128,{$N$-Net$_2$}) += (0.03234035,0) -=(0.03234035,0)
					
				};
				\addlegendentry{$k$=10}
				\addplot[xbar,black,fill=blue!50,postaction={pattern=crosshatch},error bars/.cd,
				x dir=both,x explicit] coordinates {
					(0.93709478,{$O$-Net}) += (0.023843447,0) -=(0.023843447,0)
					(0.877852564,{$S$-Net}) += (0.036387199,0) -=(0.036387199,0)
					(0.481467491,{$T$-Net}) += (0.01913583,0) -=(0.01913583,0)
					(0.682394689,{$Q$-Net}) += (0.036795072,0) -=(0.036795072,0)
					(0.482989927,{$N$-Net$_1$}) += (0.020039877,0) -=(0.020039877,0)
					(0.884672619,{$N$-Net$_2$}) += (0.032465968,0) -=(0.032465968,0)
				};
				\addlegendentry{$k$=14}
			\end{axis}
		\end{tikzpicture}
		\begin{tikzpicture}[scale=0.4,font=\large]
			\begin{axis}[
				title=\LARGE{Finger},
				width  = 0.8*\textwidth,
				height = 9cm,
				xmin = 0.5,
				%major y tick style = transparent,
				xbar=2*\pgflinewidth,
				bar width=6pt,
				xlabel = {\Large{Accuracy}},
				symbolic y coords={$O$-Net,$S$-Net,$T$-Net,$Q$-Net,$N$-Net$_1$,$N$-Net$_2$},
				ytick = data,scaled x ticks = false, 
				nodes near coords={\pgfmathprintnumber\pgfplotspointmeta},
				nodes near coords align={horizontal},
				every node near coord/.append style={font=\small,xshift=4pt,/pgf/number format/.cd,precision=3},
				xmax = 1.04,
				legend image code/.code={%
					\draw[#1, draw=none] (0cm,-0.1cm) rectangle (0.65cm,0.22cm);
				},  
				legend style={at={(0.5,-0.17)},
					anchor=north,legend columns=-1},
				every axis plot/.append style={fill opacity=1.0},
				every tick label/.append style={font=\Large}
				]
				\addplot[xbar, black,fill=red!50,postaction={pattern=north east lines},error bars/.cd,
				x dir=both,x explicit] coordinates {
					(1.0,{$O$-Net}) += (0.0,0) -=(0.0,0)
					(1.0,{$S$-Net}) += (0.0,0) -=(0.0,0) 
					(1.0,{$T$-Net}) += (0.0,0) -=(0.0,0)
					(1.0,{$Q$-Net}) += (0.0,0) -=(0.0,0) 
					(1.0,{$N$-Net$_1$}) += (0.0,0) -=(0.0,0) 
					(1.0,{$N$-Net$_2$}) += (0.0,0) -=(0.0,0)
				};
				\addlegendentry{$k$=10}
				\addplot[xbar,black,fill=yellow,postaction={pattern=horizontal lines},error bars/.cd,
				x dir=both,x explicit] coordinates {
					(1.0,{$O$-Net}) += (0.0,0) -=(0.0,0)
					(1.0,{$S$-Net}) += (0.0,0) -=(0.0,0) 
					(1.0,{$T$-Net}) += (0.0,0) -=(0.0,0)
					(1.0,{$Q$-Net}) += (0.0,0) -=(0.0,0) 
					(1.0,{$N$-Net$_1$}) += (0.0,0) -=(0.0,0) 
					(1.0,{$N$-Net$_2$}) += (0.0,0) -=(0.0,0)
				};
				\addlegendentry{$k$=40}
				\addplot[xbar,black,fill=green!50,postaction={pattern=dots},error bars/.cd,
				x dir=both,x explicit] coordinates {
					(1.0,{$O$-Net}) += (0.0,0) -=(0.0,0)
					(1.0,{$S$-Net}) += (0.0,0) -=(0.0,0) 
					(1.0,{$T$-Net}) += (0.0,0) -=(0.0,0)
					(1.0,{$Q$-Net}) += (0.0,0) -=(0.0,0) 
					(1.0,{$N$-Net$_1$}) += (0.0,0) -=(0.0,0) 
					(1.0,{$N$-Net$_2$}) += (0.0,0) -=(0.0,0)
					
				};
				\addlegendentry{$k$=70}
				\addplot[xbar,black,fill=blue!50,postaction={pattern=crosshatch},error bars/.cd,
				x dir=both,x explicit] coordinates {
					(1.0,{$O$-Net}) += (0.0,0) -=(0.0,0)
					(1.0,{$S$-Net}) += (0.0,0) -=(0.0,0) 
					(1.0,{$T$-Net}) += (0.0,0) -=(0.0,0)
					(1.0,{$Q$-Net}) += (0.0,0) -=(0.0,0) 
					(1.0,{$N$-Net$_1$}) += (0.0,0) -=(0.0,0) 
					(1.0,{$N$-Net$_2$}) += (0.0,0) -=(0.0,0)
				};
				\addlegendentry{$k$=100}
			\end{axis}
		\end{tikzpicture}
		\begin{tikzpicture}[scale=0.4,font=\large]
			\begin{axis}[
				title=\LARGE{FG-Net},
				width  = 0.8*\textwidth,
				height = 9cm,
				%major y tick style = transparent,
				xbar=2*\pgflinewidth,
				bar width=6pt,
				xlabel = {\Large{Accuracy}},
				symbolic y coords={$O$-Net,$S$-Net,$T$-Net,$Q$-Net,$N$-Net$_1$,$N$-Net$_2$},
				ytick = data,scaled x ticks = false, 
				enlarge y limits=0.1,
				nodes near coords={\pgfmathprintnumber\pgfplotspointmeta},
				nodes near coords align={horizontal},
				every node near coord/.append style={font=\small,xshift=17pt,/pgf/number format/.cd,precision=3},
				xmax = 1.01,
				legend image code/.code={%
					\draw[#1, draw=none] (0cm,-0.1cm) rectangle (0.65cm,0.22cm);
				},  
				legend style={at={(0.5,-0.17)},
					anchor=north,legend columns=-1},
				every axis plot/.append style={fill opacity=1.0},
				every tick label/.append style={font=\Large}
				]
				\addplot[xbar, black,fill=red!50,postaction={pattern=north east lines},error bars/.cd,
				x dir=both,x explicit] coordinates {
					(0.667425252,{$O$-Net}) += (0.044596619,0) -=(0.044596619,0)
					(0.599752889,{$S$-Net}) += (0.050501409,0) -=(0.050501409,0) 
					(0.399404351,{$T$-Net}) += (0.055260325,0) -=(0.055260325,0)
					(0.728025539,{$Q$-Net}) += (0.04930785,0) -=(0.04930785,0) 
					(0.566226967,{$N$-Net$_1$}) += (0.054729256,0) -=(0.054729256,0) 
					(0.566669493,{$N$-Net$_2$}) += (0.045649642,0) -=(0.045649642,0)
				};
				\addlegendentry{$k$=2}
				\addplot[xbar,black,fill=yellow,postaction={pattern=horizontal lines},error bars/.cd,
				x dir=both,x explicit] coordinates {
					(0.600939574,{$O$-Net}) += (0.055323452,0) -=(0.055323452,0)
					(0.530763523,{$S$-Net}) += (0.058957876,0) -=(0.058957876,0)
					(0.334210575,{$T$-Net}) += (0.059078064,0) -=(0.059078064,0)
					(0.618360133,{$Q$-Net}) += (0.056473714,0) -=(0.056473714,0)
					(0.542306115,{$N$-Net$_1$}) += (0.054870471,0) -=(0.054870471,0)
					(0.54478102,{$N$-Net$_2$}) += (0.047315436,0) -=(0.047315436,0)
				};
				\addlegendentry{$k$=6}
				\addplot[xbar,black,fill=green!50,postaction={pattern=dots},error bars/.cd,
				x dir=both,x explicit] coordinates {
					(0.570497893,{$O$-Net}) += (0.061429384,0) -=(0.061429384,0)
					(0.501730804,{$S$-Net}) += (0.065040604,0) -=(0.065040604,0)
					(0.312862696,{$T$-Net}) += (0.060263249,0) -=(0.060263249,0)
					(0.7399877,{$Q$-Net}) += (0.055912049,0) -=(0.055912049,0)
					(0.533587019,{$N$-Net$_1$}) += (0.054834205,0) -=(0.054834205,0)
					(0.533340121,{$N$-Net$_2$}) += (0.048256357,0) -=(0.048256357,0)
					
				};
				\addlegendentry{$k$=10}
				\addplot[xbar,black,fill=blue!50,postaction={pattern=crosshatch},error bars/.cd,
				x dir=both,x explicit] coordinates {
					(0.55058615,{$O$-Net}) += (0.06493712,0) -=(0.06493712,0)
					(0.478121138,{$S$-Net}) += (0.058175733,0) -=(0.058175733,0)
					(0.301459445,{$T$-Net}) += (0.059854021,0) -=(0.059854021,0)
					(0.549089266,{$Q$-Net}) += (0.058732456,0) -=(0.058732456,0)
					(0.526885082,{$N$-Net$_1$}) += (0.054682403,0) -=(0.054682403,0)
					(0.525499654,{$N$-Net$_2$}) += (0.04887684,0) -=(0.04887684,0)
				};
				\addlegendentry{$k$=14}
			\end{axis}
		\end{tikzpicture}
		\begin{tikzpicture}[scale=0.4,font=\large]
			\begin{axis}[
				title=\LARGE{Adience},
				width  = 0.8*\textwidth,
				height = 9cm,
				%major y tick style = transparent,
				xbar=2*\pgflinewidth,
				bar width=6pt,
				xlabel = {\Large{Accuracy}},
				symbolic y coords={$O$-Net,$S$-Net,$T$-Net,$Q$-Net,$N$-Net$_1$,$N$-Net$_2$},
				ytick = data,scaled x ticks = false, 
				enlarge y limits=0.1,
				nodes near coords={\pgfmathprintnumber\pgfplotspointmeta},
				nodes near coords align={horizontal},
				every node near coord/.append style={font=\small,xshift=14pt,/pgf/number format/.cd,precision=3},
				xmax = 1.06,
				legend image code/.code={%
					\draw[#1, draw=none] (0cm,-0.1cm) rectangle (0.65cm,0.22cm);
				},  
				legend style={at={(0.5,-0.17)},
					anchor=north,legend columns=-1},
				every axis plot/.append style={fill opacity=1.0},
				every tick label/.append style={font=\Large}
				]
				\addplot[xbar, black,fill=red!50,postaction={pattern=north east lines},error bars/.cd,
				x dir=both,x explicit] coordinates {
					(0.931201923,{$O$-Net}) += (0.02735982,0) -=(0.02735982,0)
					(0.886538462,{$S$-Net}) += (0.03690188,0) -=(0.03690188,0) 
					(0.509214744,{$T$-Net}) += (0.038530569,0) -=(0.038530569,0)
					(0.759935897,{$Q$-Net}) += (0.026957184,0) -=(0.026957184,0) 
					(0.54150641,{$N$-Net$_1$}) += (0.026957184,0) -=(0.026957184,0) 
					(0.88525641,{$N$-Net$_2$}) += (0.036097833,0) -=(0.036097833,0)
				};
				\addlegendentry{$k$=10}
				\addplot[xbar,black,fill=yellow,postaction={pattern=horizontal lines},error bars/.cd,
				x dir=both,x explicit] coordinates {
					(0.930011276,{$O$-Net}) += (0.0218294121,0) -=(0.0218294121,0)
					(0.87786859,{$S$-Net}) += (0.040906524,0) -=(0.040906524,0)
					(0.497088675,{$T$-Net}) += (0.026699443,0) -=(0.026699443,0)
					(0.723114316,{$Q$-Net}) += (0.031795072,0) -=(0.031795072,0)
					(0.502857906,{$N$-Net$_1$}) += (0.021813241,0) -=(0.021813241,0)
					(0.885229701,{$N$-Net$_2$}) += (0.031257494,0) -=(0.031257494,0)
				};
				\addlegendentry{$k$=40}
				\addplot[xbar,black,fill=green!50,postaction={pattern=dots},error bars/.cd,
				x dir=both,x explicit] coordinates {
					(0.948285256,{$O$-Net}) += (0.0258163104,0) -=(0.0258163104,0)
					(0.889871795,{$S$-Net}) += (0.037315311,0) -=(0.037315311,0)
					(0.487163462,{$T$-Net}) += (0.019761322,0) -=(0.019761322,0)
					(0.699871795,{$Q$-Net}) += (0.035795072,0) -=(0.035795072,0)
					(0.491378205,{$N$-Net$_1$}) += (0.02176812,0) -=(0.02176812,0)
					(0.885080128,{$N$-Net$_2$}) += (0.03234035,0) -=(0.03234035,0)
					
				};
				\addlegendentry{$k$=70}
				\addplot[xbar,black,fill=blue!50,postaction={pattern=crosshatch},error bars/.cd,
				x dir=both,x explicit] coordinates {
					(0.93709478,{$O$-Net}) += (0.023843447,0) -=(0.023843447,0)
					(0.877852564,{$S$-Net}) += (0.036387199,0) -=(0.036387199,0)
					(0.481467491,{$T$-Net}) += (0.01913583,0) -=(0.01913583,0)
					(0.682394689,{$Q$-Net}) += (0.036795072,0) -=(0.036795072,0)
					(0.482989927,{$N$-Net$_1$}) += (0.020039877,0) -=(0.020039877,0)
					(0.884672619,{$N$-Net$_2$}) += (0.032465968,0) -=(0.032465968,0)
				};
				\addlegendentry{$k$=100}
			\end{axis}
		\end{tikzpicture}
		\caption{The effect of varying $K$-nearest neighbor on image search accuracy.}\label{fig:comparison_acc_knn}
	\end{figure*}
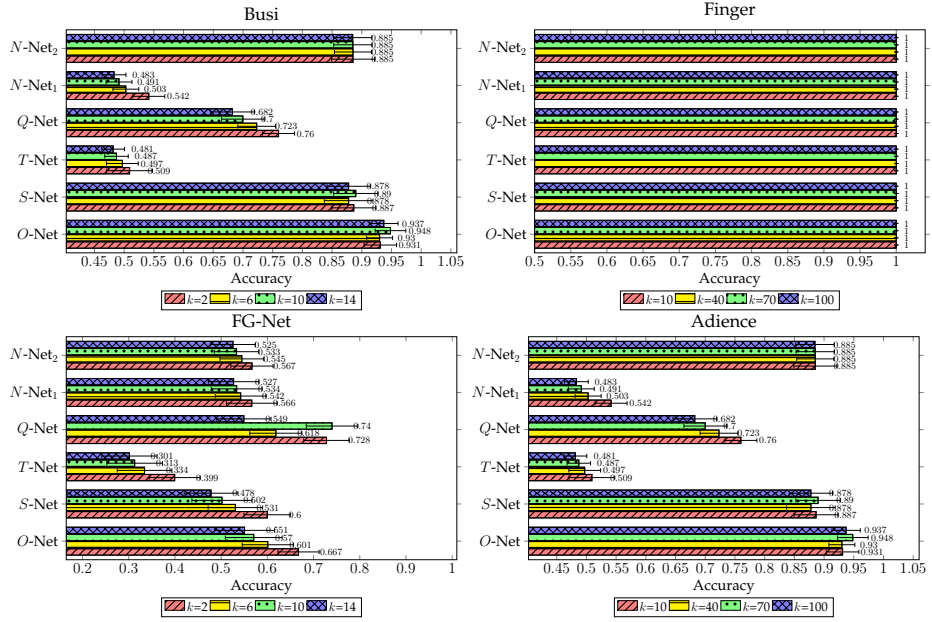
	\begin{equation}
		\label{eq:accuracy_knn}
		accuracy = \frac{\sum_{i=0}^{N-1} \sum_{j=0}^{K-1} a_i^j}{N \times K}, \; a_i^j =
		\begin{cases}
			1, & {\hat{l}}_i^j = l_i \\
			0, & \text{otherwise} \\
		\end{cases} 
	\end{equation}
	\subsection{Embedding accuracy}
	\label{subsec:embedding_acc}
	To assess the accuracy of the embedding space representation, we used the $K$-nearest neighbor algorithm. The $K$ number of similar images was obtained using a query image as the input based on the cosine similarity. We used Eq. \ref{eq:accuracy_knn} to measure the mapping accuracy of each image in the embedding space based on its label. The $a^j_i$ corresponds to the correctness between the label of the query image (${l}_i$) and the neighbor image label (${\hat{l}}_j$); where its value is 1, when the label is the same; otherwise, 0. The $K$ nearest neighbor performance of each model is shown in Fig. \ref{fig:comparison_acc_knn}. In the Busi, FG-Net, and Adience datasets, the results reveal that, generally, our proposed model outperforms existing DML models. The $S$-Net is trained based on a binary-like classification, where it is devised to distinguish between the inner and outer categories. Therefore, the performance may be less accurate than that of our $O$-Net, which is explicitly trained to simultaneously distinguish the inner, outer, and ordinal relationships among the categories. Furthermore, defining the anchor, positive, and negative samples in $T$-Net is crucial. The distance between the anchor and the positive samples should be shorter than that of the negative sample. Thus, its performance on the Busi dataset was not significantly accurate. Note that the samples of each category (normal, benign, and malignant) in the Busi dataset have similar features (Fig. \ref{fig:dataset_overview} \protect\subref{busi_data}), making them difficult to distinguish with a triplet loss. Meanwhile, in the FG-Net and Adience datasets, $T$-Net has better accuracy than the Busi dataset because it uses a pre-trained network (FaceNet) as a backbone in the embedding network. $Q$-Net outperformed $O$-Net when the $K=10$ in the Adience dataset. However, with increasing $K$, the accuracy was competitive. Because the number of batches significantly influences the $N$-Net accuracy, we employed two types of $N$-Net, namely, $N$-Net$_1$ and $N$-Net$_2$. The difference between $N$-Net$_1$ and $N$-Net$_2$ is the number of batches used during the training. $N$-Net$_1$ used the same number of batches as the other networks. For comparison, $N$-Net$_2$ was set to use a huge batch size (1,000). As shown in Fig. \ref{fig:comparison_acc_knn}, the higher the number of batches in $N$-Net, the higher its accuracy. Moreover, a fascinating finding in the Finger dataset is that all models have perfect accuracy because the separability of this data is very high, as shown in Fig. \ref{fig:dataset_overview} \protect\subref{finger_data}.
	
	Additionally, we employed a support vector machine with a standard radial basis function to assess latent separability in the embedding space representation \cite{KAMAL2022108562,8438540}. For dimensionality reduction, the embedding space representation should maintain a latent code separability similar to the original space. $F_{\varphi}$ projects $X$ for both training and testing sets into the latent variable representation $Z$. The training set of latent code trains the SVM model, and henceforth, the SVM model is tested by the latent code of the testing set. The performance of the latent code separability is shown in Table \ref{tab:classification_svm}. In line with the $K$-nearest neighbor algorithm, our proposed method outperformed existing DML models, such as $S$-Net, $T$-Net, $Q$-Net, $N$-Net$_1$, and $N$-Net$_2$. Meanwhile, all DML model performances achieved perfect accuracy in the Finger dataset, which is notable for being highly separable. Note that the SVM accuracy performance may contradict the accuracy of $K$-NN. This can occur because the total number of samples in the $K$-NN and SVM is $N\times K$ and $N$, respectively.
	\begin{table}
		\centering
		\caption{Evaluating the embedding space separability with an SVM classifier.}
		\label{tab:classification_svm}
		\resizebox{1\linewidth}{!}{
			\centering
			\begin{tabular}{l l l l l}		
				\toprule
				\multirow{2}{*}{Model} & \multicolumn{4}{c}{Accuracy (mean $\pm$ std.)}\\
				\cmidrule{2-5}
				&	\qquad Busi 							&	\qquad Finger							&	\qquad FG-Net					&	\qquad Adience			\\
				\midrule
				$O$-Net				&	\qquad $\textbf{0.950}\pm0.026$			&\qquad 	$\textbf{1.000}\pm0.000$		&\qquad 	$\textbf{0.961}\pm0.010$	&\qquad 	$\textbf{0.578}\pm0.153$	\\
				$S$-Net				&	\qquad $0.903\pm0.040$					&\qquad 	$\textbf{1.000}\pm0.000$		&\qquad 	$0.957\pm0.015$				&\qquad 	$0.547\pm0.161$	\\
				$T$-Net				&	\qquad $0.564\pm0.023$					&\qquad 	$\textbf{1.000}\pm0.000$		&\qquad 	$0.864\pm0.029$				&\qquad 	$\textbf{0.578}\pm0.178$	\\
				$Q$-Net				&	\qquad $0.879\pm0.027$					&\qquad 	$\textbf{1.000}\pm0.000$		&\qquad 	$0.905\pm0.016$				&\qquad 	$0.545\pm0.051$	\\				
				$N$-Net$_1$				&	\qquad $0.558\pm0.009$					&\qquad 	$\textbf{1.000}\pm0.000$		&\qquad 	$0.961\pm0.026$				&\qquad 	$0.570\pm0.159$	\\
				$N$-Net$_2$				&	\qquad $0.923\pm0.024$					&\qquad 	$\textbf{1.000}\pm0.000$		&\qquad 	$0.919\pm0.008$				&\qquad 	$0.458\pm0.157$	\\
				\bottomrule
			\end{tabular}
		}
	\end{table}
	
	\begin{figure*}
		\centering
		\subfloat[$O$-Net]{\includegraphics[width=0.18\linewidth]{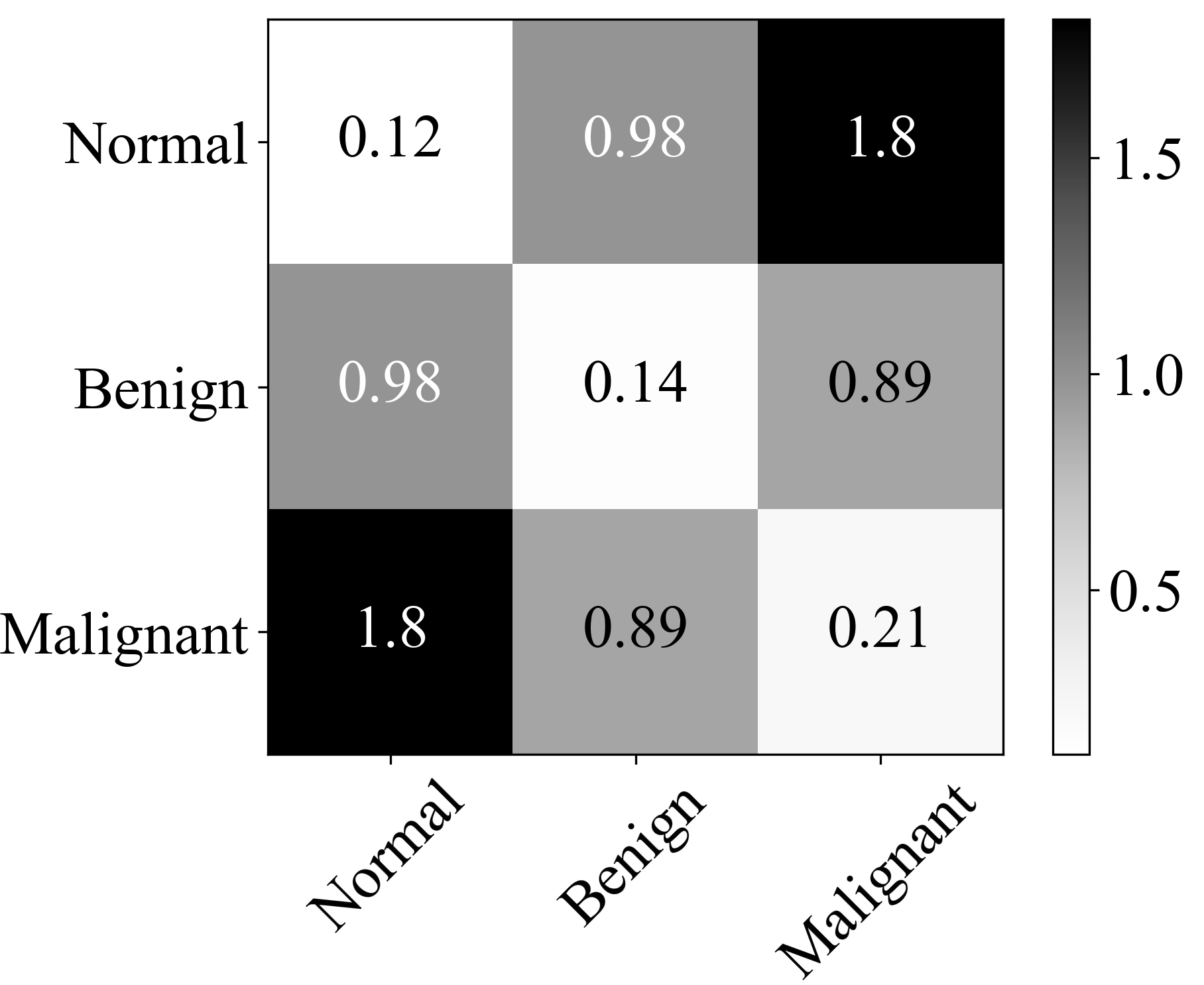}
			\label{birad_corr_onet}}\hspace{2pt}
		\subfloat[$S$-Net]{\includegraphics[width=0.18\linewidth]{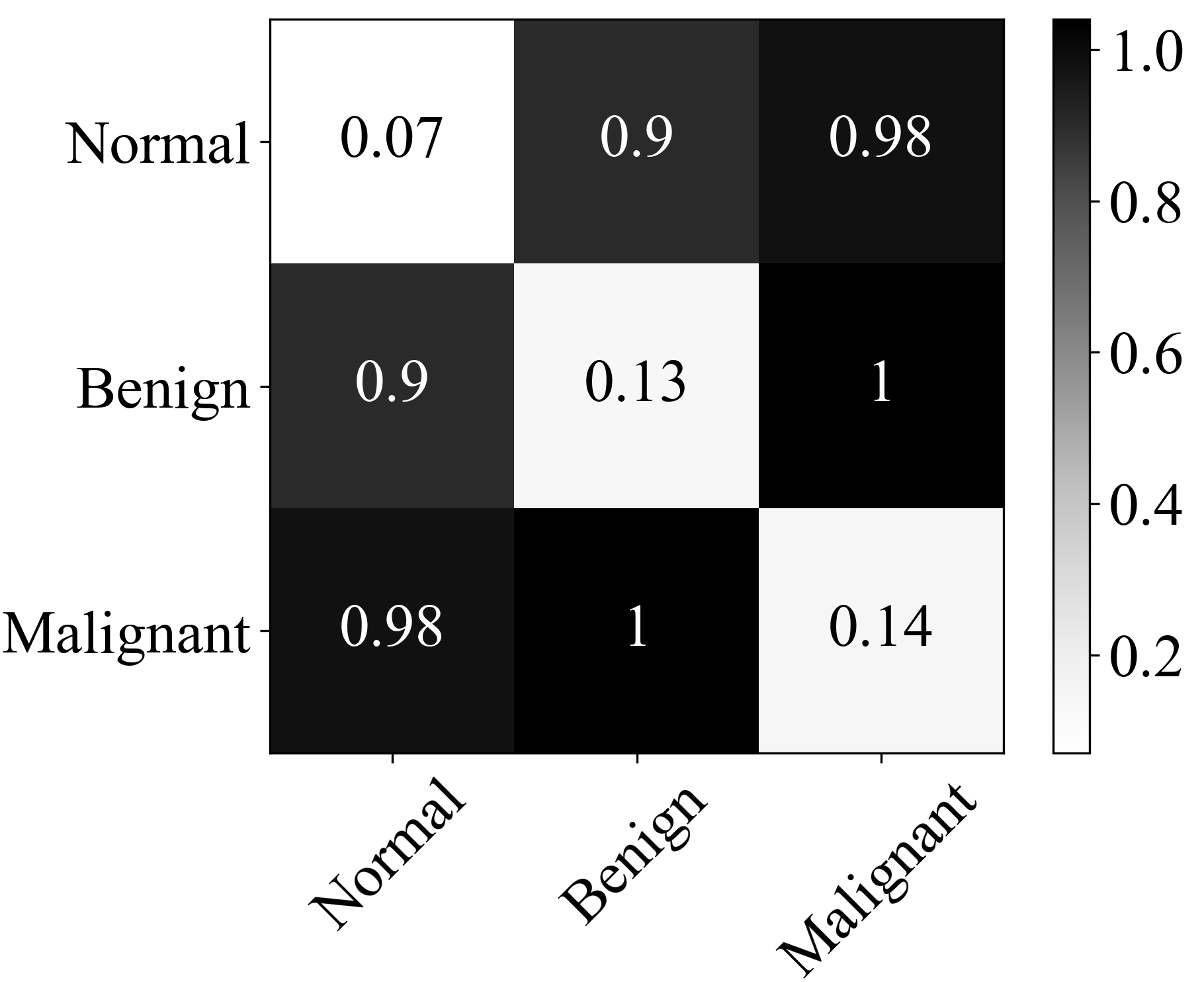}
			\label{birad_corr_snet}}\hspace{2pt}
		\subfloat[$T$-Net]{\includegraphics[width=0.20\linewidth]{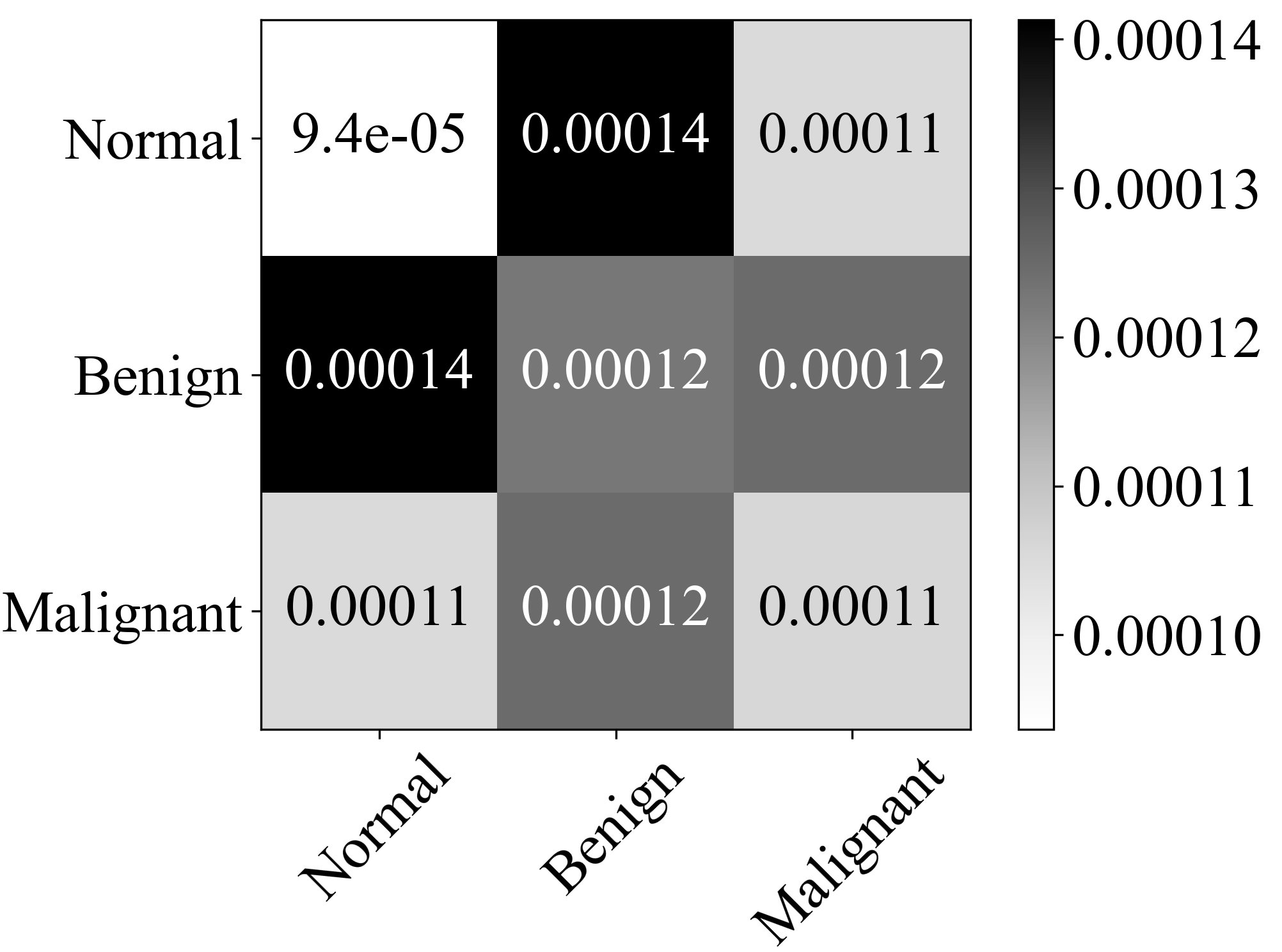}
			\label{birad_corr_tnet}}\hspace{2pt}
		\subfloat[$Q$-Net]{\includegraphics[width=0.186\linewidth]{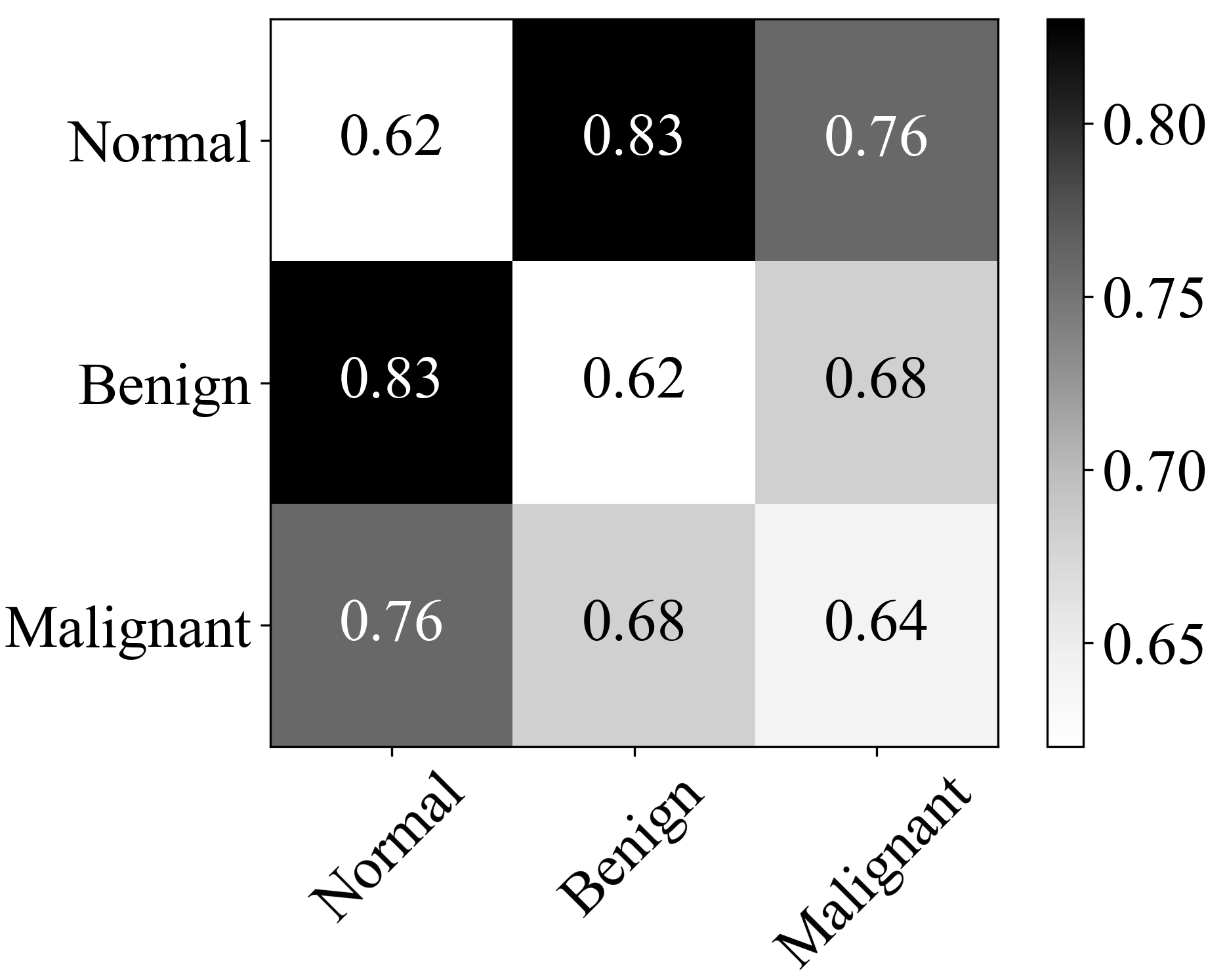}
			\label{birad_corr_qnet}}\hspace{2pt}
		\subfloat[$N$-Net$_2$]{\includegraphics[width=0.18\linewidth]{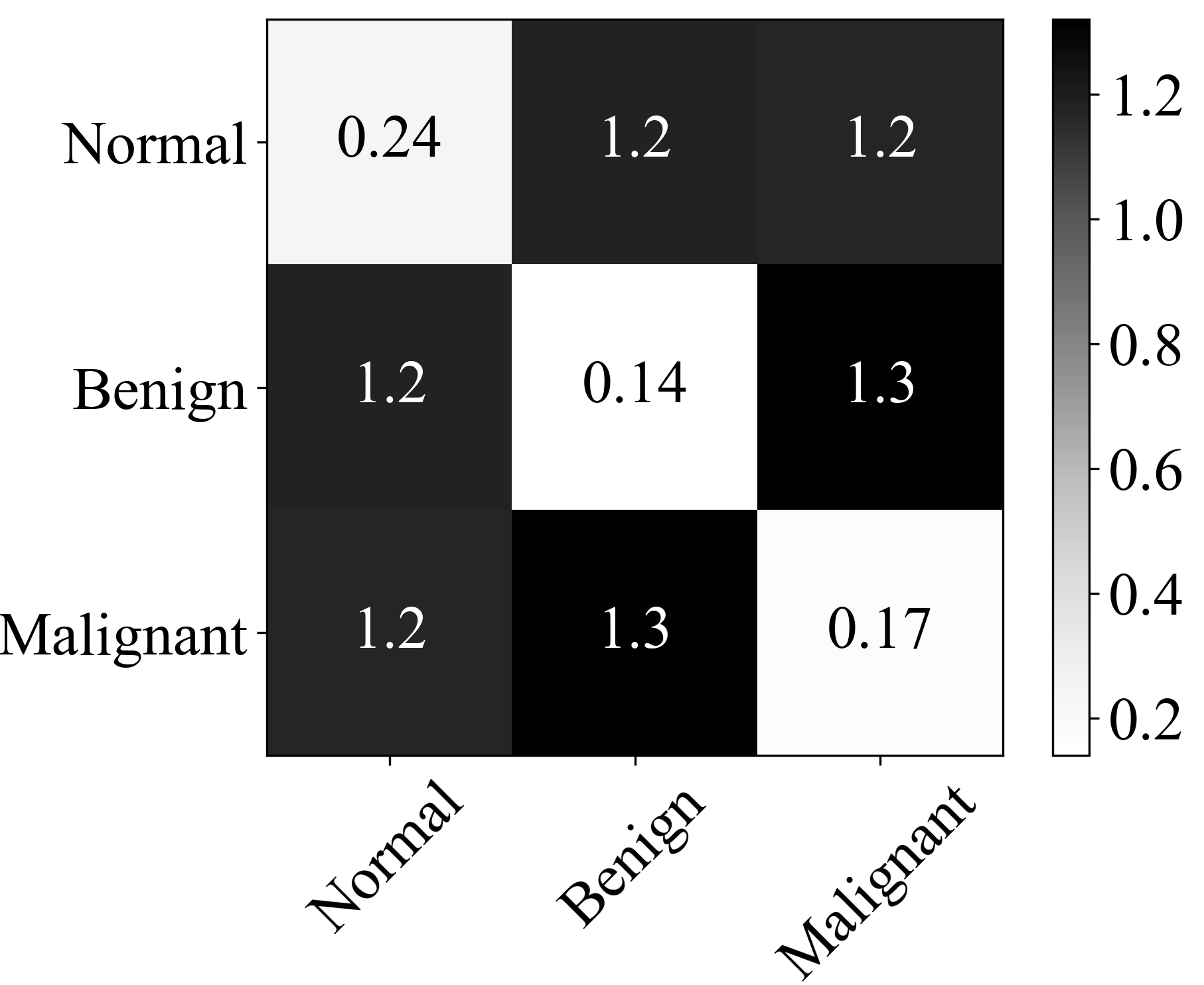}
			\label{birad_corr_nnet2}}\hspace{2pt}
		\caption{Pairwise cosine distance matrix of latent representation in the Busi dataset.}
		\label{fig:corr_matrix_birad}
	\end{figure*}
	\begin{figure*}
		\centering
		\subfloat[$O$-Net]{\includegraphics[width=0.18\linewidth]{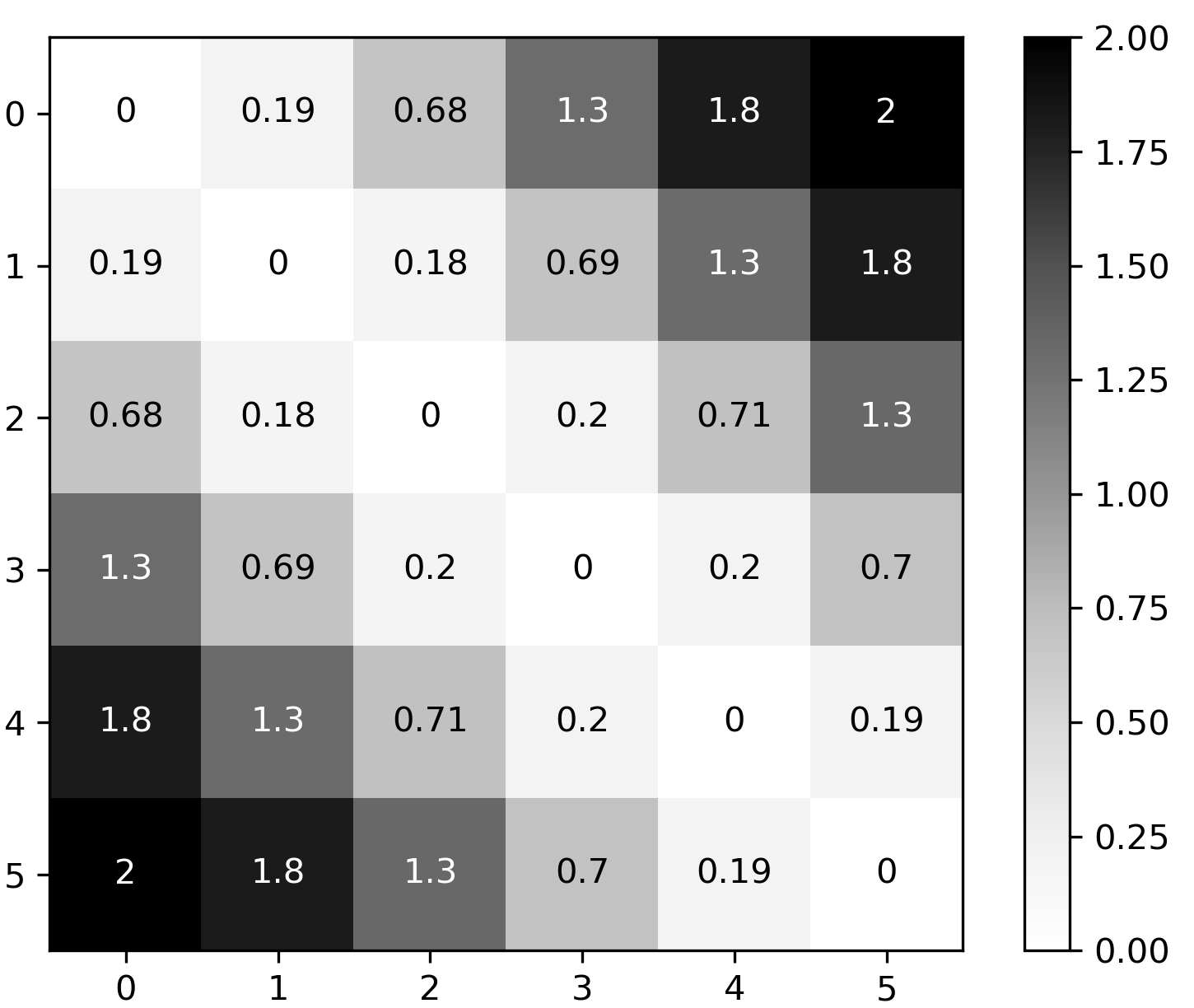}
			\label{finger_corr_onet}}\hspace{3pt}
		\subfloat[$S$-Net]{\includegraphics[width=0.18\linewidth]{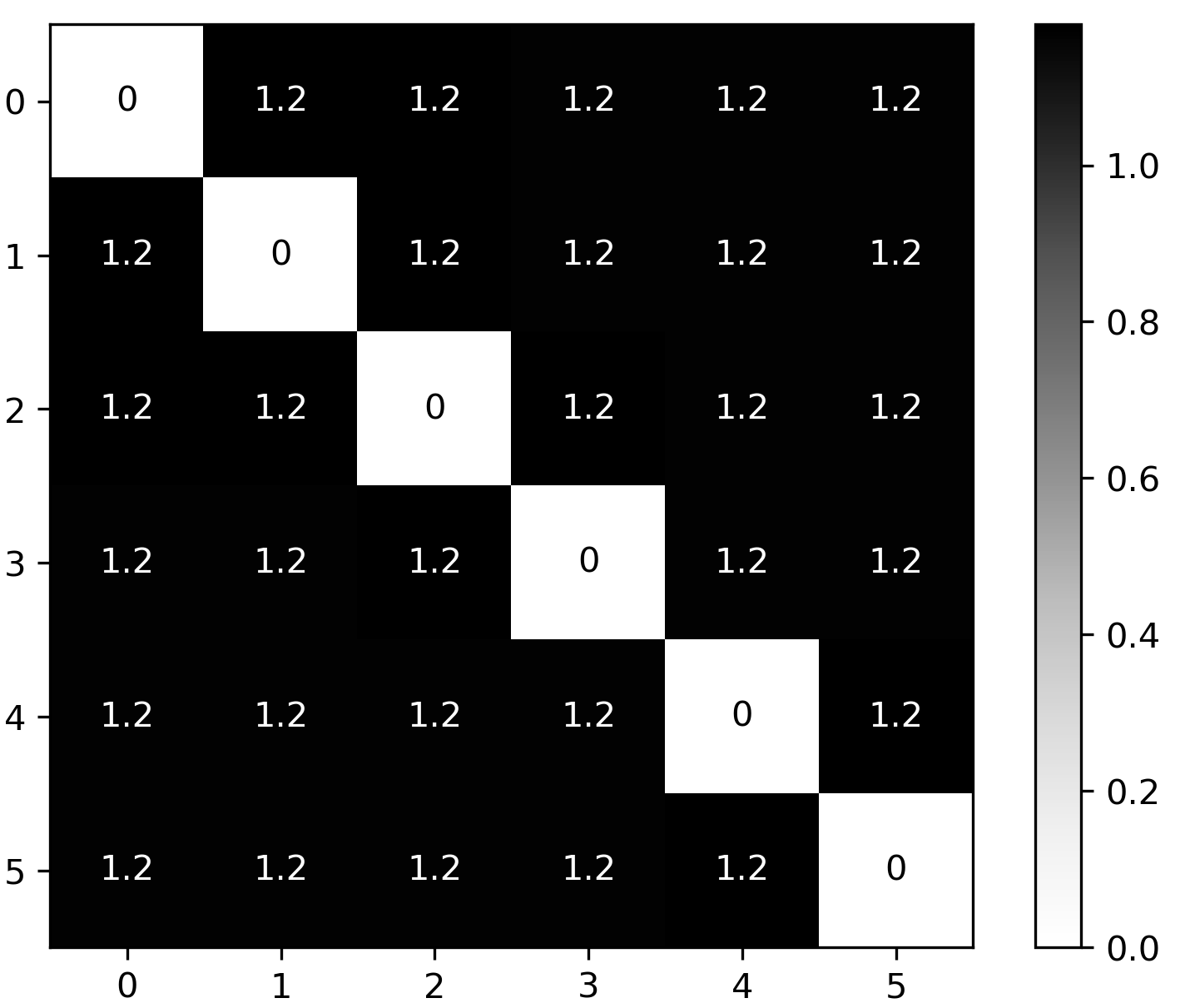}
			\label{finger_corr_snet}}\hspace{3pt}
		\subfloat[$T$-Net]{\includegraphics[width=0.18\linewidth]{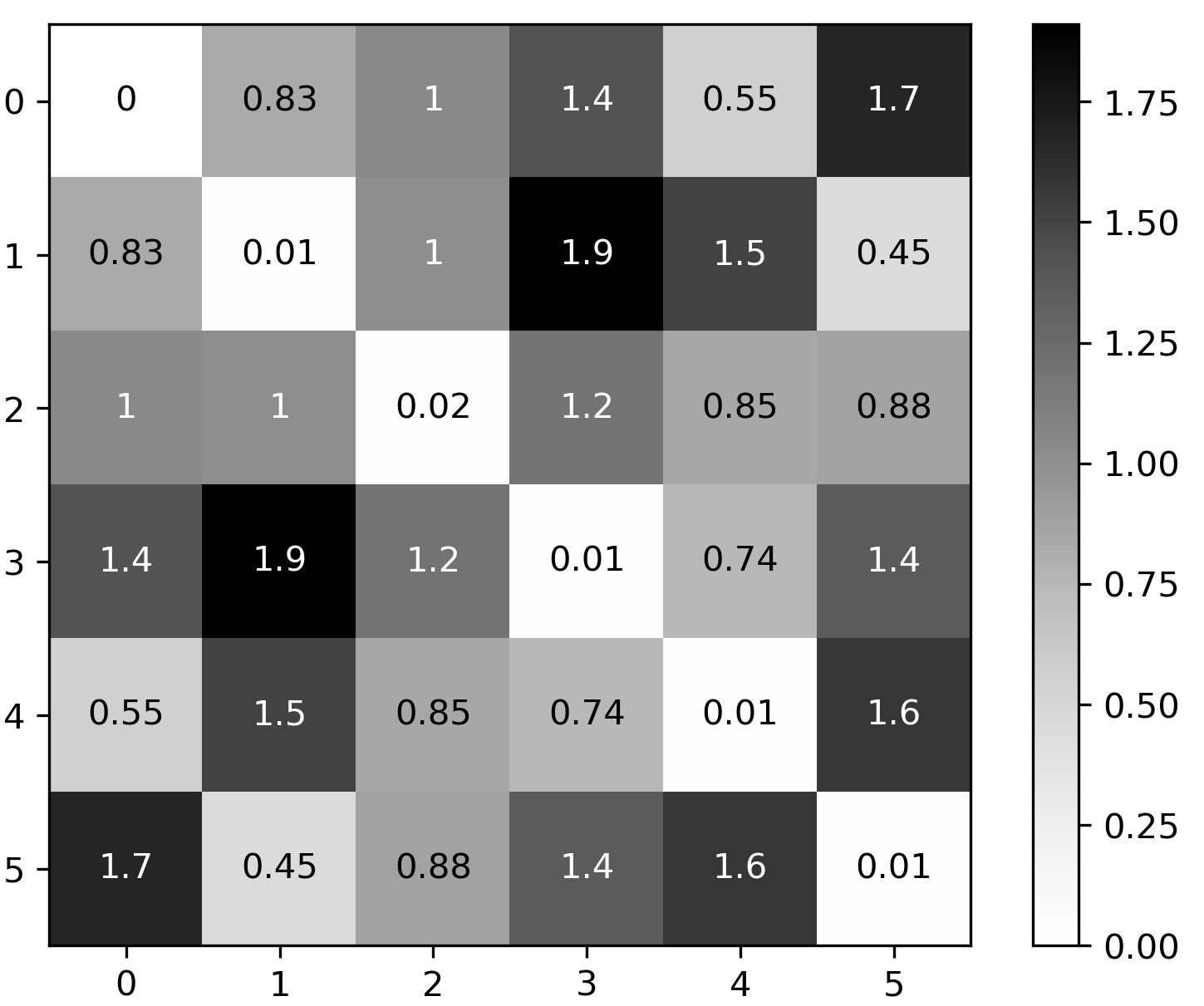}
			\label{finger_corr_tnet}}\hspace{3pt}
		\subfloat[$Q$-Net]{\includegraphics[width=0.18\linewidth]{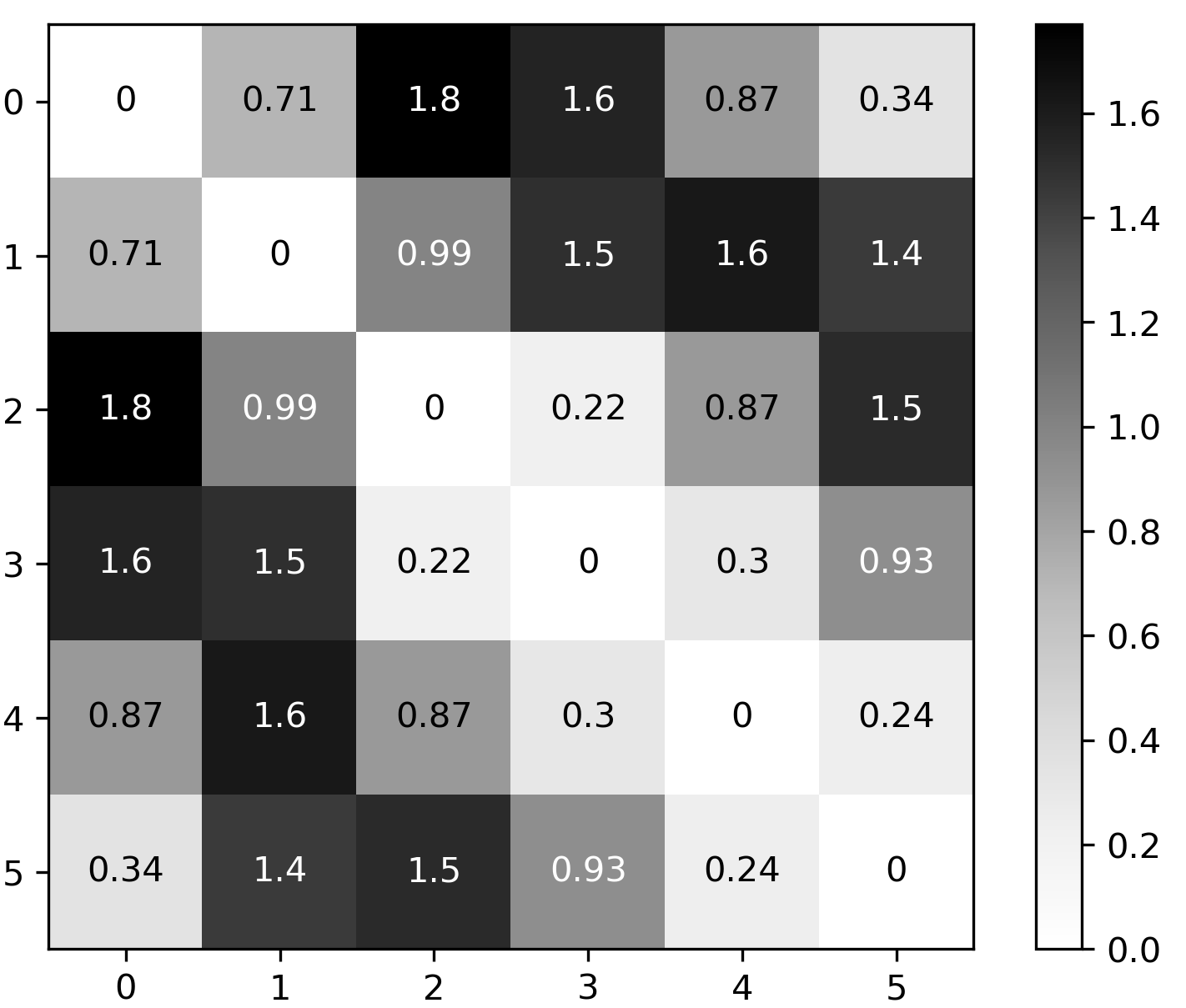}
			\label{finger_corr_qnet}}\hspace{3pt}
		\subfloat[$N$-Net$_2$]{\includegraphics[width=0.18\linewidth]{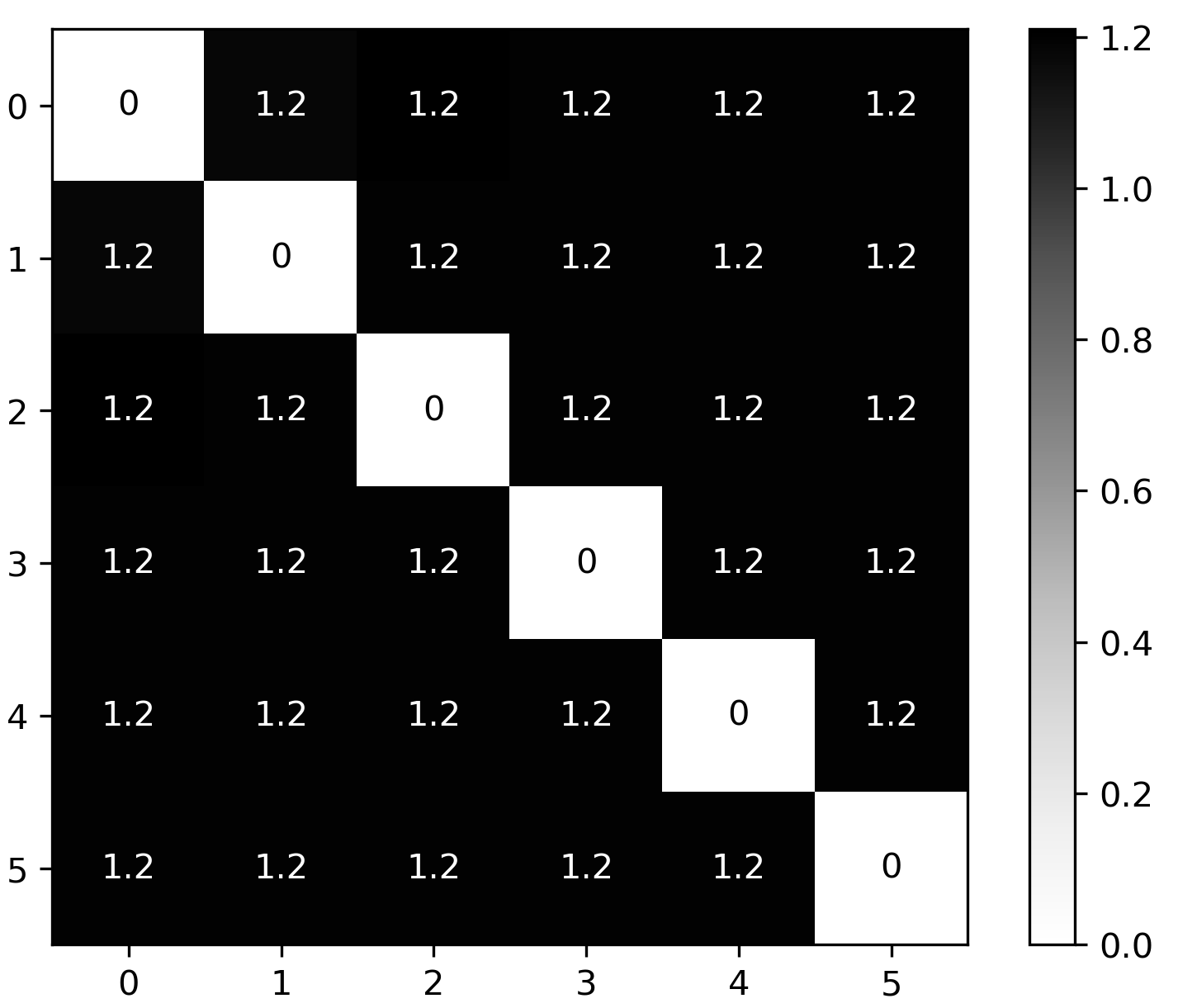}
			\label{finger_corr_nnet2}}
		\caption{Pairwise cosine distance matrix of latent representation in the Finger dataset.}
		\label{fig:corr_matrix_finger}
	\end{figure*}
	\subsubsection{Why does ATD outperform the existing DML methods?} In addition to the binary-based representation, which makes it difficult to capture the inter-category relationship, $S$-Net uses the Euclidean distance (see Table \ref{tab:model_archict_comp} of Appendix \ref{ap:detailed_architect}), which is prone to be less discriminant in high-dimensional data \cite{GIANNELLA2021106115,NEURIPS2020_0561bc7e}. Note that in this study, the default embedding size ($|Z|$) was 100, representing a vector with 100 dimensions, as described in Table \ref{tab:net_architecture} (Appendix \ref{ap:detailed_architect}). $T$-Net performed significantly well in well-defined triplet representations. Thus, it must be guaranteed that the features of the anchor and positive samples are more similar than those of the negative samples unless they are unstable. Many scholars have also reported that triplet loss is less stable and difficult to converge \cite{9238467}. $Q$-Net, which adds a hard-negative constraint to $T$-Net, does not consistently outperform $T$-Net in all datasets. In a few datasets, such as Busi and FG-Net, the quadruplet loss tends to outperform $T$-Net. In contrast to Finger and Adience, which is a large dataset, $T$-Net tends to outperform $Q$-Net. Meanwhile, in addition to its batch-dependent issue, $N$-Net with an N-pair mechanism uses cosine similarity in the merging layer, which has lower performance than angular distance, as reported in natural language processing tasks \cite{cer2018universal}. In comparison, the ATD is a variant of the angular distance and is represented in a triplet fashion. Unlike the triplet and quadruplet representations in $T$-Net and $Q$-Net, respectively, the relationship or distance between the lower, middle, and upper bound is well defined in the angular space. Therefore, it can correctly project all the images based on their categories. Moreover, unlike existing DML models, our proposed network explicitly considers an ordinal relationship as a constraint during training, which can improve the discriminant capacity \cite{SENDIK2019368}.
	\begin{figure*}
		\centering
		\subfloat[$O$-Net]{\includegraphics[width=0.18\linewidth]{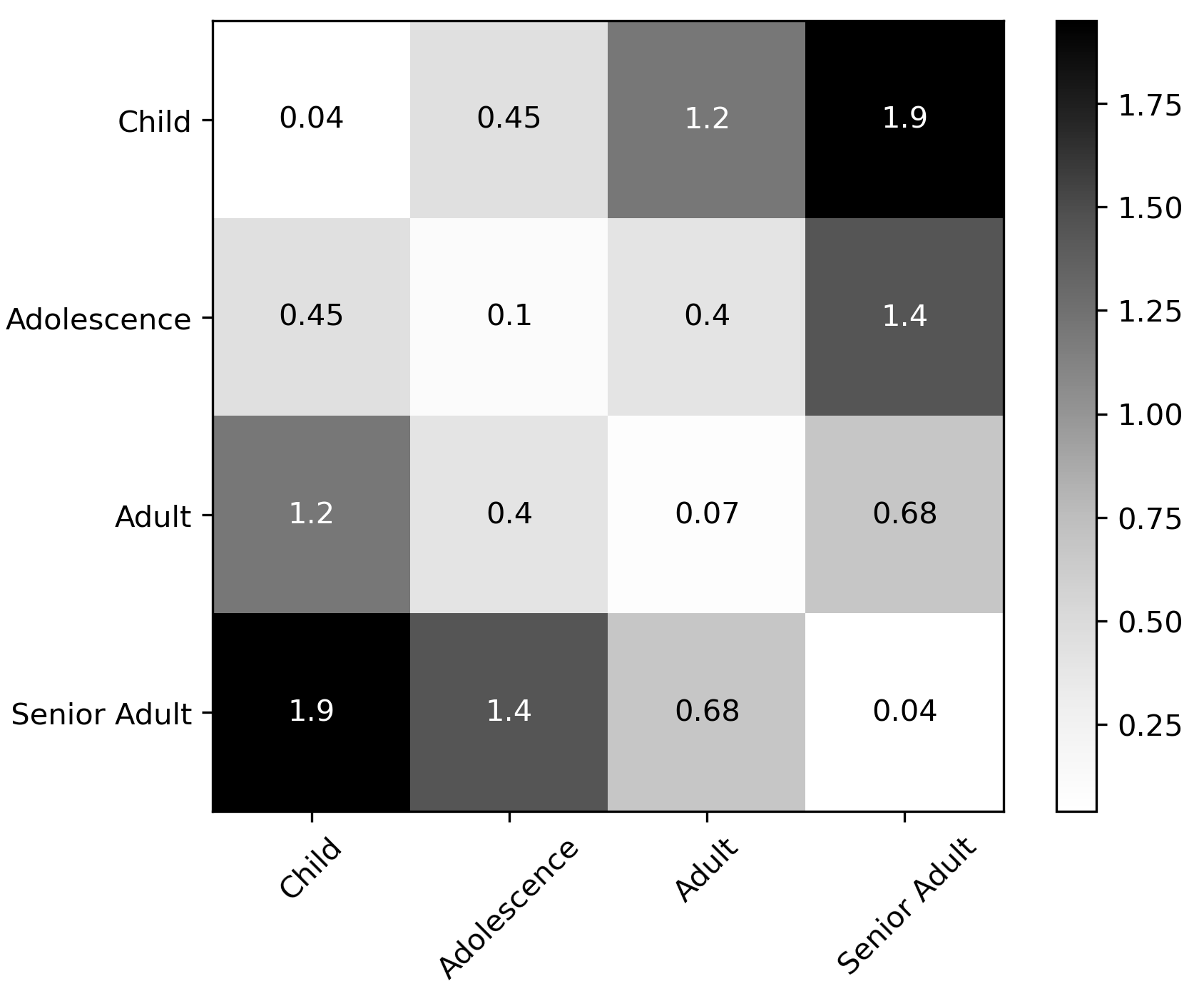}
			\label{fpg_corr_onet}}\hspace{2pt}
		\subfloat[$S$-Net]{\includegraphics[width=0.18\linewidth]{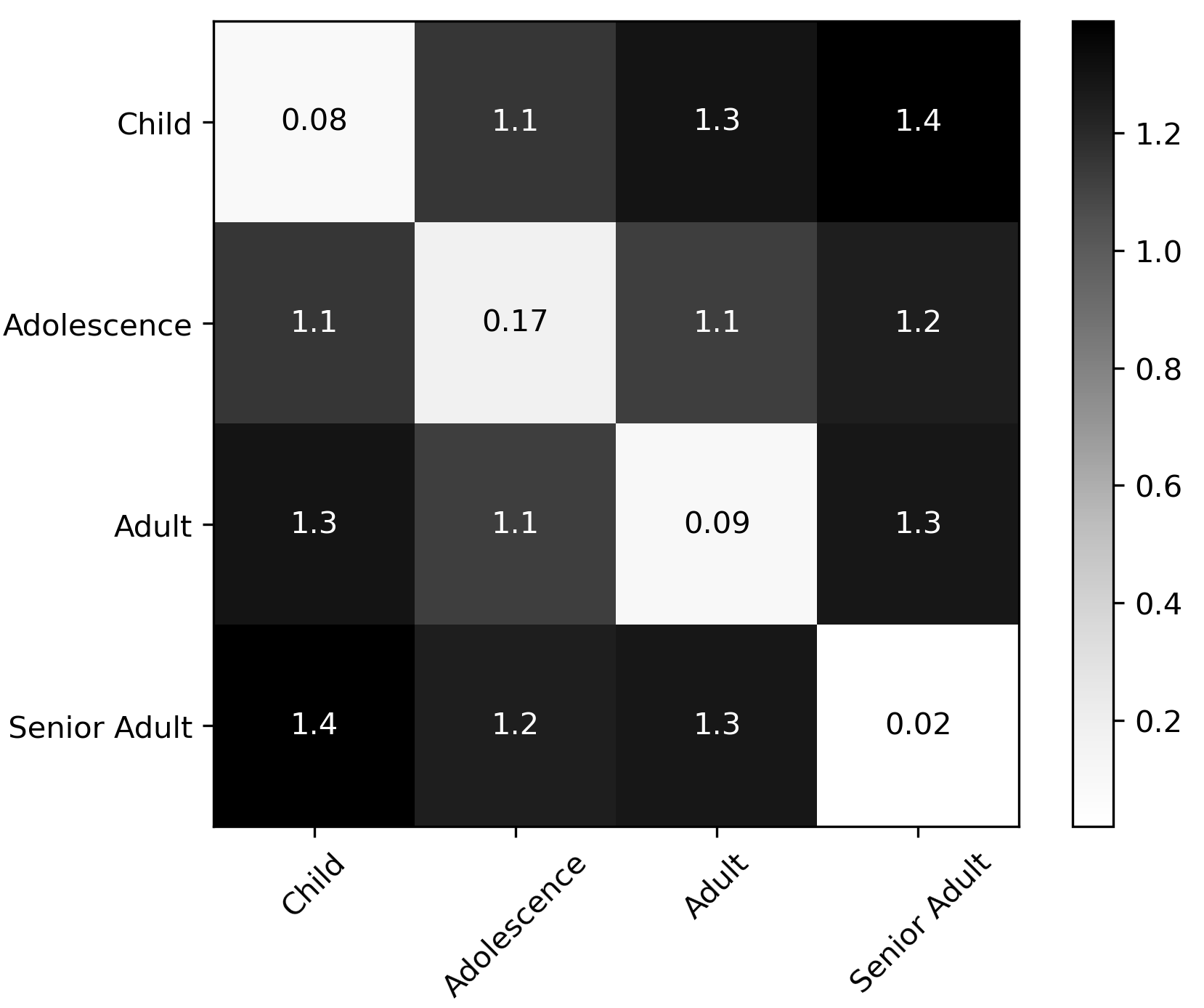}
			\label{fpg_corr_snet}}\hspace{2pt}
		\subfloat[$T$-Net]{\includegraphics[width=0.18\linewidth]{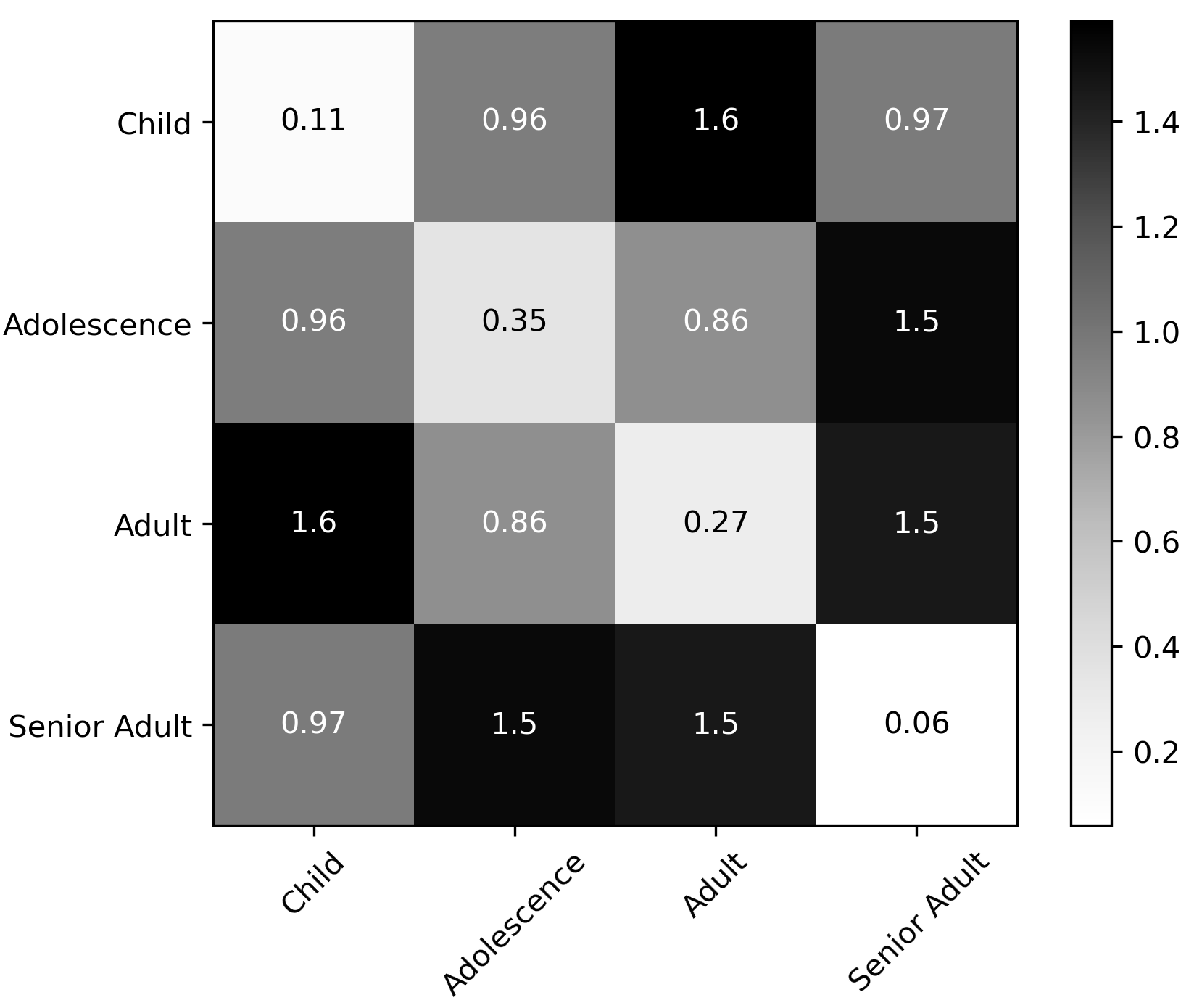}
			\label{fpg_corr_tnet}}\hspace{2pt}
		\subfloat[$Q$-Net]{\includegraphics[width=0.18\linewidth]{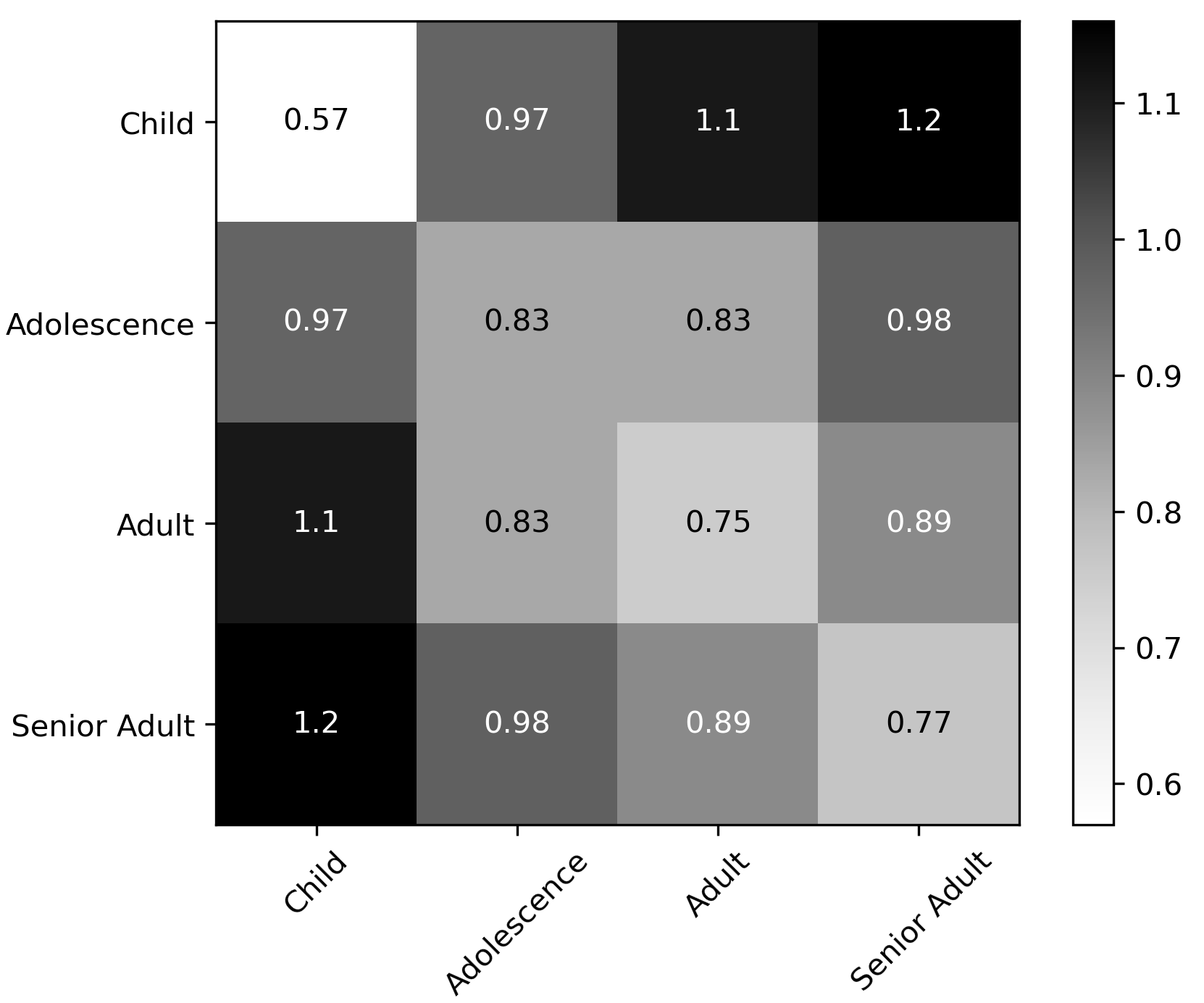}
			\label{fpg_corr_qnet}}\hspace{2pt}
		\subfloat[$N$-Net$_2$]{\includegraphics[width=0.18\linewidth]{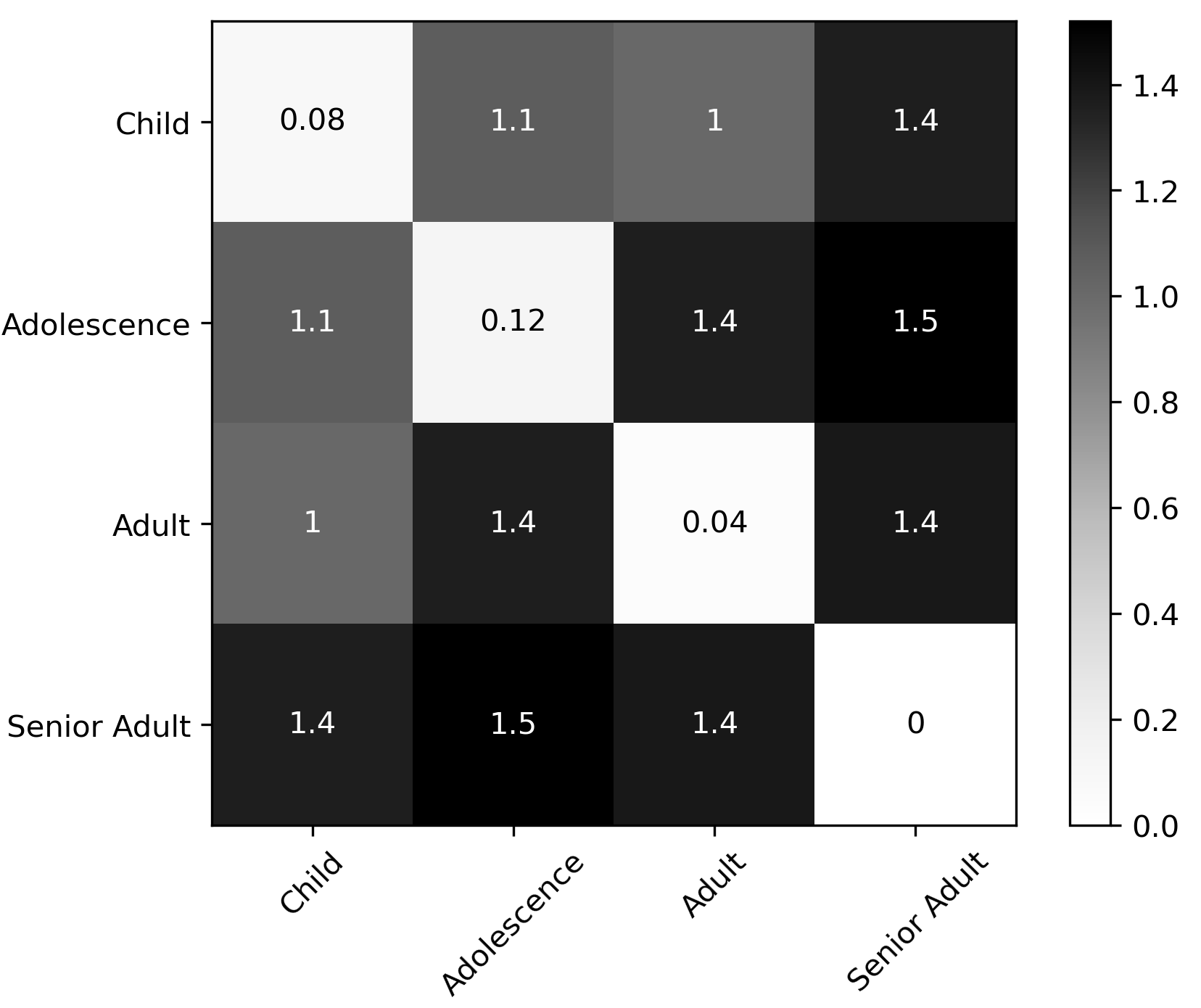}
			\label{fpg_corr_nnet2}}\hspace{2pt}
		\caption{Pairwise cosine distance matrix of latent representation in the FG-Net dataset.}
		\label{fig:corr_matrix_fpg}
	\end{figure*}
	\begin{figure*}
		\centering
		\subfloat[$O$-Net]{\includegraphics[width=0.18\linewidth]{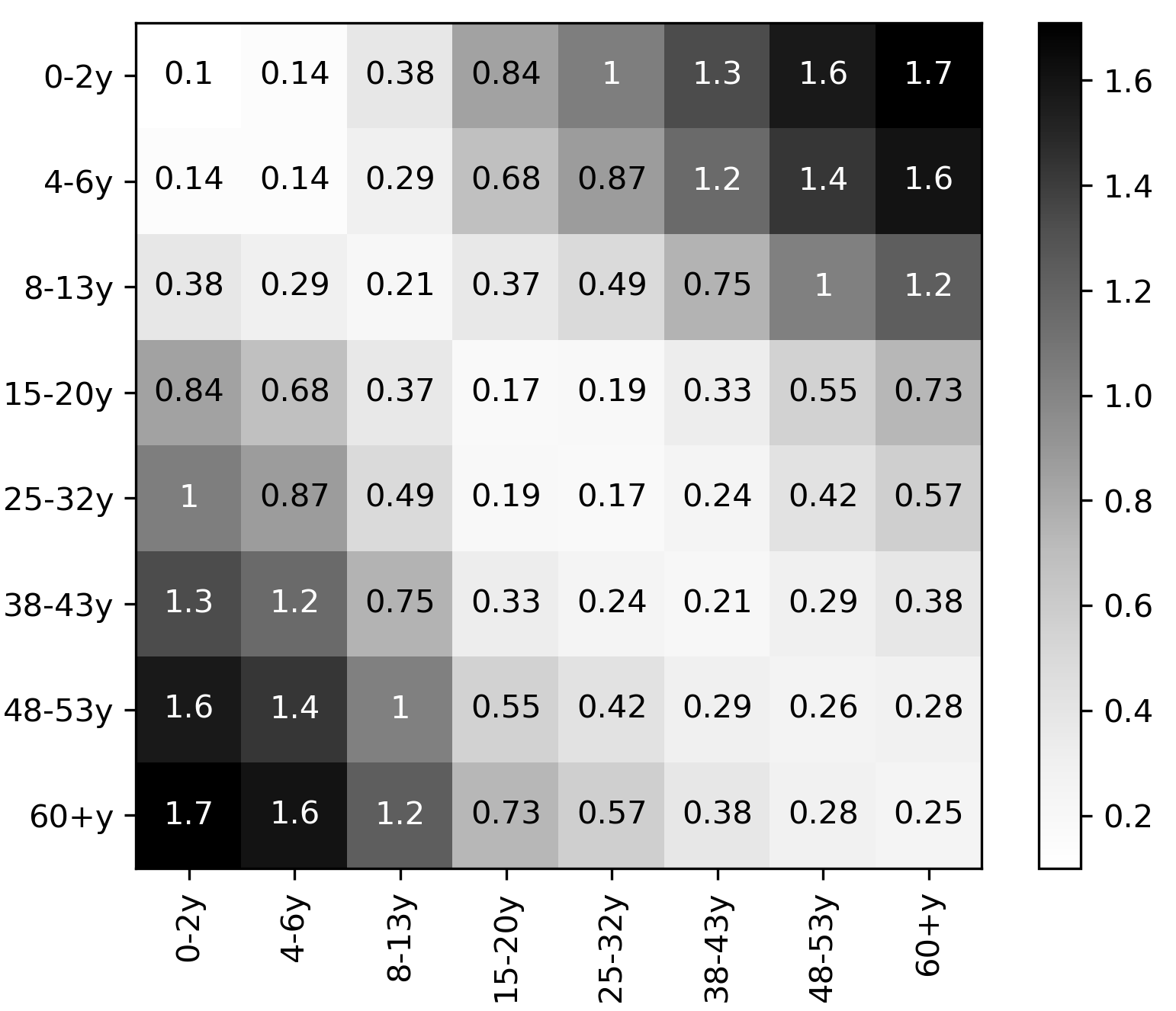}
			\label{adience_corr_onet}}\hspace{3pt}
		\subfloat[$S$-Net]{\includegraphics[width=0.18\linewidth]{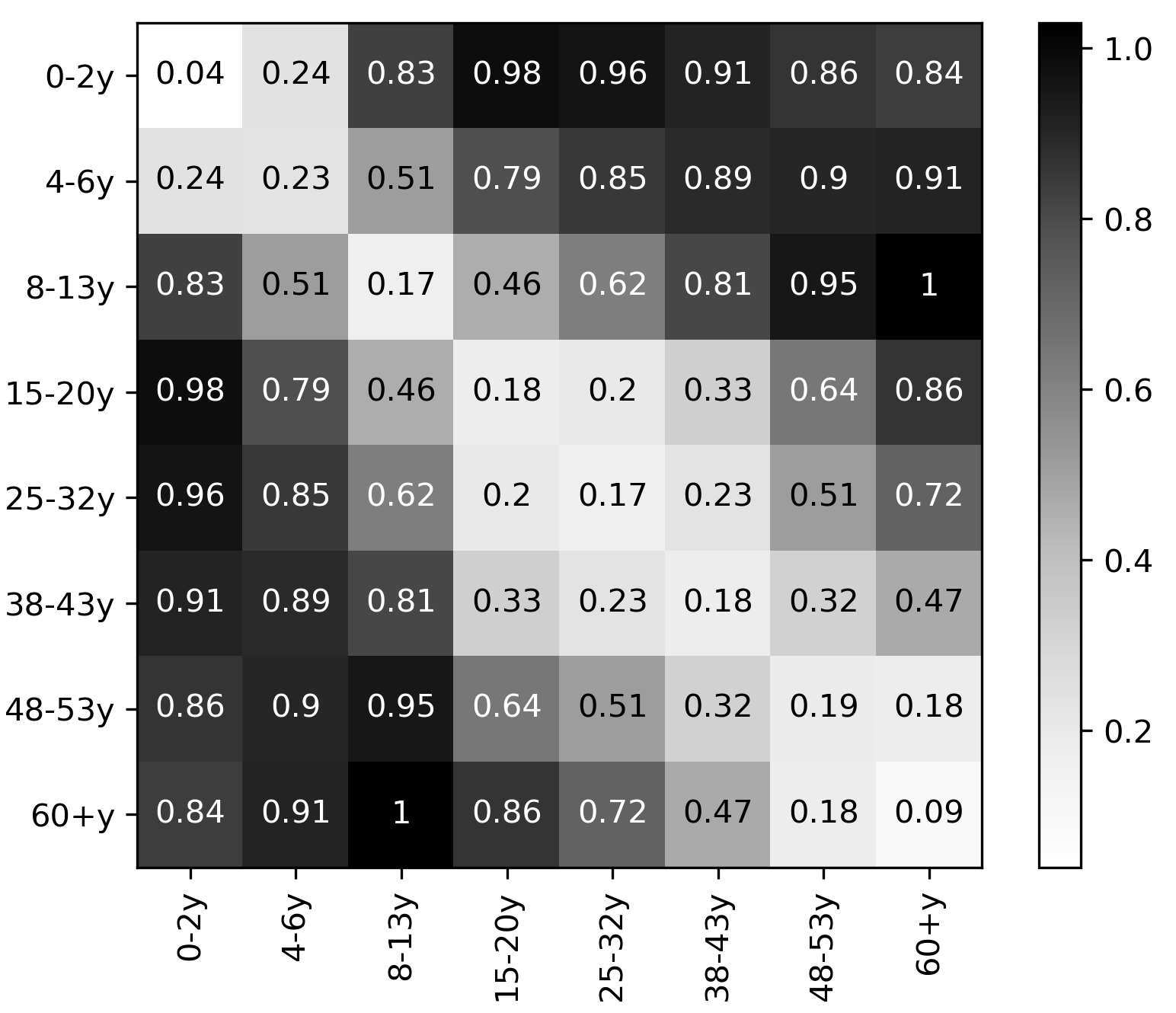}
			\label{adience_corr_snet}}\hspace{3pt}
		\subfloat[$T$-Net]{\includegraphics[width=0.18\linewidth]{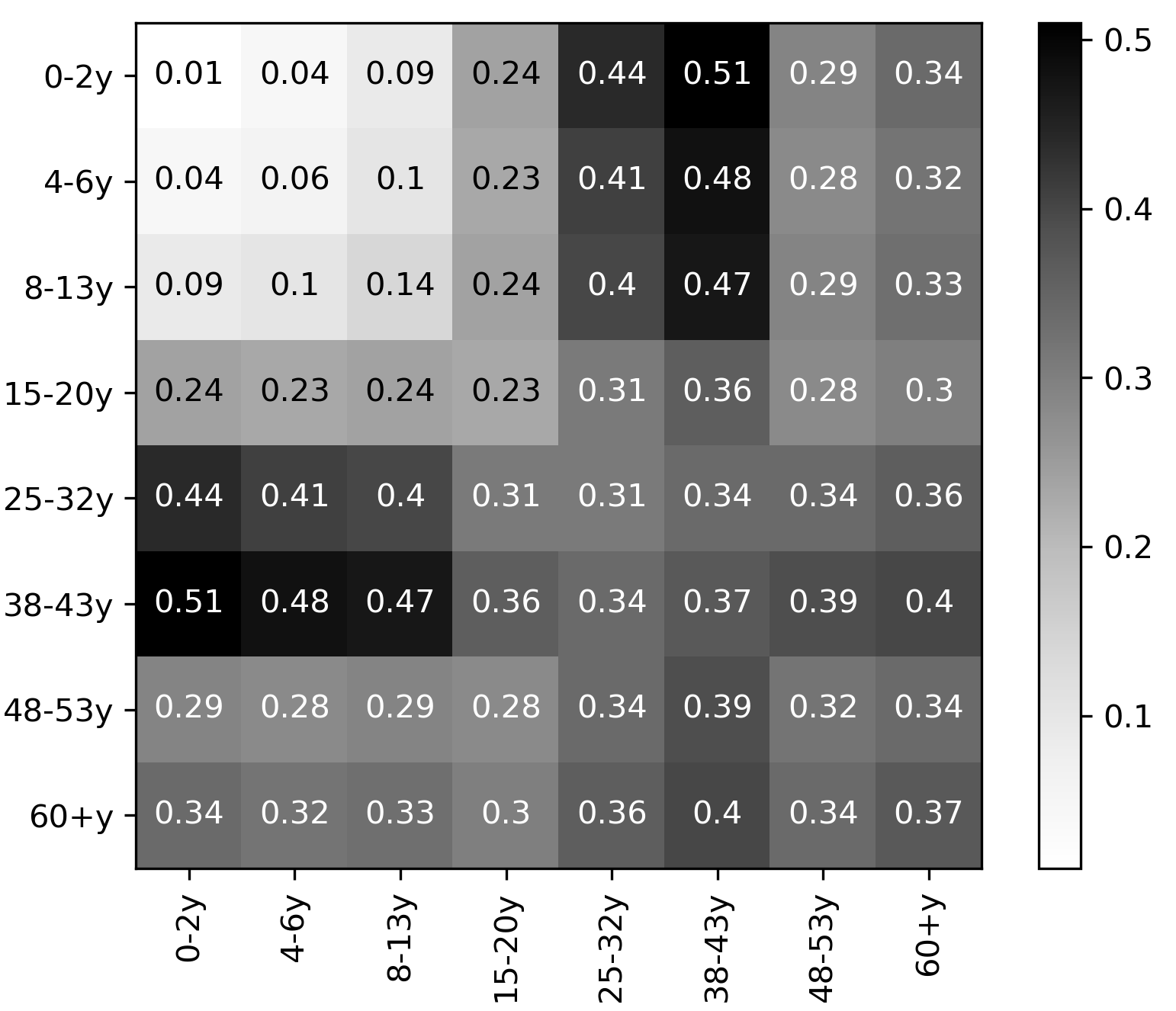}
			\label{adience_corr_tnet}}\hspace{3pt}
		\subfloat[$Q$-Net]{\includegraphics[width=0.18\linewidth]{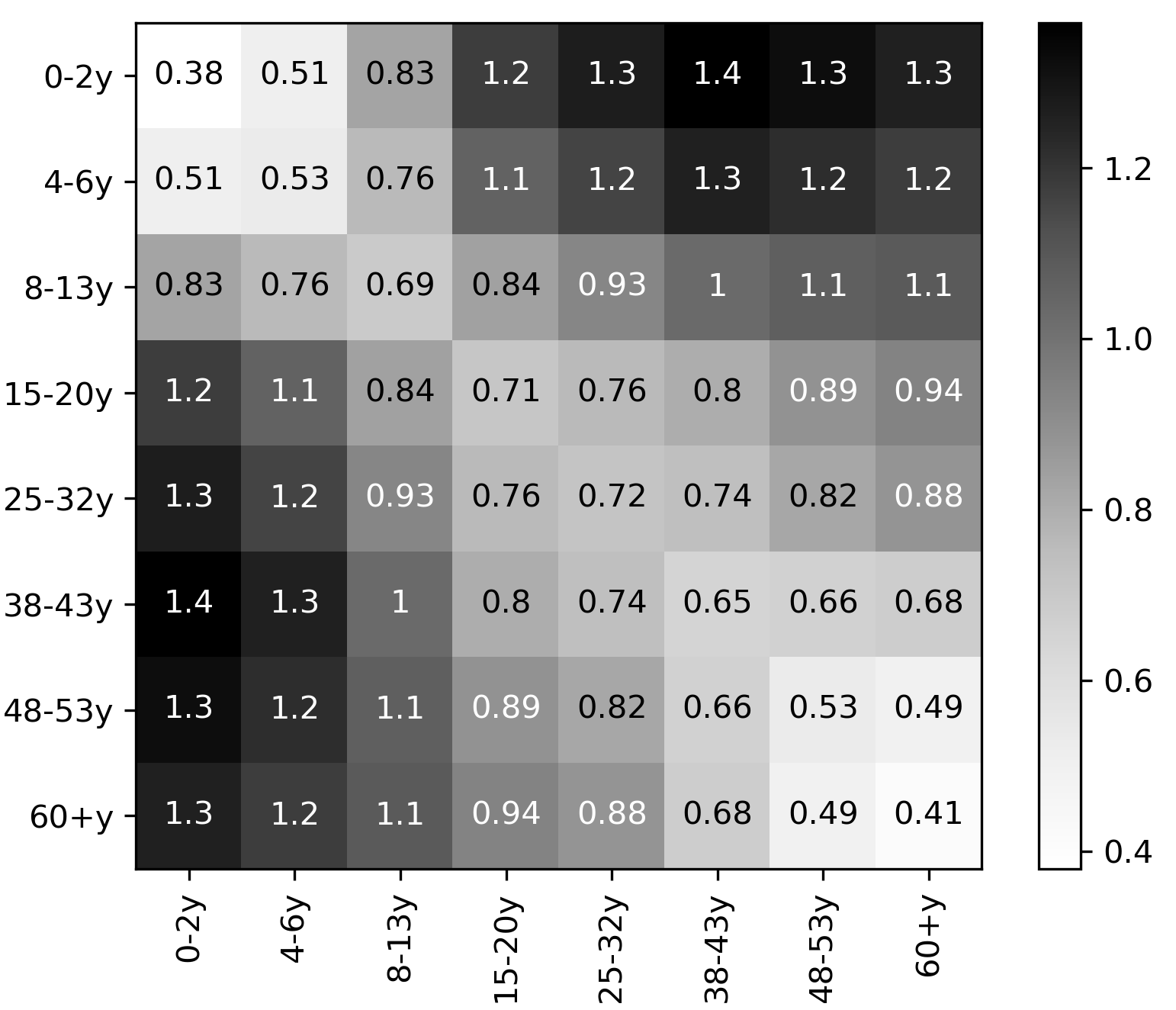}
			\label{adience_corr_qnet}}\hspace{3pt}
		\subfloat[$N$-Net$_2$]{\includegraphics[width=0.18\linewidth]{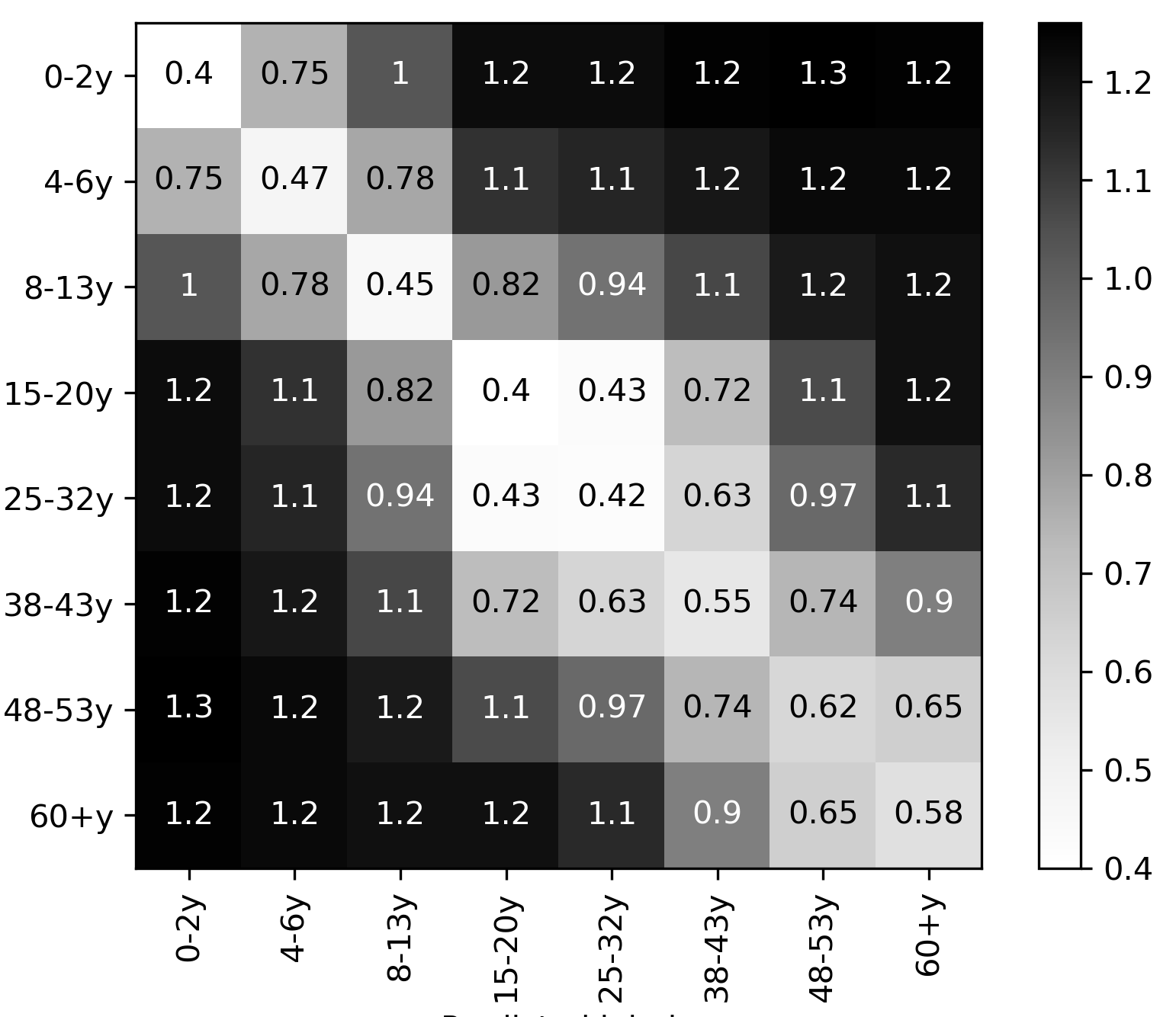}
			\label{adience_corr_nnet2}}
		\caption{Pairwise cosine distance matrix of latent representation in the Adience dataset.}
		\label{fig:corr_matrix_adience}
	\end{figure*}
	\subsubsection{Comparison against conventional ordinal metric learning}
	We present the comparison of $O$-Net against conventional ordinal metric learning. We employed four public datasets from the UCI machine learning repository. The datasets have ordinal features both in dependent and independent variables, such as Car, Nursery, Hayes-Roth, and Balance datasets. The number of instances in the Car, Nursery, Hayes-Roth, and Balance datasets is 1728, 12960, 160, and 625, respectively. We used the same experimental data, setting, and evaluation as \cite{Shi_Li_Sha_2016}. As shown in Table \ref{tab:classification_knn_comp}, our proposed method ($O$-Net) generally outperforms the conventional ordinal metric learning methods. Please note that the Hayes-Roth and Balance have a small number of instances (160 and 625 samples, respectively); thus, the deep learning-based model ($O$-Net) might not work optimally.

	\begin{table}
		\centering
		\caption{$K$-NN classification error on the datasets with ordinal features. $K=3$}
		\label{tab:classification_knn_comp}
		\resizebox{1\linewidth}{!}{
			\centering
			\begin{tabular}{l l l l l}		
				\toprule
				\multirow{2}{*}{Model} & \multicolumn{4}{c}{Error rate (mean $\pm$ std.)}\\
				\cmidrule{2-5}
				&	\qquad Car 						&	\qquad Nursery					&	\qquad Hayes-Roth					&	\qquad Balance			\\
				\midrule
				Real-Eucl \cite{Shi_Li_Sha_2016}					&	\qquad $11.4\pm0.7$				&\qquad 	$8.6\pm0.1$				&\qquad 	$38.5\pm3.1$				&\qquad 	$15.2\pm1.1$	\\
				Real-LMNN \cite{Shi_Li_Sha_2016}					&	\qquad $5.0\pm0.3$				&\qquad 	$2.4\pm0.1$				&\qquad 	$23.1\pm1.6$				&\qquad 	$17.6\pm0.9$	\\
				Binary-Eucl \cite{Shi_Li_Sha_2016}					&	\qquad $24.0\pm1.4$				&\qquad 	$24.0\pm0.2$			&\qquad 	$50.0\pm2.9$				&\qquad 	$32.7\pm1.9$	\\
				Binary-LMNN	\cite{Shi_Li_Sha_2016}					&	\qquad $4.1\pm0.3$				&\qquad 	$2.3\pm0.1$				&\qquad 	$\textbf{16.0}\pm1.0$				&\qquad 	$17.8\pm1.2$	\\
				Binary-Ord-Eucl	\cite{Shi_Li_Sha_2016}				&	\qquad $12.3\pm0.4$				&\qquad 	$8.7\pm0.1$				&\qquad 	$45.5\pm3.3$				&\qquad 	$16.7\pm0.8$	\\
				Binary-Ord-LMNN \cite{Shi_Li_Sha_2016}				&	\qquad $4.1\pm0.2$				&\qquad 	$1.9\pm0.1$				&\qquad 	$15.4\pm1.2$				&\qquad 	$13.4\pm0.8$	\\
				Ex-Gower \cite{Shi_Li_Sha_2016}						&	\qquad $12.1\pm0.7$				&\qquad 	$8.8\pm0.1$				&\qquad 	$37.2\pm1.9$				&\qquad 	$32.8\pm1.3$	\\
				Thresh-Eucl	\cite{Shi_Li_Sha_2016}					&	\qquad $4.5\pm0.4$				&\qquad 	$2.3\pm0.1$				&\qquad 	$22.3\pm1.9$				&\qquad 	$14.5\pm0.7$	\\
				Ord-LMNN-Uni \cite{Shi_Li_Sha_2016}					&	\qquad $3.8\pm0.3$				&\qquad 	$1.8\pm0.1$				&\qquad 	$20.5\pm1.3$				&\qquad 	$6.8\pm0.5$	\\
				Ord-LMNN-Beta \cite{Shi_Li_Sha_2016}				&	\qquad $3.7\pm0.3$				&\qquad 	$1.6\pm0.1$				&\qquad 	$18.6\pm1.0$				&\qquad 	$\textbf{6.1}\pm0.5$	\\
				Ord-LMNN-RecBeta \cite{Shi_Li_Sha_2016}				&	\qquad $3.4\pm0.3$				&\qquad 	$1.6\pm0.1$				&\qquad 	$18.6\pm1.0$				&\qquad 	$6.4\pm0.5$	\\
				$O$-Net												&	\qquad $\textbf{3.1}\pm0.4$				&\qquad 	$\textbf{0.8}\pm0.1$				&\qquad 	$\textbf{16.0}\pm1.5$				&\qquad 	$\textbf{6.1}\pm0.6$	\\
				\bottomrule
			\end{tabular}
		}
	\end{table}
	\subsection{Semantic analysis}
	\label{subsec:semantic_analysis}
	In this study, semantic analysis is the process of drawing meaning from embedded space representations generated by the DML model. Therefore, the interpretation of the embedding space should be similar to that of the original space. Hence, when the categories of input data are ordinal, it must also maintain its ordinal nature in the embedding space representation. For example, the distance between category I and category II samples should be less than that between category I and category III samples. We used a favored cosine distance to measure the relationships among them because the embedding space’s latent representation is a vector with 100 dimensions. Thus, pairwise comparison using cosine distance was conducted on all samples to obtain the ordinal relationships among the samples, and the results are shown in Figs. \ref{fig:corr_matrix_birad}, \ref{fig:corr_matrix_finger}, \ref{fig:corr_matrix_fpg}, and \ref{fig:corr_matrix_adience} for the Busi, Finger, FG-Net, and Adience datasets, respectively. It is worth noting that $S$-Net only obtained an adequate latent representation in the Busi dataset, which contains only three categories. Surprisingly, our model consistently preserved the ordinal relations among the categories in all datasets. In contrast to most existing DML models, the relationships among the categories tend to be scattered. Moreover, the most fascinating finding in this study is that all DML models have a perfect mapping accuracy in the Finger dataset, as reported in Section \ref{subsec:embedding_acc}. Nevertheless, its latent representation in the embedding space is still messy based on the ordinal relationship of the original space. Therefore, we can conclude that the accuracy of latent code representation cannot guarantee adequate semantic meaning of latent code representation in ordinal data.
	
	\section{Conclusion}
	\label{sec:conclusion}
	Ordinal data are widespread in real-world problems and have been frequently considered standard nominal problems, which can lead to non-optimal solutions. This study introduces an OTD with an ATD, particularly designed for solving ordinal DMLs. Our empirical experiments demonstrated that our proposed method obtains a more accurate and semantic embedding space representation compared with existing DML models. Owing to its simplicity, the ATD can be applied to other machine learning methods dealing with ordinal data. Moreover, when the embedding representation is accurate and meaningful, a simple traditional machine learning method, such as nearest neighbor and clustering algorithms, can solve a modern machine learning task that deals with complex and high-dimensional data. In addition to the effectiveness of our proposed method, we acknowledge its limitations, particularly as a triplet representation potentially requires a high computational cost in a large-scale dataset. This issue leads to further research directions to be addressed in the future.

	\appendices
	\label{sec:appendix}
	\section{ATD is a distance metric}
	\label{ap:proof_dist_metric}
	\begin{proof}
		Nonnegativity, identity, symmetry, and triangle inequality properties of ATD. \\
		\textbf{Non-negativity}: $D_{AT}(z^{l_{m}}_i, z^{l_{n}}_j, z^{l_{o}}_k)\geq0$ \\ 
		$D_A(z^{l_{m}}_i, z^{l_{n}}_j) + D_A(z^{l_{n}}_j, z^{l_{o}}_k)\geq0 $ \\
		$D_A(z^{l_{m}}_i, z^{l_{n}}_j) = \frac{\cos^{-1}(S_C(z^{l_{m}}_i,z^{l_{n}}_j))}{\pi} = \frac{{\theta}_{z^{l_{m}}_i,z^{l_{n}}_j}}{\pi} = [0,1],$\\ 
		$D_A(z^{l_{n}}_j, z^{l_{o}}_k) = \frac{\cos^{-1}(S_C(z^{l_{n}}_j,z^{l_{o}}_k))}{\pi} = \frac{{\theta}_{z^{l_{n}}_j,z^{l_{o}}_k}}{\pi} = [0,1],$\\
		$\because {\theta}_{z^{l_{m}}_i,z^{l_{n}}_j}, {\theta}_{z^{l_{n}}_j,z^{l_{o}}_k} \in [0,\pi]$,\\ 
		$\therefore D_A(z^{l_{m}}_i, z^{l_{n}}_j) + D_A(z^{l_{n}}_j, z^{l_{o}}_k) \geq0$.\\
		\textbf{Identity}: $D_{AT}(z^{l_m}_i, z^{l_m}_j, z^{l_m}_k)=0$ \\
		$D_A(z^{l_m}_i, z^{l_m}_j) + D_A(z^{l_m}_j, z^{l_m}_k) = 0$\\
		$\frac{\cos^{-1}(S_C(z^{l_m}_i,z^{l_m}_j))}{\pi} + \frac{\cos^{-1}(S_C(z^{l_m}_j,z^{l_m}_k))}{\pi}=0$\\
		$\frac{{\theta}_{z^{l_m}_i,z^{l_m}_k}}{\pi}=0$\\
		$\because z_i$ and $z_k$ are in the same category, $\therefore z^{l_m}_i,z^{l_m}_k = 0$. \\
		\textbf{Symmetry}: $D_{AT}(z^{l_m}_i, z^{l_n}_j, z^{l_o}_k) = D_{AT}(z^{l_o}_i, z^{l_n}_j, z^{l_m}_k)$\\
		$D_A(z^{l_m}_i, z^{l_n}_j) + D_A(z^{l_n}_j, z^{l_o}_k) = D_A(z^{l_o}_i, z^{l_n}_j) + D_A(z^{l_n}_j, z^{l_m}_k)$ \\
		$\frac{\cos^{-1}(S_C(z^{l_m}_i,z^{l_n}_j))}{\pi} + \frac{\cos^{-1}(S_C(z^{l_n}_j,z^{l_o}_k))}{\pi}=\frac{\cos^{-1}(S_C(z^{l_o}_i,z^{l_n}_j))}{\pi} + \frac{\cos^{-1}(S_C(z^{l_n}_j,z^{l_m}_k))}{\pi}$\\
		$\frac{\theta_{z^{l_m}_i,z^{l_o}_k}}{\pi} = \frac{\theta_{z^{l_o}_i,z^{l_m}_k}}{\pi}$\\
		Note that the degree value remains the same whether we read from the left or right,\\
		$\therefore \theta_{z^{l_m}_i,z^{l_o}_k} = \theta_{z^{l_o}_i,z^{l_m}_k}$.\\
		\textbf{Triangle inequality}: $D_{AT}(z^{l_m}_i, z^{l_n}_j, z^{l_o}_k)+D_{AT}(z^{l_o}_i, z^{l_p}_j, z^{l_q}_k)\geq D_{AT}(z^{l_m}_i, z^{l_o}_j, z^{l_q}_k)$\\
		$D_{AT}(z^{l_m}_i, z^{l_n}_j, z^{l_o}_k) = D_A(z^{l_m}_i, z^{l_n}_j) + D_A(z^{l_n}_j, z^{l_o}_k) = \frac{\theta_{z^{l_m}_i,z^{l_o}_k}}{\pi}$, \\
		$D_{AT}(z^{l_o}_i, z^{l_p}_j, z^{l_q}_k) = D_A(z^{l_o}_i, z^{l_p}_j) + D_A(z^{l_p}_j, z^{l_q}_k) = \frac{\theta_{z^{l_o}_i,z^{l_q}_k}}{\pi}$, \\
		$D_{AT}(z^{l_m}_i, z^{l_o}_j, z^{l_q}_k) = D_A(z^{l_m}_i, z^{l_o}_j) + D_A(z^{l_o}_j, z^{l_q}_k) = \frac{\theta_{z^{l_m}_i,z^{l_q}_k}}{\pi}$, \\
		$\theta_{z^{l_m}_i,z^{l_o}_k} = \cos^{-1}(Sc(z^{l_m}_i,z^{l_o}_k)),$\\
		$\theta_{z^{l_o}_i,z^{l_q}_k} = \cos^{-1}(Sc(z^{l_o}_i,z^{l_q}_k)),$\\
		$\theta_{z^{l_m}_i,z^{l_q}_k} = \cos^{-1}(Sc(z^{l_m}_i,z^{l_q}_k)),$\\
		
		For simplicity, we denote $\theta_{z^{l_m}_i,z^{l_o}_k} = \cos{\theta}_{mo},\,$ $\theta_{z^{l_o}_i,z^{l_q}_k} = \cos{\theta}_{oq},\,$ $\theta_{z^{l_m}_i,z^{l_q}_k} = \cos{\theta}_{mq}$. \\
		Then, using the following Gram matrix: 
		\[
		A =
		\begin{bmatrix}
			1 & \cos{\theta}_{mo} & \cos{\theta}_{mq}\\
			\cos{\theta}_{mo} & 1 & \cos{\theta}_{oq}\\
			\cos{\theta}_{mq} & \cos{\theta}_{oq} & 1
		\end{bmatrix}
		\]
		Where $\det(A)\geq0$, we obtain\\
		$1-\cos^{2}\theta_{mo}-\cos^{2}\theta_{oq}-\cos^{2}\theta_{mq}+2\cos\theta_{mo}\cos\theta_{oq}\cos\theta_{mq}\geq 0$.\\
		%This implies that\\
		$(1-\cos^{2}\theta_{mo}-\cos^{2}\theta_{oq}+\cos^{2}\theta_{mo}\cos^{2}\theta_{oq})-(\cos^{2}\theta_{mo}\cos^{2}\theta_{oq}-2\cos\theta_{mo}\cos\theta_{oq}\cos\theta_{mq}+\cos^{2}\theta_{mq})\geq 0$.\\
		%Thus \\
		$(1-\cos^{2}\theta_{mo})(1-\cos^{2}\theta_{oq})-(\cos\theta_{mo}\cos\theta_{oq}-\cos\theta_{mq})^{2}\geq 0.$\\
		$\sin^{2}\theta_{mo}\sin^{2}\theta_{oq}\geq (\cos\theta_{mo}\cos\theta_{oq}-\cos\theta_{mq})^{2}.$	\\
		Taking the square roots and using $\theta_{mo},\theta_{oq}\in[0,\pi]:$\\
		$\sin\theta_{mo}\sin\theta_{oq}\geq \cos\theta_{mo}\cos\theta_{oq}-\cos\theta_{mq}$ \\
		Using $\cos(\theta_{mo}+\theta_{oq})=\cos\theta_{mo}\cos\theta_{oq}-\sin\theta_{mo}\sin\theta_{oq}$, we obtain: \\
		$\cos\theta_{mq}\geq \cos(\theta_{mo}+\theta_{oq})$.\\
		As ${\cos}^{-1}$ is decreasing, this implies \\
		$\theta_{mq}=\cos^{-1}(\cos(\theta_{mq}))\leq \cos^{-1}(\cos(\theta_{mo}+\theta_{oq}))\leq \theta_{mo}+\theta_{oq}.$ \\
		$\because \theta_{z^{l_m}_i,z^{l_q}_k} \leq \theta_{z^{l_m}_i,z^{l_o}_k} + \theta_{z^{l_o}_i,z^{l_q}_k},$ \\
		$\therefore D_{AT}(z^{l_m}_i, z^{l_n}_j, z^{l_o}_k)+D_{AT}(z^{l_o}_i, z^{l_p}_j, z^{l_q}_k)\geq D_{AT}(z^{l_m}_i, z^{l_o}_j, z^{l_q}_k)$. \\
	\end{proof}
	
	\section{Detailed network architecture}
	\label{ap:detailed_architect}
	An extensive trial-and-error experiment was conducted to determine the optimal network hyperparameters, including network depth, which considers both accuracy and simplicity. The detailed network architecture of the embedding network ($F_{\varphi}$) used for each model ($O$-Net, $S$-Net, $T$-Net, $Q$-Net, and $N$-Net) in this study is presented in Table \ref{tab:net_architecture}. The $Conv(a,b,c)$ denotes a two-dimensional convolutional layer with $a$ number of filters, $b \times b$ kernel size, and $c$ activation function, respectively. The $MaxPool(d,e)$ is a two-dimensional max-pooling layer with $d \times d$ pooling size and $e \times e$ stride size. The $BatchNorm$ represents a batch normalization layer. The $Dense(f,g)$ is a dense layer with $f$ units of neurons and $g$ activation function. The $ZeroPad(h)$ corresponds to a two-dimensional zero-padding layer with a padding size $h \times h$. The $GlobAvgPool$ is a two-dimensional global average pooling layer. The $L2Norm$ is an $L2$ normalization layer. We used an Adam optimizer with an initial learning rate of $10^{-6}$ for $O$-Net and $Q$-Net and $10^{-4}$ (default) for $S$-Net, $T$-Net, and $N$-Net, which runs for 50 epochs for the model from scratch, such as in the Busi and Finger datasets. The model with pre-trained FaceNet backbones, such as in the FG-Net and Adience datasets, is run for 20 epochs.
	
	\begin{table*}[h!]
		\centering
		\caption{Embedding network ($F_{\varphi}$) architecture of $O$-Net, $S$-Net, $T$-Net, $Q$-Net, and $N$-Net.}
		\label{tab:net_architecture}
		\resizebox{0.75\linewidth}{!}{
			\begin{tabular}{l}		
				\toprule
				\textbf{Busi \& Finger}:\\
				$Conv(64,3,Relu)-Conv(64,3,Relu)-MaxPool(2,2)-Conv(128,3,Relu)-Conv(128,3,Relu)-$ \\
				$MaxPool(3,1)-BatchNorm-Conv(256,3,Relu)-Conv(256,3,Relu)-MaxPool(3,1)-$ \\
				$BatchNorm-Conv(512,3,Relu)-Conv(512,3,Relu)-MaxPool(3,1)-BatchNorm-GlobAvgPool-$ \\
				$Dense(4096,Relu)-Dense(4096,Relu)-Dense(100,Relu)-L2Norm.$\\
				\midrule
				\textbf{FG-Net \& Adience}:\\
				$ZeroPad(5)-FaceNet-BatchNorm-GlobAvgPool-Dense(4096,Relu)-Dense(4096,Relu)-$\\
				$Dense(100,Relu)-L2Norm.$ \\
				\bottomrule
			\end{tabular}
		}
	\end{table*}
	
	The overall framework comparisons for $O$-Net, $S$-Net, $T$-Net, $Q$-Net, and $N$-Net are summarized in Table \ref{tab:model_archict_comp}. Unlike the other DML networks, as a binary classifier, the $S$-Net requires a tail layer to squeeze the output of the Euclidean distance bound to the $[0,1]$ interval. Moreover, $Q$-Net uses a neural network to learn the distance in a merging technique [13].
	
	\begin{table*}[h!]
		\centering
		\caption{Network architecture comparison of $O$-Net, $S$-Net, $T$-Net, $Q$-Net, and $N$-Net in all datasets.}
		\label{tab:model_archict_comp}
		\resizebox{0.75\linewidth}{!}{
			\centering
			\begin{tabular}{l l l l l l }		
				\toprule
				Model					&	Pairing method		& Embedding network 		&	Merging technique				&	Tail layer					&	Loss function	 				\\
				\midrule
				$O$-Net					&	Triplet 				&	$F_{\varphi}$, (Table \ref{tab:net_architecture})			&	Angular tirangle distance		&	--							&	MSE loss						\\
				\midrule
				$S$-Net					&	Duplet 	 				&	$F_{\varphi}$, (Table \ref{tab:net_architecture})			&	Euclidean distance					&	$BatchNorm-Dense(1, Sigmoid)$	&	Contrastive loss				\\
				\midrule
				$T$-Net					&	Triplet 				&	$F_{\varphi}$, (Table \ref{tab:net_architecture})			&	Squared Euclidean distance						&	--							&	Triplet loss					\\
				\midrule
				$Q$-Net					&	Quadruplet  			&	$F_{\varphi}$, (Table \ref{tab:net_architecture})			&	Neural network						&	--							&	Quadruplet loss					\\
				\midrule
				$N$-Net					&	Duplet 				&	$F_{\varphi}$, (Table \ref{tab:net_architecture})			&	Cosine similarity					&	--							&	N-pair loss						\\
				\bottomrule
			\end{tabular}
		}
	\end{table*}
	The experiment was conducted using Python (ver. 3.7.0)\footnote{License: General Public License} and Keras (ver. 2.4.3)\footnote{License: Apache License 2.0} under the TensorFlow (ver. 2.3.0)\footnote{License: Apache License 2.0} with an NVIDIA TITAN RTX 24 GB processor. 
	
	\section{Additional experiments}
	This section presents additional experiments, such as visualization of image search similarity on ordinal datasets, the effect of embedding space size on mapping accuracy, and a discussion of computation time comparison.
	
	\subsection{Image search similarity}
\begin{figure}[h!]
		\centering
		\subfloat[$O$-Net]{\includegraphics[width=0.46\linewidth]{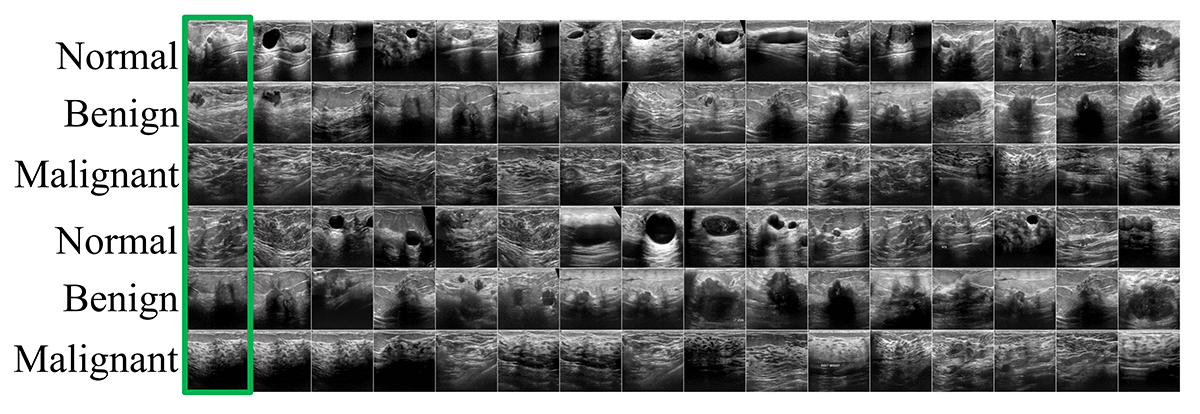}
			\label{img_search_busi_onet}}\hspace{2pt}
		\subfloat[$S$-Net]{\includegraphics[width=0.46\linewidth]{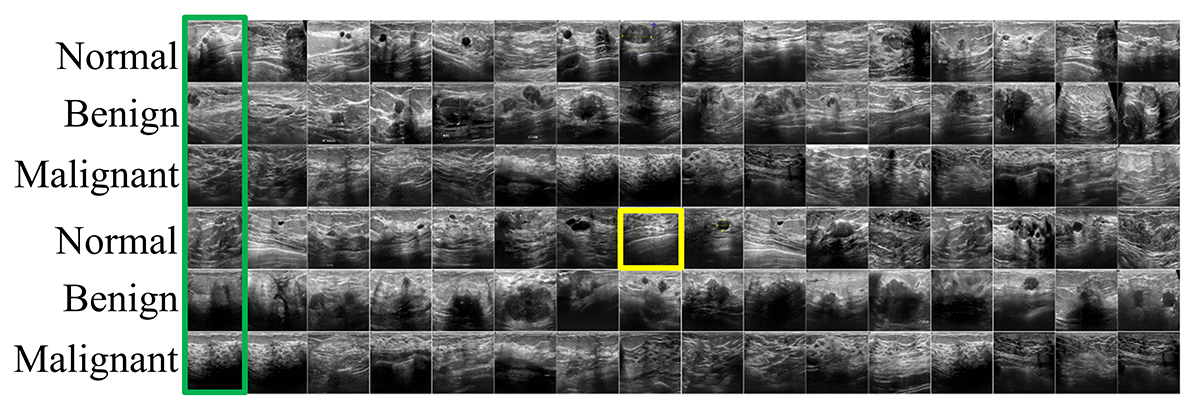}
			\label{img_search_busi_snet}}
		\subfloat[$T$-Net]{\includegraphics[width=0.46\linewidth]{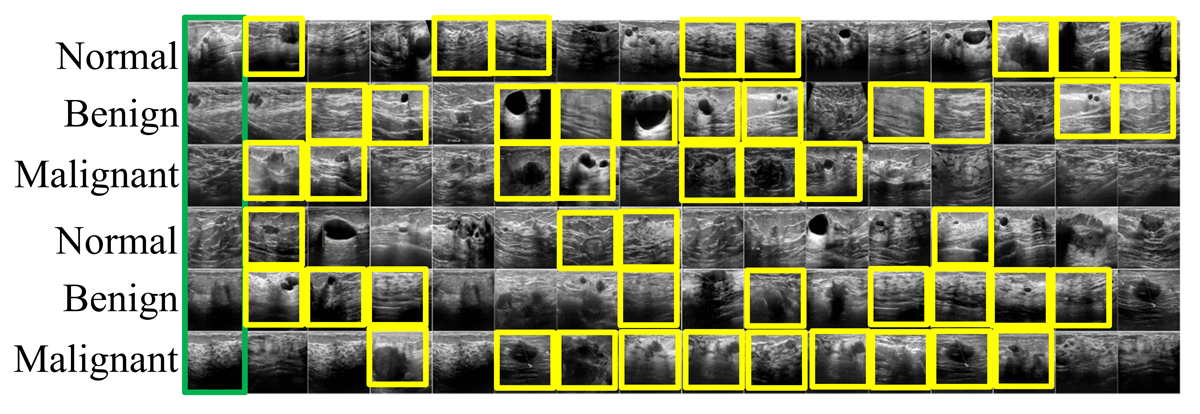}
			\label{img_search_busi_tnet}}\hspace{2pt}
		\subfloat[$Q$-Net]{\includegraphics[width=0.46\linewidth]{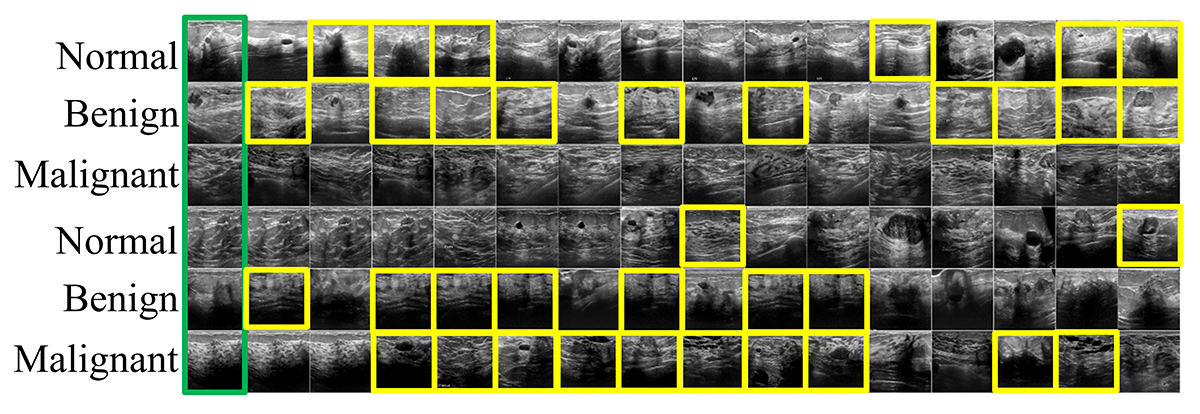}
			\label{img_search_busi_qnet}}\hspace{2pt}
		\subfloat[$N$-Net$_2$]{\includegraphics[width=0.46\linewidth]{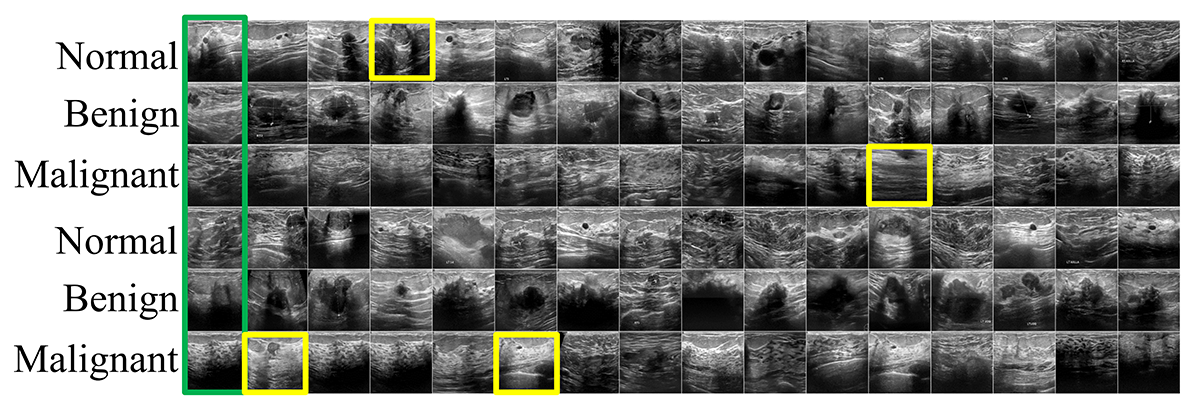}
			\label{img_search_busi_nnet}}
		\caption{Image similarity search in the Busi dataset with 15 nearest neighbors, the green and yellow boxes represent the query images, and incorrect images, respectively.}
		\label{fig:img_search_busi}
	\end{figure}

\begin{figure}[h!]
		\centering
		\subfloat[$O$-Net]{\includegraphics[width=0.45\linewidth]{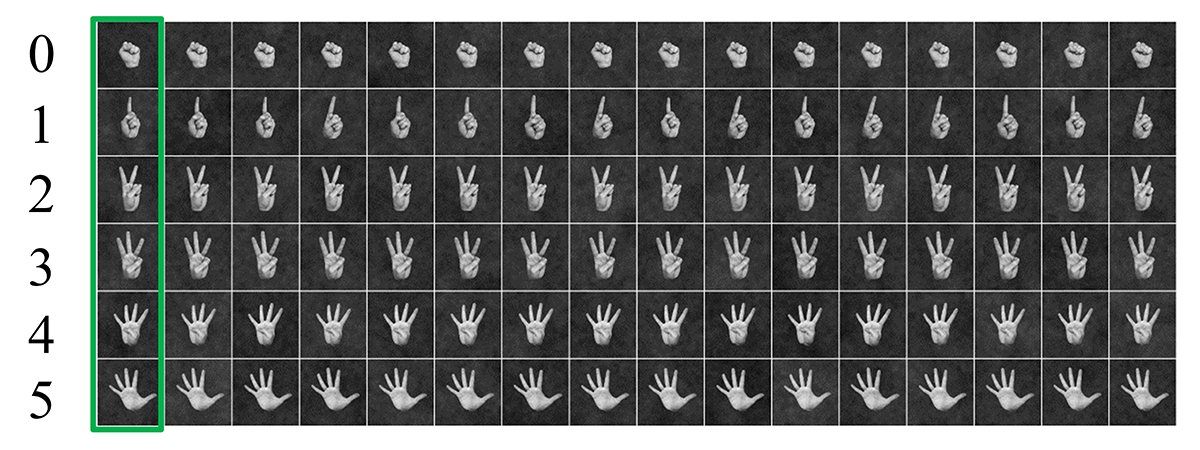}
			\label{img_search_finger_onet}}\hspace{5pt}
		\subfloat[$S$-Net]{\includegraphics[width=0.45\linewidth]{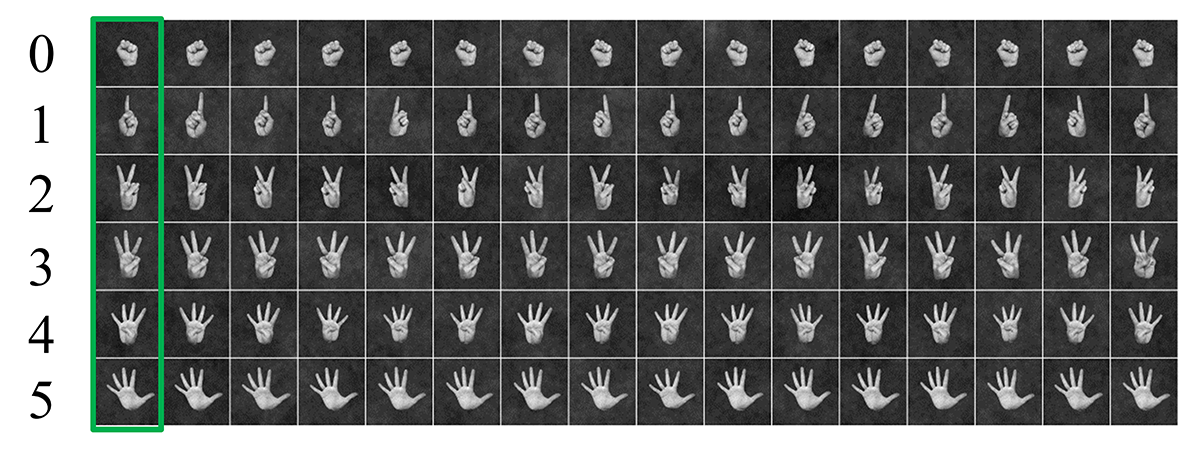}
			\label{img_search_finger_snet}}
		\subfloat[$Q$-Net]{\includegraphics[width=0.45\linewidth]{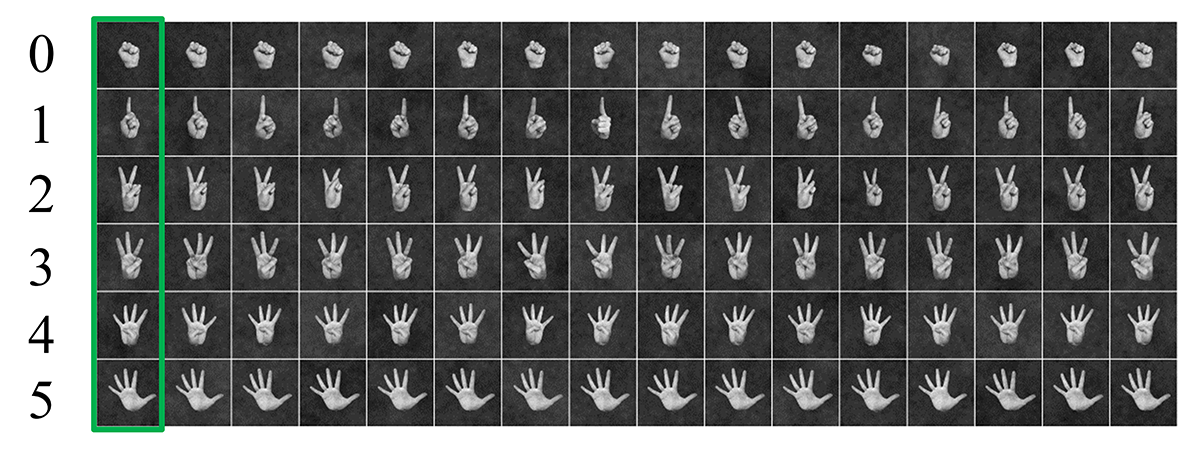}
			\label{img_search_finger_qnet}}\hspace{5pt}
		\subfloat[$T$-Net]{\includegraphics[width=0.45\linewidth]{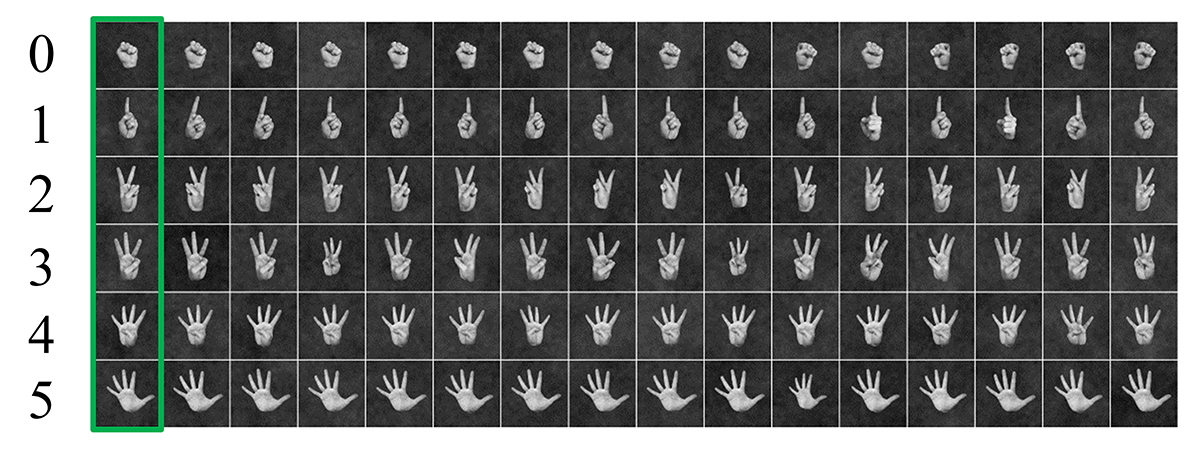}
			\label{img_search_finger_tnet}}\hspace{3pt}
		\subfloat[$N$-Net$_2$]{\includegraphics[width=0.45\linewidth]{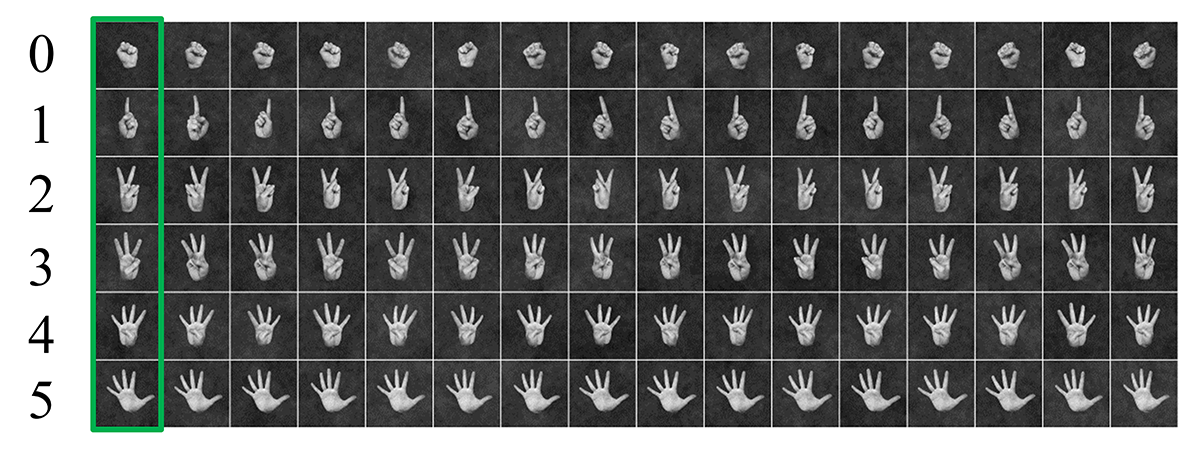}
			\label{img_search_finger_nnet}}
		\caption{Image similarity search in the Finger dataset with 15 nearest neighbors, the green and yellow boxes represent the query and incorrect images, respectively.}
		\label{fig:img_search_finger}
	\end{figure}

\begin{figure}[h!]
		\centering
		\subfloat[$O$-Net]{\includegraphics[width=0.45\linewidth]{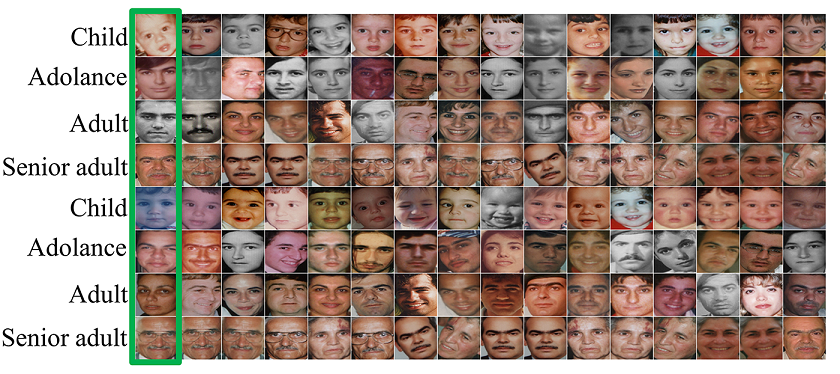}
			\label{img_search_FG-Net_onet}}\hspace{5pt}
		\subfloat[$S$-Net]{\includegraphics[width=0.45\linewidth]{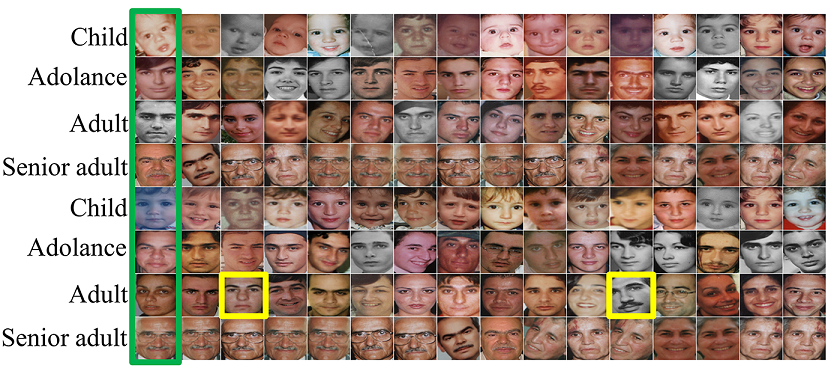}
			\label{img_search_FG-Net_snet}}
		\subfloat[$T$-Net]{\includegraphics[width=0.45\linewidth]{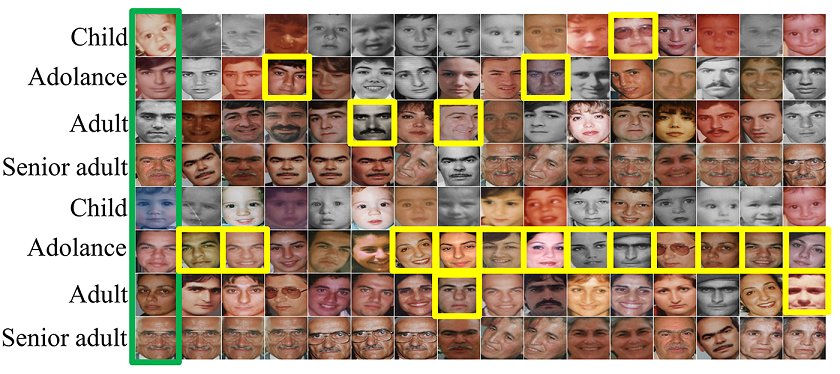}
			\label{img_search_FG-Net_tnet}}\hspace{5pt}
		\subfloat[$Q$-Net]{\includegraphics[width=0.45\linewidth]{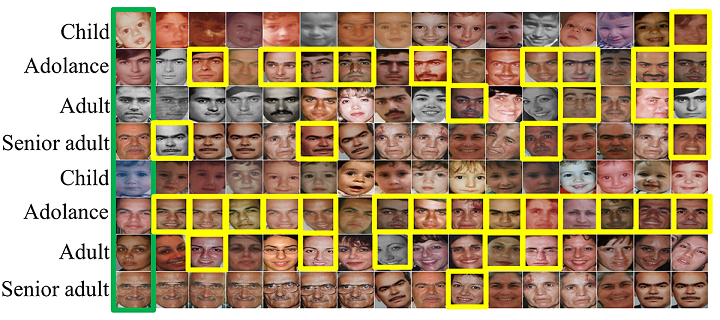}
			\label{img_search_FG-Net_qnet}}\hspace{2pt}
		\subfloat[$N$-Net$_2$]{\includegraphics[width=0.45\linewidth]{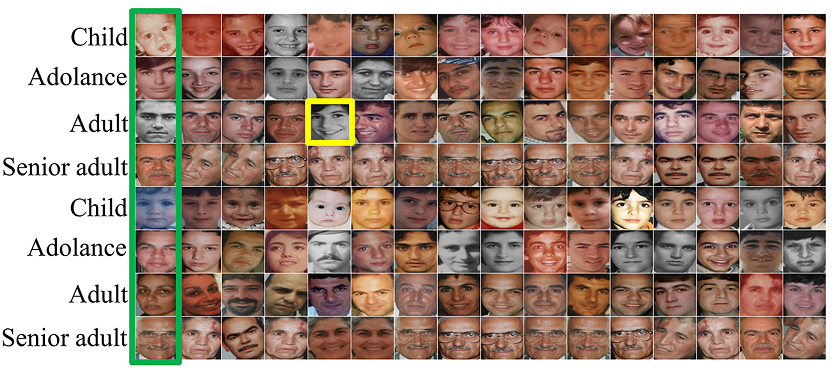}
			\label{img_search_FG-Net_nnet}}
		\caption{Image similarity search in the FG-Net dataset with 15 nearest neighbors, the green and yellow boxes represent the query and incorrect images, respectively.}
		\label{fig:img_search_FG-Net}
	\end{figure}

\begin{figure}[h!]
		\centering
		\subfloat[$O$-Net]{\includegraphics[width=0.45\linewidth]{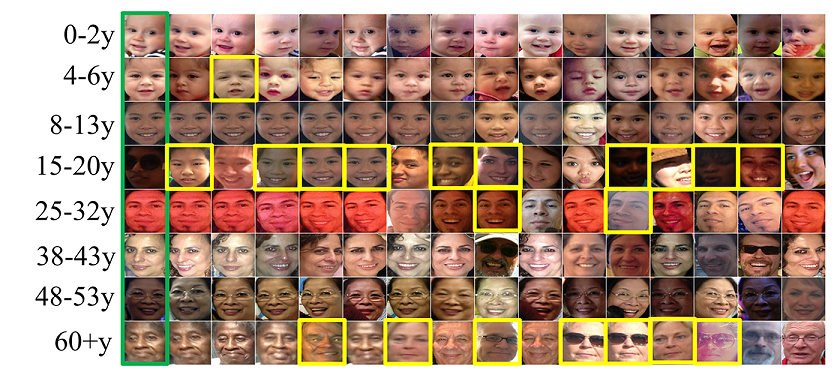}
			\label{img_search_adience_onet}}\hspace{5pt}
		\subfloat[$S$-Net]{\includegraphics[width=0.45\linewidth]{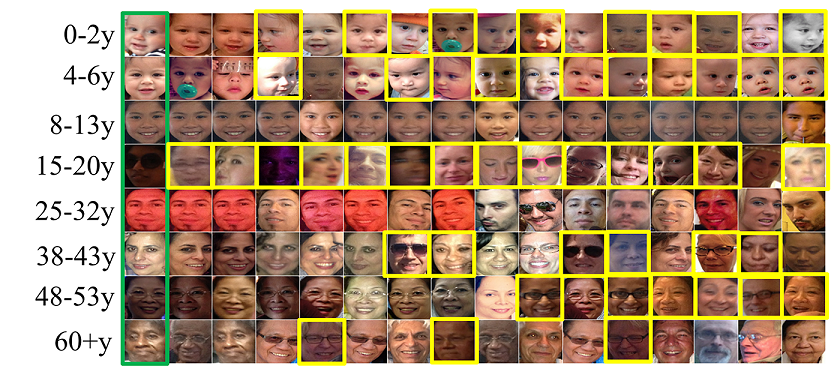}
			\label{img_search_adience_snet}}
		\subfloat[$T$-Net]{\includegraphics[width=0.45\linewidth]{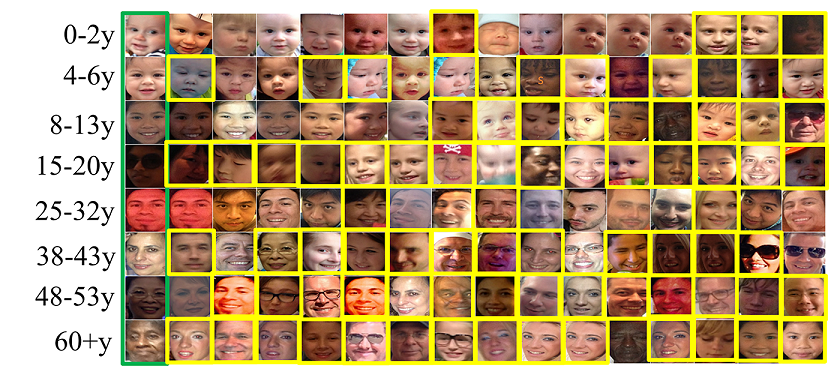}
			\label{img_search_adience_tnet}}\hspace{5pt}
		\subfloat[$Q$-Net]{\includegraphics[width=0.45\linewidth]{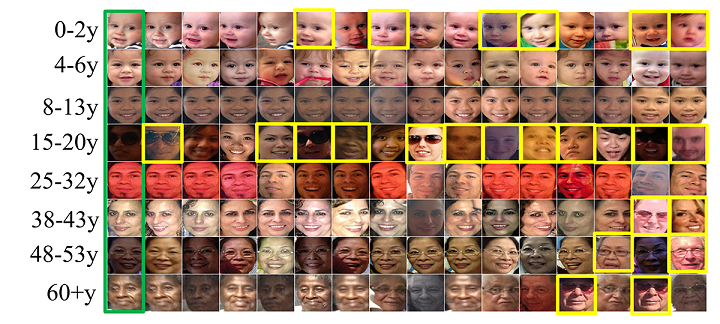}
			\label{img_search_adience_qnet}}\hspace{2pt}		\subfloat[$N$-Net$_2$]{\includegraphics[width=0.45\linewidth]{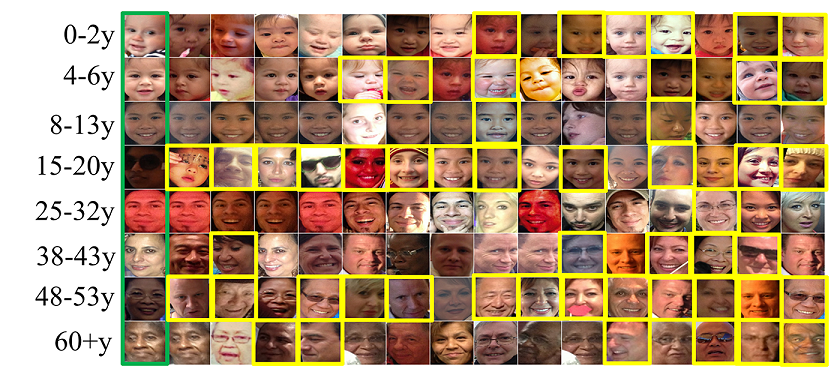}
			\label{img_search_adience_nnet}}
		\caption{Image similarity search in the Adience dataset with 15 nearest neighbors, the green and yellow boxes represent the query and incorrect images, respectively.}
		\label{fig:img_search_adience}
	\end{figure}

	We used the $K$-nearest neighbor algorithm with cosine similarity to obtain 15 similar images from a query image in the image search similarity task. The results are depicted in Figs. \ref{fig:img_search_busi}, \ref{fig:img_search_finger}, \ref{fig:img_search_FG-Net}, and \ref{fig:img_search_adience} for the Busi, Finger, FG-Net, and Adience datasets, respectively. In the figures, the first column (green box) is a query image, and the remaining columns represent its 15 nearest-neighbor images.
	
	\vspace{-1pt}
	\vspace{-1pt}
	\vspace{-1pt}
	
	%	\subsubsection{Ordinal regression}
	%	We conducted an ordinal regression task to assess the performance of $O$-Net in the larger categories (ordinal-continuous data). We used Asian Face Database (AFAD), which was introduced by Niu et al. \cite{7780901}. The AFAD dataset contains $165,501$ faces in the range of $15-40$ years. We used the same setting as in \cite{5995437,7780901,CAO2020325}, where each image was randomly divided into 80\% training data and 20\% test data. Due to the O-Net being devised for the DML model, we employed it as a backbone. We used the existing ordinal ranking loss (CORAL) to conduct an ordinal regression task with a convolutional neural network. The result and comparison against the state-of-the-art model are shown in Table \ref{tab:ordinal_reg}. We can infer that adding $O$-Net as a backbone has a competitive result with the existing benchmark.
	
	%	\begin{table}[h]
	%		\makegapedcells
	%		\centering
	%		\caption{Age prediction error in the AFAD dataset. The performance is measured by the mean absolute error (MAE) metric.}
	%		\label{tab:ordinal_reg}
	%		\resizebox{0.275\linewidth}{!}{
	%			\centering
	%			\begin{tabular}{l l}		
	%				\toprule
	%				Model 						 & 	MAE 	\\
	%				\midrule
	%				BIFs + OHRank \cite{5995437} &	3.84	\\
	%				MR-CNN \cite{7780901} 		 &	3.51	\\
	%				OR-CNN \cite{7780901} 		 &	3.34	\\
	%				CE-CNN \cite{CAO2020325} 	 &	3.60	\\
	%				CORAL-CNN \cite{CAO2020325}  &	3.47	\\
	%				$O$-Net + CORAL 			 &	3.52	\\
	%				\bottomrule
	%			\end{tabular}
	%		}
	%		\end{table}
	
	\subsection{Impact of embedding space size}
\begin{figure}[h!]
		\centering
		\begin{pics}
			\centering
			\begin{tikzpicture}[scale=0.3,every node/.style={scale=1.8}]
				\begin{axis}[title=Busi, legend pos=south east,
					xmin=0,   xmax=200,
					ymin=0.935,   ymax=0.965,
					xlabel=$|Z|$,
					ylabel=SVM Accuracy, every axis plot/.append style={ultra thick},yticklabel style={/pgf/number format/fixed,/pgf/number format/precision=5}]
					\addplot [solid,
					color=red,
					thick,
					mark=*,
					mark options={thick,solid,scale=1.5, black}, %<--- Change
					mark layer=like plot,      %<--- Change
					/pgfplots/error bars/.cd,
					x dir=none,
					y dir=both,
					y explicit,
					error bar style={color=black,dashed,ultra thick}]
					table[x =x, y =y, y error =err]{
						x	y	err
						10	0.946153846	0.002732012
						100	0.950235654	0.002519387
						190	0.953846154	0.001933070
					};
					\legend{Mean accuracy}
				\end{axis}
			\end{tikzpicture}&
			\begin{tikzpicture}[scale=0.3,every node/.style={scale=1.8}]
				\begin{axis}[title=Finger, legend,
					xmin=0,   xmax=200,
					ymin=0.8,   ymax=1.2,
					xlabel=$|Z|$,
					ylabel=SVM Accuracy, every axis plot/.append style={ultra thick},yticklabel style={/pgf/number format/fixed,/pgf/number format/precision=5},scaled ticks=false]
					\addplot [solid,
					color=red,
					thick,
					mark=*,
					mark options={thick,solid,scale=1.5, black}, %<--- Change
					mark layer=like plot,      %<--- Change
					/pgfplots/error bars/.cd,
					x dir=none,
					y dir=both,
					y explicit,
					error bar style={color=black,dashed,ultra thick}]
					table[x =x, y =y, y error =err]{
						x	y	err
						10	1.0	0.00
						100	1.0	0.00
						190	1.0	0.00
					};
					\legend{Mean accuracy}
				\end{axis}
			\end{tikzpicture}\\
			
			\begin{tikzpicture}[scale=0.3,every node/.style={scale=1.8}]
				\begin{axis}[title=FG-Net, legend pos=south east,
					xmin=0,   xmax=200,
					ymin=0.90,   ymax=0.99,
					xlabel=$|Z|$,
					ylabel=SVM Accuracy, every axis plot/.append style={ultra thick},yticklabel style={/pgf/number format/fixed,/pgf/number format/precision=5}]
					\addplot [solid,
					color=red,
					thick,
					mark=*,
					mark options={thick,solid,scale=1.5, black}, %<--- Change
					mark layer=like plot,      %<--- Change
					/pgfplots/error bars/.cd,
					x dir=none,
					y dir=both,
					y explicit,
					error bar style={color=black,dashed,ultra thick}]
					table[x =x, y =y, y error =err]{
						x	y	err
						10	0.926765068	0.007363103
						100	0.961027421	0.006429082
						190	0.961067528	0.006643089
					};
					\legend{Mean accuracy}
				\end{axis}
			\end{tikzpicture}&
			\begin{tikzpicture}[scale=0.3,every node/.style={scale=1.8}]
				\begin{axis}[title=Adience, legend,
					xmin=0,   xmax=200,
					ymin=0.465,   ymax=0.7,
					xlabel=$|Z|$,
					scaled ticks=false,
					ylabel=SVM Accuracy, every axis plot/.append style={ultra thick},yticklabel style={/pgf/number format/fixed,/pgf/number format/precision=5}]
					\addplot [solid,
					color=red,
					thick,
					mark=*,
					mark options={thick,solid,scale=1.5, black}, %<--- Change
					mark layer=like plot,      %<--- Change
					/pgfplots/error bars/.cd,
					x dir=none,
					y dir=both,
					y explicit,
					error bar style={color=black,dashed,ultra thick}]
					table[x =x, y =y, y error =err]{
						x	y	err
						10	0.534123923	0.013508917
						100	0.578019761	0.01598917	
						190	0.622290819	0.015091158
					};
					\legend{Mean accuracy}
				\end{axis}
			\end{tikzpicture}
		\end{pics}
		\caption{Effect of adjusting embedding space size $|Z|$ to latent code separability.} \label{fig:effect_z}
	\end{figure}
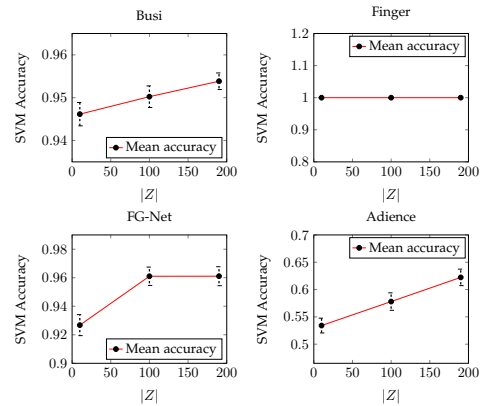

	As shown in Table \ref{ap:detailed_architect}, the default embedding latent size $|Z|$ is 100. In this section, we vary $|Z|$ and observe the embedding-space mapping accuracy using the SVM classifier, as shown in Fig. \ref{fig:effect_z}. The results reveal that increasing $|Z|$ value can also increase the accuracy, except in the Finger dataset, which is a highly separable dataset.
	
	\subsection{Computation time} 
	Finally, we report the computation time comparison for each method in Table \ref{tab:computation_time}. In DML, the computation time is mainly compressed into two phases: the pairing process and network training time, as shown in Table \ref{tab:model_archict_comp}. Unlike $T$-Net and $N$-Net, $O$-Net and $S$-Net generate labels in the pairing process; thus, they have higher computational costs than $T$-Net and $N$-Net. Since $Q$-Net has a quadruple input, it requires a slower computation time than duplet ($S$-Net) and triplet ($T$-Net). Our $O$-Net has the slowest computation time in the pairing process because it uses more combinations than others in pairing the triplet representation. Nevertheless, the overall time complexity is comparable with the existing DML methods. It is worth noting that $N$-Net computes the loss value solely once per batch; hence training the network requires less computational effort than others. However, $N$-Net achieves an inadequate result with the default batch ($N$-Net$_1$), as explained in Section 4.2. Moreover, with a large batch ($N$-Net$_2$), $N$-Net achieves the highest time complexity for DML models. 
	
	\begin{table}[h!]
		\centering
		\caption{Comparison of computation time (seconds) of each method in the Busi dataset.}
		\label{tab:computation_time}
		\resizebox{1\linewidth}{!}{
			\centering
			\begin{tabular}{l l l l l l l}		
				\toprule
				\multirow{2}{*}{Phase} & \multicolumn{6}{c}{Computation time (mean $\pm$ std.)}\\
				\cmidrule{2-7}
				&	 $O$-Net 								&	 $S$-Net								&	 $T$-Net								&	 $Q$-Net								&	 $N$-Net$_1$								&	 $N$-Net$_2$				\\
				\midrule
				Pairing process     &	$0.767\pm0.014$						&	$0.511\pm0.009$						&	$0.400\pm0.007$						&	$0.531\pm0.007$						&	$0.401\pm0.009$ 					&	$0.401\pm0.009$		\\
				Network training    &	\multirow{2}{*}{$21.010\pm0.170$}	&	\multirow{2}{*}{$13.796\pm0.118$}	&	\multirow{2}{*}{$11.529\pm0.085$}	&	\multirow{2}{*}{$14.095\pm0.175$}	&	\multirow{2}{*}{$2.876\pm0.018$} 	&	\multirow{2}{*}{$30.167\pm0.264$}	\\
				(per-epoch)    		&										&										&										&										&		 								&	 									\\
				\bottomrule
			\end{tabular}
		}
	\end{table}

	% or
	

	\section*{Acknowledgment}
	This work was supported by the National Research Foundation of Korea (NRF) grant funded by the Korea government (MSIT) (No. 2020R1A2C1102294).

	
	
	%\bibliographystyle{IEEEtran}
	\bibliographystyle{IEEEtran}
	\bibliography{bare_jrnl_compsoc}
	

	
	%\vfill
	

\end{document}